\documentclass[final,1p,times,nopreprintline]{elsarticle}
\journal{Neurocomputing}

\usepackage{amsmath,amssymb,amsfonts}
\usepackage{algorithmic}
\usepackage{graphicx}
\usepackage{textcomp}
\usepackage{xcolor}
\usepackage{pgffor}  % \foreach

\graphicspath{{Figs/}{Figs/Results/}}
\usepackage[makeroom]{cancel} 

\DeclareMathOperator{\sgn}{sgn}
\usepackage{multirow} 
\DeclareMathOperator*{\argmin}{arg\,min} 

\usepackage{siunitx} 
\usepackage{gensymb} 
\usepackage{changepage} % \adjustwidth
\usepackage{xurl} % break url
\newcolumntype{x}[1]{>{\centering\arraybackslash\hspace{0pt}}p{#1}} % fixed size centered column
\usepackage[colorlinks=true, linkcolor=blue, citecolor=blue, urlcolor=blue]{hyperref}
\usepackage{caption}   % For caption handling

\begin{document}
	
	\begin{frontmatter}
		
		\title{Addressing imperfect symmetry: \\A novel symmetry-learning actor-critic extension}
		
		\tnotetext[preprint]{Preprint. Published in \textit{Neurocomputing}. \url{https://doi.org/10.1016/j.neucom.2024.128771}}
		
		\author[label1]{Miguel Abreu}
		\author[label1]{Lu\' {i}s Paulo Reis}
		\author[label2]{Nuno Lau}
		
		\affiliation[label1]{organization={LIACC/LASI/FEUP, Artificial Intelligence and Computer Science Laboratory, Faculty of Engineering, \\ University of Porto},%Department and Organization
			%addressline={}, 
			city={Porto},
			%postcode={}, 
			%state={},
			country={Portugal}}
			
		\affiliation[label2]{organization={IEETA/LASI/DETI, Institute of Electronics and Informatics Engineering of Aveiro, Department of Electronics, Telecommunications and Informatics, University of Aveiro},%Department and Organization
			%addressline={}, 
			city={Aveiro},
			%postcode={}, 
			%state={},
			country={Portugal}}
		
		\begin{abstract}
			Symmetry, a fundamental concept to understand our environment, often oversimplifies reality from a mathematical perspective. Humans are a prime example, deviating from perfect symmetry in terms of appearance and cognitive biases (e.g. having a dominant hand). Nevertheless, our brain can easily overcome these imperfections and efficiently adapt to symmetrical tasks. The driving motivation behind this work lies in capturing this ability through reinforcement learning. To this end, we introduce Adaptive Symmetry Learning (ASL) --- a model-minimization actor-critic extension that addresses incomplete or inexact symmetry descriptions by adapting itself during the learning process. ASL consists of a symmetry fitting component and a modular loss function that enforces a common symmetric relation across all states while adapting to the learned policy. The performance of ASL is compared to existing symmetry-enhanced methods in a case study involving a four-legged ant model for multidirectional locomotion tasks. The results show that ASL can recover from large perturbations and generalize knowledge to hidden symmetric states. It achieves comparable or better performance than alternative methods in most scenarios, making it a valuable approach for leveraging model symmetry while compensating for inherent perturbations.
		\end{abstract}

		%%Research highlights

		
	\end{frontmatter}

\section{Introduction} \label{sec:introduction}

Symmetry is a multidisciplinary concept that plays a fundamental role in how we perceive the world, from abstract beauty and harmony, up to precise definitions in physics, biology, mathematics, and many other disciplines. In the scope of this paper, symmetry is studied through geometry, and, more precisely, group theory. Weyl \cite{weyl} takes a top-down approach to define symmetry, starting with a vague notion of proportion and balance, then delving into the intuitive notion of reflections and rotations, and ending with the characterization of invariance with respect to a group of automorphisms (i.e. mappings of a mathematical object onto itself that preserve the structure of space).  

Weyl acknowledges that in reality, the actual group of automorphisms may be unknown, leading researchers to narrow the invariant conditions, for a precise analysis to be possible. In the context of this paper, let us start with a biology example that doubles as motivation. Consider the act of throwing a crumpled paper into a bin at a reasonable distance. Assume a bilaterally symmetrical non-ambidextrous human. The dominant arm/hand will easily find several trajectories for the paper to reach the bin and, with enough training, the other arm will be able to find equivalent mirrored trajectories.

This means that the human brain is able to learn how to compensate for internal imbalances to produce a symmetric outcome. Internal imbalances include differences in arm strength, weight and other physical attributes, as well as handedness and other cognitive biases. Additionally, the brain may also have to compensate for external factors, e.g. wearing a thick glove on the left hand. 

\subsection{Symmetry Perturbations}

These symmetry perturbations do not imply an asymmetric relation between both actions. Assume the human's initial state $s_0$ is a bilaterally symmetrical pose. The motor cortex generates muscle contractions by producing neural activity patterns. The patterns that result in throwing with the dominant arm/hand represent action $a_d$, while the patterns that deal with the other arm represent action $a_z$. There might be a map $g_{s_0}:\mathcal{A}_{s_0}\rightarrow\mathcal{A}_{s_0}$, where, in state $s_0$, for every action $a_d \in \mathcal{A}_{s_0}$ there is an action $a_z \in \mathcal{A}_{s_0}$ that results in a symmetric outcome, where $\mathcal{A}_{s_0}$ is a group of symmetrizable actions in state $s_0$. In this context, actions are not symmetrizable when, for instance, a shorter arm cannot perform the same role as a longer arm.

When the map's estimate $\hat{g}_{s_0}$ is a simplification of the ground truth, we say that the symmetry description it provides is incomplete or inexact. Symmetry perturbation, hereinafter referred to simply as perturbation, is any factor that stands between $\hat{g}_{s_0}$ and $g_{s_0}$, affecting the mapping between pairs of symmetric actions, i.e., actions that lead to two symmetric states when applied to the same initial state. Note that this analysis only makes sense for state-action pairs, as the invariance that characterizes symmetric actions concerns the state that follows. These notions will be formalized in Section \ref{sec:prelim}, while defining Markov Decision Processes and their homomorphisms.

Despite the search for symmetry for aesthetic or functional purposes, our brain has evolved to be asymmetric \cite{mainzer2005}, both in terms of specialized regions and motor skills. About 9.33\% of humans are mixed-handed, while only around 1\% or 2\% are ambidextrous, depending on the classification criteria \cite{handedness}. Symmetry might not always mean better performance, and this uncertainty must be taken into account. Translating this concept to reinforcement learning (RL), it would mean that symmetry should guide the learning process, but never to the detriment of the primary optimization objective.

\subsection{Model Minimization} \label{sec:model_min}

This paper presents a case study of a four-legged ant model with a total order of symmetry of 8, used to learn multidirectional locomotion tasks under the influence of multiple symmetry perturbations. The robot's symmetry group includes 4 planes of symmetry and 4-fold rotational symmetry, where one rotation is an identity operation, resulting in 7 non-identity transformations. This is a challenging task, where finding a good policy in an efficient way heavily depends on how the learning algorithm explores the symmetry group.

Most symmetry leveraging techniques fall within temporal or spatial domains. The former, which pertains to invariability under temporal transformations, such as scaling or inversion, is relatively uncommon in robotic tasks.  Techniques centered around spatial symmetry are more prevalent, and can be divided into multiple areas, as described in following paragraphs. 

\subsubsection{Relabeling states and actions}

The first spatial symmetry technique concerns the permutation of roles of equivariant parts in the state and action spaces. For instance, consider a bilaterally symmetrical robot that throws a ball with one arm, while stabilizing itself with the other. The policy could learn independent actions for a left throw and a right throw. However, it could abstract the side by just considering a \textit{throwing arm} and a \textit{stabilizing arm}. The user must relabel both arms when assigning the state, depending on the preferred throwing side for the next episode. After the policy computes an action, the user has to relabel them again to assign the resulting action correctly to both arms. This approach can be implemented directly in the simulator without requiring modifications to the optimization algorithm.

\subsubsection{Data augmentation}

Data augmentation is commonly used in RL to enhance sample efficiency and stability, being experience replay, introduced by Lin~\cite{lin1992}, one of its main precursors. In the scope of this work, data augmentation involves the creation of symmetric copies of real experiences. Using the last example, after the robot threw the ball with its left arm, the RL algorithm would also receive artificially created samples of the same experience, but executed with the right arm. Ultimately, there is twice as much information to learn from. 

\subsubsection{Symmetric Networks}

This approach enforces symmetry constraints directly on the policy, by modifying the network architecture. In domains where symmetry is a requirement of optimal solutions, this is a robust method. However, this assumption cannot be made for many robotic tasks, including locomotion, where it is unclear whether perfect symmetry is mechanically preferable. Real robots exhibit imperfections like humans, who learn to compensate for asymmetries and perceive the resulting gaits as normal or unimpaired \cite{browne21,handvzic15}.

\subsubsection{Symmetry Loss Function}

In optimization problems, a loss function quantifies the cost of an event. In RL, this function is responsible for maximizing a cumulative reward over time, while also sustainably guiding the learning process. In the ball throwing robot example, the reward could measure how close the ball is to the target. Also, if symmetry was desired, it could be added to the reward as a scalar value, but there is a better alternative.

A reward is useful when evaluating a metric that requires interaction with the environment. In contrast, symmetry does not have this limitation, as its compliance can be measured through analytical functions that take the policy as input. By incorporating this knowledge into the primary loss function, the algorithm is able to steer the policy's parameters in the right direction through a gradient.

Despite not providing symmetry guarantees, like the relabeling method or symmetric networks, a loss function is the most flexible approach, both at design and runtime stages. It allows users to dynamically regulate the importance of symmetry and fine-tune the learning process. However, current contributions fall short in addressing symmetry perturbations. To bridge this gap, our work introduces a novel approach to leverage model symmetry, while learning and compensating for symmetry perturbations in the system. 

The main contributions of this work can be summarized as:

\begin{enumerate}
	\item We introduce Adaptive Symmetry Learning (ASL) \footnote{\url{https://github.com/m-abr/Adaptive-Symmetry-Learning}} \;---\; a model-minimization actor-critic extension that is able to handle incomplete or inexact symmetry descriptions by adapting itself during the learning process;
	\item ASL is composed of two parts: symmetry fitting, described above; and a new loss function, capable of applying the learned descriptions based on their importance, while actively avoiding neutral states and disadvantageous updates;
	\item We propose modifications to two existing symmetry loss functions ---  MSL \cite{yu2018learning} and PSL \cite{psl} --- extending them with value losses and the capacity to handle involutory transformations, such as most rotations;
	\item We present the case study of an ant robot, where we compare the performance of MSL, PSL, ASL, and vanilla PPO \cite{schulman2017ppo} in a set of ant locomotion scenarios. We test controlled and realistic symmetry perturbations, as well as partial goal observation, where the policy must discover the best symmetric gait without exploring the symmetric side, while acknowledging the existence of perturbations.
\end{enumerate}

\section{Preliminaries} \label{sec:prelim}

Typical reinforcement learning problems can be described as a Markov Decision Process (MDP) -- a tuple $\left\langle \mathcal{S,A},\Psi,p,r\right\rangle$, with a set of states $\mathcal{S}$, a set of actions $\mathcal{A}$, a set of possible state-action pairs $\Psi\subseteq \mathcal{S}\times \mathcal{A}$, a transition function $p(s,a,s'):\Psi \times \mathcal{S} \rightarrow[0,1]$, and a reward function $r(s,a):\Psi \rightarrow {\rm I\!R}$.

\subsection{MDP transformations} \label{sec:prelim_MDP_trans}

Model reduction allows the exploitation of redundant features, such as symmetry. To this end, Ravindran and Barto~\cite{ravindran2001} proposed a mathematical formalism to describe MDP homomorphisms --- a transformation that groups equivalent states and actions. An MDP homomorphism $h$ from $M\doteq\left\langle \mathcal{S,A},\Psi,p,r\right\rangle$ to $\check{M}\doteq\left\langle \mathcal{\check{S},\check{A}},\check{\Psi},\check{p},\check{r}\right\rangle$ can be defined as a surjection $h:\Psi \rightarrow \check{\Psi}$, which is itself defined by a tuple of surjections $\left\langle f,\{g_s|s\in \mathcal{S}\} \right\rangle$. In other words, equivalent state-action pairs in $M$ are mapped by $h$ to the same abstract state-action pair in $\check{M}$. For $(s,a)\in \Psi$, the surjective function $h((s,a))=(f(s),g_s(a))$, where $f:\mathcal{S}\rightarrow \mathcal{\check{S}}$ and $g_s:\mathcal{A}_s\rightarrow \mathcal{\check{A}}_{f(s)}$ for $s\in \mathcal{S}$, satisfies

\begin{align}
\check{p}(f(s),g_s(a),f(s')) & = \sum_{s''\in [s']_B}p(s,a,s''), \label{eq:homTran} \\ 
& \quad\; \forall s, s' \in \mathcal{S},a\in \mathcal{A}_s,  \nonumber
\\
\mathrm{and} \quad  \check{r}(f(s),g_s(a)) & =r(s,a), \quad \forall s \in \mathcal{S}, a \in \mathcal{A}_s,  \label{eq:homRew}
\end{align}

\noindent
where $B$ is a partition of $\mathcal{S}$ into equivalence classes, and $[s']_B$ denotes the block of partition $B$ to which state $s'$ belongs.

MDP symmetries constitute a specialization of the described framework, where $f$ and $g_s, s\in \mathcal{S}$ are bijective functions and, consequently, the homomorphism $h=\left\langle f,\{g_s|s\in \mathcal{S}\} \right\rangle$ from $M$ to $\check{M}$ is an isomorphism. Additionally, since symmetries can be characterized as MDP isomorphisms from and to the same MDP, they are automorphisms, which simplifies the homomorphism conditions~\eqref{eq:homTran} and~\eqref{eq:homRew} into

\begin{align}
p(f(s),g_s(a),f(s')) & = p(s,a,s'), \quad \forall s, s' \in \mathcal{S},a\in \mathcal{A}_s, \label{eq:homTran2}
\\[3pt]
\mathrm{and} \quad  r(f(s),g_s(a)) & =r(s,a), \quad \forall s \in \mathcal{S}, a \in \mathcal{A}_s. \label{eq:homRew2}
\end{align}

When referring to $g_s(a)$, the state $s$ corresponding to the state-action pair $(s,a)\in\Psi$ is implicit, and will thus be omitted in the remainder of this document, being the action transformation denoted as $g(a)$. As MDP automorphisms, symmetries can be defined in a single MDP, $M$. In the aforementioned ball throwing robot example, assume that starting with the ball on the left hand (state $s$), the robot will perform an optimal left throw (action $a$). In a mirrored initial state $f(s)$, a mirrored action $g(a)$ is guaranteed to be optimal if all states in $\Psi$ are symmetrizable. Equation \eqref{eq:homTran2} holds in this scenario because the probability of ending in an optimal state by performing action $a$ in $s$ is the same as performing $g(a)$ in $f(s)$. As for \eqref{eq:homRew2}, it also holds, since the reward in both cases should be the same, otherwise it would contain an asymmetric bias.

\subsection{Policy notation} \label{sec:policy_notation}

When optimizing a policy $\pi$ parameterized by $\theta$, the notion of symmetric action has two possible interpretations: $g(a)$, which is the result of mapping $a$ through function $g$; and $\pi_\theta(f(s_t))$, which represents the action chosen by the policy for the symmetric state. Frequently, the former is used as a target value for the latter. Stochastic policies that follow a normal distribution ${\mathcal {N}}(\mu ,\sigma)$ are characterized by a mean $\mu$, usually parameterized in RL by a neural network, and a standard deviation $\sigma$, optimized independently for exploration. The parameterized mean value $\mu_{\theta}$ is particularly useful for symmetry losses, as it represents the current policy without exploration noise, or, more formally, $\mu_{\theta}(\cdot) = \pi_{\theta}(\cdot \mid \sigma=0)$. Hereinafter, $a_t$ denotes an action sampled from a stochastic policy $\pi_\theta$ at time $t$, while $\overline{a}_t$ represents the mean of the distribution when $a_t$ was sampled. Extending this notation to the already introduced symmetry mapping yields:

\begin{align}
    \overline{a}_t &\doteq \mu_{\theta_\text{old}}(s_t), \label{eq:mean_at}
    \\[3pt]
    \overline{a}'_t &\doteq \mu_{\theta_\text{old}}(f(s_t)), \label{eq:mean_a't}
\end{align}

\noindent where $\theta_\text{old}$ means that $\overline{a}_t$ and $\overline{a}'_t$ were obtained from a snapshot of $\theta$ and thus are not a function of the current $\theta$. In other words, variables obtained from $\theta_\text{old}$ are constants, not trainable during optimization.

\subsection{Symmetric policy} \label{sec:sym_policy}

Zinkevich and Balch \cite{Zinkevich} define that for a policy to be functionally homogeneous, if $\pi(s)=a$, then $\pi(f(s))=g(a)$, for all $(s,a)\in \Psi$. When dealing with reflection symmetries, objects are invariant under involutory transformations, i.e., obtaining the symmetric object and the original again just requires applying the same transformation twice. Mathematically, a function $f$ that maps reflection symmetries is an involution ($f=f^{-1}$). In this context, a policy $\pi$ is symmetric if $\pi(s)=g(\pi(f(s))), \forall s \in \mathcal{S}$.

However, some symmetry operations, including scaling, rotation and translation cannot always be inverted through the same function. Rotation around an axis, for instance, is only involutory when the rotation angle $r$ in radians is a multiple of $\pi$ ($r=k\pi, k \in \mathbb{Z}$). For example, if $r=\pi/4$, the previous symmetric policy definition fails because after the state is rotated by $\pi/4$ radians, the action is also rotated by the same angle, not yielding the original action.

Generalizing the symmetric policy definition to non-involutory transformations only requires changing the side of one operation, such that $g(\pi(s))=\pi(f(s)), \forall s \in \mathcal{S}$. Care has to be taken when devising symmetry loss functions, to abide by this principle and avoid the negative effect that non-involutory transformations would otherwise cause.

\subsection{Neutral states} \label{sec:prelim_neutral}

A state $s$ is said to be neutral if it is invariant under $f$, i.e., $s=f(s)$. Under a symmetric policy, neutral states are inescapable, in the sense that all future states are guaranteed to also be neutral, unless the environment introduces its own bias. This problem occurs in numerous scenarios, from board games to robotic tasks. Consider the problem of a humanoid robot trying to initiate walking by taking its first step, but at the beginning of the episode, the environment is in a neutral state, $s_0=f(s_0)$. The symmetric policy cannot raise the left foot to take the first step in $s_0$ because that would mean the right foot should be raised in $f(s_0)$, but this is impossible because there are two actions for the same state and $\pi(s_0)\neq g(\pi(s_0))$. So, one possible option would be to jump with both legs, but after that, it would still be stuck in a neutral state unless the environment introduced some noise. 

Perfectly symmetric policies are therefore inadequate for neutral states. Moreover, neighboring states are also affected due to the generalization ability of neural networks and the fact that they represent continuous functions. A solution based on a symmetry exclusion region around neutral states is presented in Section \ref{sec:neutral_exclusion}.

\section{Related work}

This section builds upon the model minimization overview from Section~\ref{sec:model_min} and explores specific techniques, along with their advantages and disadvantages. In terms of temporal symmetry, achieving time inversion requires a conservative system without energy loss, such as a frictionless pendulum controlled through RL, as proposed by Agostini and Celaya \cite{agostini2009exploiting}. However, this assumption is impractical for real-world systems, making this approach less desirable. As for spatial symmetry, the contributions fall into the previously established categories:

\subsection{Relabeling states and actions} Zeng and Graham \cite{Zeng21sym} and Ildefonso et al. \cite{ildefonso21} create symmetry-reduced experiences in actor-critic and value-based domains, respectively, by shrinking the state space, running the policy, and finally restoring the actions using data from the original state. Surovik et al. \cite{surovik2019adaptive} combine relabeling states and actions with reflection transformations to swap frame-dependent values, such as lateral rotation, to effectively reduce the state volume of a tensegrity robot. 

However, this same idea can be applied for robot locomotion with simpler models. If, instead of left and right, we think of "stance" and "non-stance" leg for a biped robot, we can relabel them every half cycle to obtain a symmetric controller~\cite{xie2020learning,hereid2018dynamic}. Peng et al.~\cite{peng2017deeploco} apply this relabeling method at fixed intervals of 0.5~s, forcing a fixed period on the gait cycle. This idea is simple to implement but the policy is constrained by some symmetry switch, whether it is based on time or behavioral pattern.

\subsection{Data augmentation} In symmetry-oriented solutions, data augmentation can be used with model-based~\cite{bree2021data} or model-free RL algorithms, although the scope of this work is limited to the latter alternative. Examples of successful applications of this technique include a real dual-armed humanoid robot that moves objects~\cite{lin2020}, the walking gait of several humanoid models~\cite{abdolhosseini2019learning} and a quadruped with more than one plane of symmetry~\cite{mishra2019augmenting}, among others~\cite{agostini2009exploiting,silver2016go}. If changing the loss function is possible, it is always preferable to data augmentation in terms of computational efficiency and memory footprint, despite the additional initial effort. 

\subsection{Symmetric Network} Concerning humanoid models, Abdolhosseini et al. \cite{abdolhosseini2019learning} introduced a symmetric network architecture that forces perfect symmetry. This method guarantees that if states and actions are symmetrically normalized, the behavior has no asymmetric bias. However, as the authors acknowledge, neutral states are inescapable. Pol et al.~\cite{Pol2020mdp} generalize this approach to additional problems by introducing MDP homomorphic networks, which can be automatically constructed by stacking equivariant layers through a numerical algorithm. The equivariance constraints (under a group of reflections or rotations) are applied to the policy and value networks. Encoding invariances in neural networks \cite{encoding22,mondal20,rav2017,san2019,cohen2016} and the group of network solutions in general lacks the ability to control the symmetry enforcement at runtime and the ability to symmetrize asymmetric models.

\subsection{Symmetry Loss function} Mahajan and Tulabandhula~\cite{Mahajan17_abstract,mahajan2017symmetry} proposed an automated symmetry detection process and a method of incorporating that knowledge in the Q-learning algorithm \cite{watkins1992q}. A soft constraint was presented as an additional loss for the Q-function approximator, $\hat{\mathbb{E}}_t[(Q_\theta(f(s_t),g(a_t))-Q_\theta(s,a))^2]$, where $\theta$ is the parameter vector of $Q$, and the expectation $\hat{\mathbb{E}}_t$ indicates the empirical average over a batch of samples. Yu et al. \cite{yu2018learning} transpose this idea to policy gradients by fusing a new curriculum learning method with an homologous loss function --- the Mirror Symmetry Loss (MSL), $\sum_i\|\pi_\theta(s_i)-g(\pi_\theta(f(s_i)))\|^2$. The authors present very good results when both approaches are used simultaneously. 

In previous work, we have presented the Proximal Symmetry Loss (PSL) \cite{psl}, a loss function aimed at increasing the sample efficiency of the Proximal Policy Optimization (PPO) algorithm~\cite{schulman2017ppo} by leveraging static model symmetries. The loss function was built to allow asymmetric exploration in models without symmetry perturbations. It leverages the trust region concept to harmonize its behavior with PPO's exploration, which is especially noticeable in smaller batch sizes. 

In the next section we will delve into the details of MSL and PSL, while suggesting some modifications to expand the functionality of both loss functions. Then, we introduce Adaptive Symmetry Learning, with the ability to handle imperfect symmetry by adapting itself during the learning process.

\section{Symmetry Loss Function} \label{sec:proxSymLoss}

A symmetry loss function computes a value that we seek to minimize in order to obtain a symmetric behavior. Formally, if the robot performs action $a \in \mathcal{A}_s$ in state $s \in \mathcal{S}$, it should also perform the symmetric action $g(a)$ in the symmetric state $f(s)$. Since symmetry is not always the main objective while learning a new behavior, a good symmetry objective function should allow asymmetric exploration in favor of a better policy.

\begin{figure}[!t]
\centering
\includegraphics[scale = 1.3]{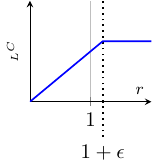} \hspace{1.5cm}
\includegraphics[scale = 1.3]{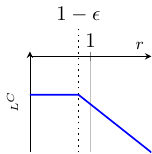}

\caption{Plots for PPO's surrogate objective function $L^{C}$ as a function of ratio $r$, for a single time step, for a positive advantage estimate (on the left) or a negative advantage estimate (on the right).}
\label{fig:ppo_plots}
    
\end{figure}

To understand the effect of the proposed loss function, we must analyze the most common implementation of PPO, where the policy and value networks do not share parameters. The PPO's objective $L^{PPO}$ can then be expressed as:

\begin{align}
    L^{PPO}(\theta, \omega)&=\hat{\mathbb{E}}_t \left[\; L_{t}^{C}(\theta) - L_{t}^{VF}(\omega) + cH(\theta) \;\right], \\
    \text{with} \quad
    L_{t}^{C}(\theta)&=\min\left(r_t(\theta)\hat{A}_t \;,\; (1+\sgn(\hat{A}_t)\epsilon)\hat{A}_t\right), \label{eq:lclip}\\
    \text{and} \quad\;\;\; r_t(\theta)&=\frac{\pi_{\theta}(a_t \mid s_t)}{\pi_{\theta_{old}}(a_t \mid s_t)},\label{eq:ppo_ratio}
\end{align}
\noindent
where the stochastic policy $\pi_\theta$ is parameterized by $\theta$ and the value function by $\omega$. $\pi_{\theta_{old}}$ is a copy of the policy before each update, $\hat{A}_t$ is the estimator of the advantage function, $L_{t}^{VF}$ is a squared error loss to update the value function, $\epsilon$ is a clipping parameter, $c$ is a coefficient and $H$ is the policy's distribution entropy. The expectation $\hat{\mathbb{E}}_t$ indicates the empirical average over a finite batch of samples. The main objective of this algorithm is to keep the policy update within a trust region, preventing greedy updates that can be detrimental to learning. This behavior is formalized in the surrogate objective $L_{t}^{C}$, as depicted in Fig.~\ref{fig:ppo_plots} for a single time step. 

\subsection{Extended loss function}

PPO's objective $L^{PPO}$ can be extended with an arbitrary symmetry loss $L^S$, such that 

\begin{align}
    \label{eq:ppo+s}
    L^{PPO+S}(\theta, \omega)&=L^{PPO}(\theta, \omega) -  L^S(\theta, \omega), 
    \\[3pt]
    \text{with} \quad
    L^S(\theta, \omega) &= \hat{\mathbb{E}}_{t} \left[\,\text{\textit{w}}_\pi \cdot L_t^{\pi}(\theta) + \text{\textit{w}}_V \cdot L_t^{V}(\omega)\,\right], \label{eq:generic_sym_loss}
\end{align}

\noindent where $L_t^{\pi}$ and $L_t^{V}$ are symmetry loss functions that influence the update gradient of the policy (actor) and value function (critic), respectively.

In $L_{t}^{C}$, the ratio $r_t$ starts at 1 and tends to stay close to that value during optimization. The advantage estimate $\hat{A}t$ is z-score normalized in typical PPO implementations \cite{openAIbaselines,stable_baselines}, resulting in a mean of 0. Consequently, $L_{t}^{C}$'s value is usually stable, with a low order of magnitude, which is important to retain when devising extensions, to preserve harmonious interactions among all losses.

\subsection{Generalized Symmetry Loss}

The symmetry loss function can be expanded to an arbitrary number of symmetries with minimal changes. Let $j \in \{1,2,3,...,N\}$ be a symmetry index for a total of $N$ symmetries. Functions $f_j(s)$ and $g_j(a)$ apply symmetry transformation $j$ to state $s$, and $s$-dependent action $a$, respectively. Accordingly, \eqref{eq:generic_sym_loss} can be rewritten as

\begin{alignat}{3}
	L^{S}(\theta,\omega) &= \hat{\mathbb{E}}_{t} \big[&&(\text{\textit{w}}^\pi_1 \cdot L_{1,t}^{\pi}(\theta) + \text{\textit{w}}^V_1 \cdot L_{1,t}^{V}(\omega)) &+ \nonumber
	\\[3pt]
	& &&(\text{\textit{w}}^\pi_2 \cdot L_{2,t}^{\pi}(\theta) + \text{\textit{w}}^V_2 \cdot L_{2,t}^{V}(\omega)) &+ \nonumber
	\\[3pt]
	& &&\hspace{54pt} \cdots &+ \nonumber
	\\[3pt]
	& &&(\text{\textit{w}}^\pi_N \cdot L_{N,t}^{\pi}(\theta) + \text{\textit{w}}^V_N \cdot L_{N,t}^{V}(\omega)) \big] & \nonumber
	\\[3pt]
	&= \sum_{j=1}^N &&\hat{\mathbb{E}}_{t} \big[\text{\textit{w}}^\pi_j \cdot L_{j,t}^{\pi}(\theta) + \text{\textit{w}}^V_j \cdot L_{j,t}^{V}(\omega)\big] &. \label{eq:generic_generalized}
\end{alignat} 

\section{Mirror Symmetry Loss} \label{sec:MSL}

The Mirror Symmetry Loss (MSL) function by Yu et al. \cite{yu2018learning} \mbox{$\text{\textit{w}}_\pi \cdot \sum_i\|\mu_\theta(s_t)-g(\mu_\theta(f(s_t)))\|^2$}, where $\text{\textit{w}}_\pi$ is a weight factor, computes the square error between the mean action taken by the stochastic policy before and after the symmetry transformation. On the one hand, this means that after crossing a certain asymmetry threshold, the loss dominates the policy gradient. On the other hand, if that threshold is crossed to the symmetric side, the symmetry loss loses influence in the policy update. Therefore, $\text{\textit{w}}_\pi$ does not dictate the weight of the symmetry loss in a consistent way, when computing the gradient, but rather the position of the symmetry threshold. Moreover, since the square error is based on actions, instead of probabilities, the weight of the symmetry loss is also dependent on the action space.

One other problem of this equation is the assumption of involutory transformations, a pitfall introduced in Section \ref{sec:sym_policy}. As aforementioned, this issue can be solved by changing where the action transformation is applied. Additionally, the method can be expanded with a value loss, in order to leverage the critic in actor-critic methods. Note that these modifications are a generalization of the original method and do not restrict its previous abilities in any way. The improved mirror symmetry loss function becomes

\begin{align}
    L^{MSL}(\theta,\omega) &= \hat{\mathbb{E}}_{t} \left[\,\text{\textit{w}}_\pi \cdot L_t^{\pi}(\theta) + \text{\textit{w}}_V \cdot L_t^{V}(\omega)\,\right], 
    \\[10pt]
    \text{with} \quad
    L^{\pi}(\theta) &= \hat{\mathbb{E}}_{t}  \big\|\,g(\mu_\theta(s_t))-\mu_\theta(f(s_t))\,\big\|^2, \label{eq:MSL_Lpi}
    \\[3pt]
    L^{V}(\omega) &= \hat{\mathbb{E}}_{t} \left[\,(V_\omega(f(s_t)) - V_t^\text{targ})^2 \,\right], \label{eq:MSL_LV}
\end{align}

\noindent where $L^{\pi}$ and $L^{V}$ are the policy and value losses for symmetry, and $\text{\textit{w}}^\pi$ and $\text{\textit{w}}^V$ are the respective weight factors. The value loss reduces the difference between the value function (for the symmetric state), and the same value target used by PPO.

The generalization in \eqref{eq:generic_generalized} can be applied to MSL by deriving $L_{j,t}^{\pi}$ and $L_{j,t}^{V}$ from \eqref{eq:MSL_Lpi} and \eqref{eq:MSL_LV}, respectively,

\begin{align}
	L_{j,t}^{\pi}(\theta) &= \big\|\,g_j(\mu_\theta(s_t))-\mu_\theta(f_j(s_t))\,\big\|^2,
	\\[3pt]
	L_{j,t}^{V}(\omega) &= (V_\omega(f_j(s_t)) - V_t^\text{targ})^2.
\end{align}
\vspace{0mm}

\section{Proximal Symmetry Loss} \label{sec:PSL}

The Proximal Symmetry Loss (PSL) builds on the trust region concept of PPO to reduce the model's asymmetry iteratively \cite{psl}. In contrast with MSL, PSL can already handle non-involutory operations. However, it was also generalized in this work to include a value loss component, such that

\begin{align}
    L^{PSL}(\theta,\omega) &= \hat{\mathbb{E}}_{t} \left[\,\text{\textit{w}}_\pi \cdot L_t^{\pi}(\theta) + \text{\textit{w}}_V \cdot L_t^{V}(\omega)\,\right], \label{eq:psl}
    \\[10pt]
    \text{with} \quad
    L^{\pi}(\theta) &= -\hat{\mathbb{E}}_{t} \left[\, \min(x_t(\theta),1+\epsilon)\,\right], \label{eq:psl_pi_loss}
    \\[3pt]
    L^{V}(\omega) &= \hat{\mathbb{E}}_{t} \left[\,(V_\omega(f(s_t)) - V_t^\text{targ})^2 \,\right], 
\end{align}

\noindent where $\epsilon$ is a clipping parameter shared with $L^{PPO}$, and $x_t$ is a symmetry probability ratio. 

In contrast to \eqref{eq:lclip}, $L^{\pi}$ does not rely on an advantage estimator to determine the update direction, as its goal is to consistently increase or preserve symmetry. However, the RL algorithm can still opt to decrease symmetry if it benefits the policy, prioritizing $L_{t}^{C}$ over $L_t^S$ in \eqref{eq:ppo+s}. Furthermore, although the symmetry probability ratio shares similarities with \eqref{eq:ppo_ratio}, there are some distinctions:

\begin{equation}
    x_t(\theta)= \frac{\min(\pi_\theta(g(\overline{a}_t) \mid f(s_t)), \pi_{\theta_{old}}(a_t \mid s_t))}{\pi_{\theta_{old}}(g(\overline{a}_t) \mid f(s_t))} , \label{eq:sym_ratio}
\end{equation}

\noindent where $\overline{a}_t = \mu_{\theta_\text{old}}(s_t)$ as defined in \eqref{eq:mean_at}. Parameters $\pi_{\theta}(\cdot \mid s)$ and $\pi_{\theta_{old}}(\cdot \mid s)$ represent probability distributions during, and before the update, respectively, in accordance with the notation provided in Section \ref{sec:policy_notation}. The only term that is not constant during a policy update is $\pi_\theta(g(\overline{a}_t) \mid f(s_t))$, and only $\theta$ is being optimized. Note that $\theta_{old}$ is constant because it represents the policy's parameters before the update.

\subsection{Unidirectional update} \label{sec:unidir_update}

The policy is modified so that the action $\pi_\theta(f(s_t))$ carried out in the symmetric state $f(s_t)$ becomes symmetric to the action taken in the explored state $s_t$. Note that ratio $x_t$ in \eqref{eq:sym_ratio} promotes an adjustment to $\pi_\theta(f(s_t))$ without changing $\pi_\theta(s_t)$. In contrast, the mirror symmetry loss attempts to reduce the distance between $\pi_\theta(f(s_t))$ and $\pi_\theta(s_t)$ by promoting changes to both. This approach is generally adequate, except when there are exploration imbalances in symmetric states. One instance is when a humanoid robot is learning how to navigate a maze, but it initially experiences more left turns than right turns. 

While exploring left turns, it makes sense to adapt the right turning ability to match the symmetric technique, but not vice versa. Adopting the right turning knowledge to change the left turning skill would be detrimental in this case. Yet, when the robot actually explores right turns, it will try to adapt the symmetric action in the wrong direction, but the effect will be diluted by the low probability of experiencing right turns. In summary, adapting the symmetric action is better than the explored action, because the explored action is more likely to be optimal than the symmetric action. In an extreme case, the symmetric state may be unreachable, making the symmetric action unreliable.

\subsection{Generalized Proximal Symmetry Loss}

Similarly to MSL, PSL can also be expanded to an arbitrary number of symmetries by employing \eqref{eq:generic_generalized} and deriving $L_{j,t}^{\pi}$ and $L_{j,t}^{V}$, yielding:

\begin{align}
    L_{j,t}^{\pi}(\theta) &= -  \min(x_{j,t}(\theta),1+\epsilon),
    \\[3pt]
    L_{j,t}^{V}(\omega) &= (V_\omega(f_j(s_t)) - V_t^\text{targ})^2,
    \\[3pt]
    \text{with} \quad
    x_{j,t}(\theta)&= \frac{\min\left(\pi_\theta(g_j(\overline{a}_t) \mid f_j(s_t)), \pi_{\theta_{old}}(a_t \mid s_t)\right)}{\pi_{\theta_{old}}(g_j(\overline{a}_t) \mid f_j(s_t))}. 
\end{align}
\vspace{0mm}

\section{Adaptive Symmetry Learning: Overview} \label{sec:ASLearn}

The previous symmetry loss functions follow the provided symmetry transformations strictly and update the policy accordingly. The main purpose of these loss functions is to find a symmetric outcome, independently of perturbations in the reward function, the agent (control process), the robot or the environment. As an example, consider a humanoid robot with a rusty left arm. The limb still has the same range of motion as the right one, but the amount of torque required to produce the same outcome is different. Assuming that for a given task, no motor needs to reach its maximum torque, there is a symmetry transformation that maps the torque difference between both arms to achieve the equivalent outcome in the symmetric state. For this example, the proposed algorithm finds a function that rectifies biases and achieves a symmetric outcome.

Nevertheless, a symmetric outcome is not always desirable, either due to biases introduced by the environment in favor of asymmetric behaviors, or due to explicit problem constraints expressed through the reward function. To generalize the algorithm to these cases, its purpose can be redefined as adapting symmetry transformation(s) in order to maximize the return. In general terms, this can be achieved by finding functions that best describe the relation between sets of two action elements across all states and symmetry transformations.

\begin{figure}[htbp]
\centering
\includegraphics[trim={20 0 0 0}, width=0.65\columnwidth]{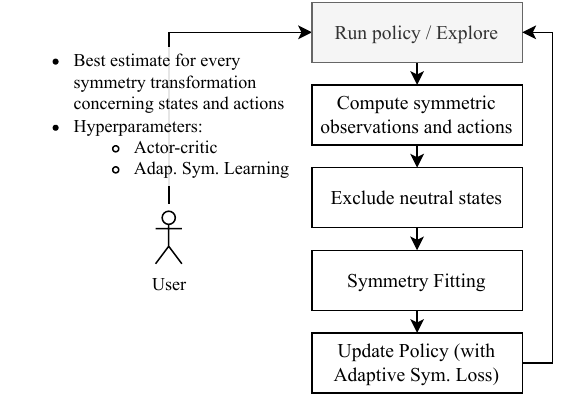}
\caption{Adaptive Symmetry Learning algorithm overview}
\label{fig:learn_algo}
\end{figure}

An ASL overview is shown in Fig.\ref{fig:learn_algo}. Initially, in addition to defining hyperparameters for the actor-critic method (PPO) and the symmetry algorithm, the user has to provide the same information required by previous symmetry loss functions --- the best-known estimate for every symmetry transformation. Each transformation must include a vector of functions to transform states and actions. Symmetry transformations can be omitted if the user deems them to be unrelated or detrimental to the problem being optimized, but this is not required. 

The algorithm starts by running the current policy and exploring the action space to gather a batch of samples (states, actions, rewards). Symmetric observations and actions are computed and stored, and neutral states are excluded based on a user-defined threshold (see Section~\ref{sec:neutral_exclusion} for details on the exclusion method). Then, the provided symmetry transformations suffer a small adjustment in the direction of the policy and, finally, the policy is updated. The final loss used to compute the gradient that updates the policy is a sum of PPO's loss and the Adaptive Symmetry Loss. So, in this last step, the policy suffers a small adjustment in the direction of the symmetry transformation.

Consequently, there is a circular relation between reinforcement learning and symmetry fitting. However, the former learns the behavior (the best action for each state), while the latter learns constant state-independent perturbations from several sources, as enumerated in the beginning of this section. An alternative perspective is to consider the symmetry fitting as a moving average of the symmetry transformations learned by the policy (where the most recent average sample is restricted to the last batch of explored states). A more detailed explanation is given in the following sections.

\section{Adaptive Symmetry Learning: symmetry fitting } \label{sec:sym_learning}

% As described in Section \ref{sec:prelim_MDP_trans}, in a MDP, for a finite set of states $\mathcal{S}$, there is a function $f:\mathcal{S}\rightarrow \mathcal{S}$ that transforms state $s$ into its symetrically equivalent state $f(s)$ for $s\in \mathcal{S}$. Analogously, for a finite set of actions $\mathcal{A}$, there is a function $g_s:\mathcal{A}_s\rightarrow \mathcal{A}_{f(s)}, \forall s \in \mathcal{S}$ that transforms action $a$ into its symetrically equivalent action $g_s(a)$ for $a\in \mathcal{A}_s$, where $\mathcal{A}_s$ is the set of actions allowed in $s$. Since state $s$ is implicit in $g_s(a)$ it will be omitted in following deductions, being the action transformation denoted as $g(a)$ as mentioned in Section \ref{sec:prelim_MDP_trans}.

Each robot can have multiple symmetry elements, i.e., planes or axes about which we apply symmetry transformations. When dealing with imperfectly symmetrical robots, these transformations may not completely preserve the robot's shape and motion properties. Each element may be used to describe one or multiple symmetry transformations (e.g. one axis may be used for several rotation operations). The symmetry elements must be declared as if the robot were perfectly symmetrical, as the objective of ASL is to compensate for imperfections by adapting the transformations. 

Assuming a robot is characterized by a set of symmetry transformations $\mathcal{J}$, function $f_j(s), j \in \mathcal{J}$ maps state $s$ to its symmetrical equivalent under transformation $j$. Function $g(a)$ is formally defined as the solution for equations \eqref{eq:homTran2}  and \eqref{eq:homRew2}, where p and r are characteristics of the environment. Thus, $g_j(a), j \in \mathcal{J}$ is the action that, when applied to state $f_j(s)$, results in the same transition probability to $f_j(s')$, as applying $a$ does from $s$ to $s'$. Additionally, the expected reward in both scenarios is the same.

In typical robotic scenarios, $a$ represents a vector of $n$ scalars that control independent actuators. Consequently, $g_j$ becomes a vector of functions, one for each element of $a$, such that

\begin{equation}
	g_j(a) = \Bigl[G_{j,0}(a[\cdot]), G_{j,1}(a[\cdot]), \dots, G_{j,n}(a[\cdot]) \Bigr], \label{eq:asl_per_scalar}
\end{equation}

\noindent where $G_{j,i}:{\rm I\!R} \rightarrow {\rm I\!R}$ always represents an operation on a scalar value, and $[\cdot]$ indicates an action index which depends on the robot model. We use the notation $a[i]$ instead of $a_i$ to represent elements of an action vector to avoid confusion with the time step $t$ to which the action corresponds. The objective of ASL is to find an estimator $\hat{g}_\nu$ for $g$, or more precisely, an estimator $\hat{G}_{j,i}^\nu$ parameterized by $\nu$ for each symmetry transformation $j$ and each action element $i$. We optimize $\nu$ by extracting these symmetry relationships from the policy for states that were previously explored:

\begin{equation}
	\label{eq:asl_general_fitting}
	\argmin_\nu \sum_{t} \big(\overline{a}'_t [i] - \hat{G}_{j,i}^\nu(\overline{a}_t[i])\big)^2.
\end{equation}

However, there are several ways of improving the symmetry fitting algorithm which will be discussed in the remainder of this section.

\subsection{Overview}

\begin{figure}[htbp]
	\centering
	\includegraphics[width=0.56\columnwidth]{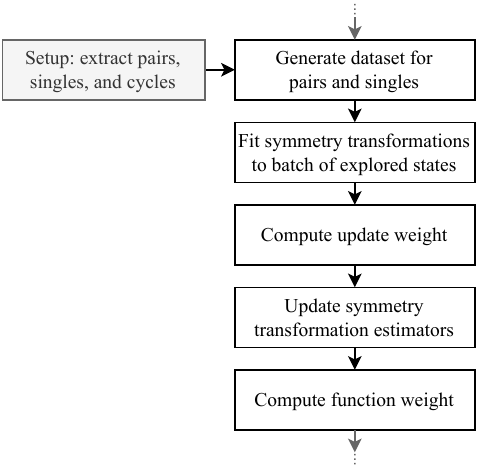}
	\caption{Overview of the symmetry fitting process}
	\label{fig:learn_process}
\end{figure}

An overview of the symmetry fitting process is shown in Fig.~\ref{fig:learn_process}. Before learning, the algorithm extracts and groups types of mappings between action elements (from the provided symmetry transformations). These groups describe dependencies across different planes of symmetry or rotation axes, which can be leveraged to reduce the fitting problem's number of dimensions, improving the solution's efficiency and accuracy (see Section \ref{sec:fitting_setup}).

Using humans as an example, with respect to the sagittal plane, each two distinct body parts that mirror each other form symmetric pairs, while central body parts like the head form reflexive relations (singles). Circular symmetry relations of 3 or more body parts (cycles) do not exist in humans but can exist in certain robots. 

The fitting process is executed after the exploration phase, when local symmetry transformations are fitted to the policy. The term \textit{local} indicates that only the last batch of states from the exploration phase is used to generate the dataset (see Section \ref{sec:pairs} and \ref{sec:singles}). Local transformations are used as targets for the global transformation estimators, according to an update weight. This weight is based on cycle restrictions, if they exist, otherwise a fixed weight is used, as defined by the user (see Section \ref{sec:cycles}).

Finally, the targets used to update each function in the transformation vectors are assessed in terms of stability to compute a function weight employed by the symmetry loss to penalize unstable functions (see Section \ref{sec:f_weight}). These topics will be reviewed in detail through an example robot model. 

\begin{figure}[htbp]
\centering
\includegraphics[trim={2mm 0 4mm 0},clip,width=0.72\columnwidth]{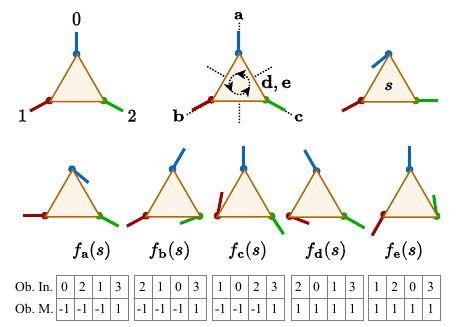}
\caption{2D robot shaped as an equilateral triangle with 3 limbs, 3 symmetry planes (\textbf{a}, \textbf{b}, \textbf{c}) and 3-fold rotational symmetry (\textbf{d}, \textbf{e}). $f_\textbf{a}(s)$, $f_\textbf{b}(s)$, $f_\textbf{c}(s)$, $f_\textbf{d}(s)$ and $f_\textbf{e}(s)$ are symmetry transformations of $s$ relative to $\textbf{a}$, $\textbf{b}$, $\textbf{c}$, $\textbf{d}$ and $\textbf{e}$, respectively. The observations indices and multipliers characterize the transformations above.}
\label{fig:sym_example}
\end{figure}

Inspired by a triangular lamina example by McWeeny \cite{McWeeny}, consider a 2D robot shaped as an equilateral triangle with 3 controllable limbs that are able to rotate around axes located at each vertex of the robot (see Fig.~\ref{fig:sym_example}, top left). For the symmetry of a figure to be absolutely known, all its non-equivalent symmetry properties must be determined \cite{lectures1917}. Assuming the robot cannot be flipped, it has 6 non-equivalent symmetry operations characterized by 3 symmetry planes (\textbf{a}, \textbf{b}, \textbf{c}) and 3-fold rotational symmetry (\textbf{d}, \textbf{e}), i.e., it is invariant under rotations of 120, 240 or 360 deg, although the last one is an identity operation. Therefore, in total, only 5 symmetry operations are to be considered (see Fig.~\ref{fig:sym_example}, top middle).

Rotating 240 deg is equivalent to performing two 120 deg rotations. This means that, although both operations must be declared, in theory, only the specification of 120 deg should be needed. Yet, 
currently, ASL does not allow the specification of combined symmetry operations, which means that 5 specifications are required. Despite this limitation, the algorithm will automatically simplify redundant operations, as explained at the end of the next section.

\subsection{Setup} \label{sec:fitting_setup}

The state space of the robot is represented by 4 variables $[X_0,X_1,X_2,Z]$, where $X_i$ is the angular position of limb $i$ relative to its neutral position (let the positive rotation direction be counterclockwise), and $Z$ is the battery level. The purpose of $Z$ is to introduce a variable that is not affected by symmetry transformations. In the top right corner of Fig.~\ref{fig:sym_example} is represented an arbitrary state $s$, and in the bottom are represented the states that result from applying the 5 aforementioned symmetry transformations (e.g. $f_\textbf{a}(s)$ is the reflection of $s$ in relation to symmetry plane $\textbf{a}$). 

Below $f_\textbf{a}(s)$ to $f_\textbf{e}(s)$ are two rows of values that characterize the respective transformation. The top row specifies how to rearrange the variables of $s$ (observation indices) and the bottom row indicates if an inversion is necessary (observation multipliers). Transformation $f_\textbf{a}$, for instance, is a vector of functions such that  
\begin{equation}
    \label{eq:f_a}
    f_\textbf{a}(s)=[F_{\textbf{a},0}(s[0]),\; F_{\textbf{a},1}(s[2]),\; F_{\textbf{a},2}(s[1]),\; F_{\textbf{a},3}(s[3])],  \quad F_{\textbf{a},i}:{\rm I\!R} \rightarrow {\rm I\!R} .
\end{equation}

Given the observation multipliers in Fig.~\ref{fig:sym_example}, it is possible to simplify the equation to
\begin{equation}
    \label{eq:f_a_2}
    f_\textbf{a}(s)=[-s[0], -s[2], -s[1], s[3] ].
\end{equation}

The new value for $X_0$ is given by $X_0\cdot-1$ (same limb but inverted rotation direction), but limb 1 gets the inverted value of limb 2 and vice versa. Since the battery level is not affected by this transformation, $X_3 \leftarrow X_3\cdot 1$. As observed, the rotation direction was inverted for every limb, and the same is true for $f_\textbf{b}(s)$ and $f_\textbf{c}(s)$, but not for $f_\textbf{d}(s)$ and $f_\textbf{e}(s)$, which require no inversion.

\begin{figure}[htbp]
\centering
\includegraphics[trim={0 0 13 0},clip,width=0.75\columnwidth]{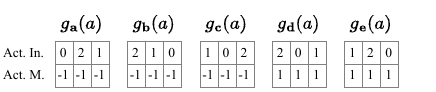}
\caption{Action indices and multipliers that characterize the 2D robot's symmetry transformations $g_\textbf{a}(a)$, $g_\textbf{b}(a)$, $g_\textbf{c}(a)$, $g_\textbf{d}(a)$ and $g_\textbf{e}(a)$}
\label{fig:sym_example_act_tr}
\end{figure}

The robot has 3 action variables $[Y_0,Y_1,Y_2]$, where $Y_i$ is the torque applied to limb $i$. In this case, given action $a$, the symmetry transformations $g_\textbf{a}(a)$, $g_\textbf{b}(a)$, $g_\textbf{c}(a)$, $g_\textbf{d}(a)$ and $g_\textbf{e}(a)$ are analogous to the already introduced transformations, excluding the battery level (see the respective actions indices and multipliers in Fig.~\ref{fig:sym_example_act_tr}). 

\begin{figure}[htbp]
\centering
\includegraphics[trim={0 0 13 0}, clip,width=0.75\columnwidth]{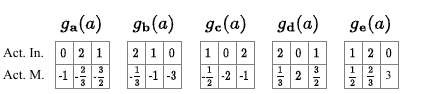}
\caption{Hypothetical action indices and multipliers in a scenario where limb 1 is rusty and requires double the normal torque and limb 2 requires triple the normal torque. This transformation holds while both affected motors are below saturation.}
\label{fig:sym_example_asym}
\end{figure}

As mentioned in the beginning of Section \ref{sec:ASL_loss_function}, in some circumstances, it is possible to achieve a symmetric outcome with perturbations in the control process, the robot or the environment. Therefore, it is possible to leverage the same symmetric principles if the motors of the 2D robot behave differently due to several factors (e.g. different power specifications, unbalanced control drivers, or external factors such as rust). As a basic example, consider that limb 1 is rusty and requires double the normal torque to perform the same movement, and limb 2 requires triple the normal torque. Until either motor reaches its maximum torque, the symmetry transformations can be described as in Fig.~\ref{fig:sym_example_asym}.

In some situations, the symmetry transformations that lead to the highest return may not be known a priori. In Fig.~\ref{fig:sym_example_asym}, the transformation operation can be defined as a linear equation of the form $y=mx$, where $x$ is an arbitrary action and $m$ is the action multiplier vector. On the one hand, different problems may require more complex functions to describe the required operation. On the other hand, the policy (typically parameterized by a neural network) is already able to represent complex nonlinear relations. So, if a linear function can approximate the symmetric relation to a satisfactory degree, it should be preferred, as complex functions have increased learning time, potentially counteracting the sample efficiency benefit of using symmetries.

\begin{figure}[htbp]
\centering
\includegraphics[trim={2 6 13 0}, clip,width=0.75\columnwidth]{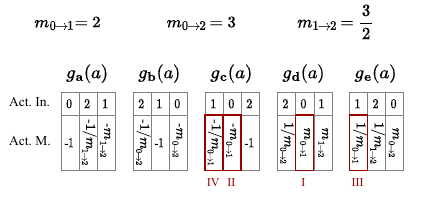}
\caption{Rewriting the action multipliers from Fig.~\ref{fig:sym_example_asym} as a function of three pair relations of the form $a'[y]=m_{x\rightarrow y}\cdot a[x]$. Highlighted in red are action multipliers that depend on $m_{0\rightarrow 1}$.}
\label{fig:sym_example_pairs}
\end{figure}

During its initialization, the symmetry fitting process extracts pairs of symmetric action elements from the user-defined action indices. In the current scenario, there are 3 pairs \{(0,1),(0,2),(1,2)\}, which means that the 15-dimensional operation represented in Fig.~\ref{fig:sym_example_act_tr} and Fig.~\ref{fig:sym_example_asym} can be reduced to a 3-dimensional problem. The relations between each pair of action elements can be written in the form $a'[y]=m_{x\rightarrow y}\cdot a[x]$. Fig.~\ref{fig:sym_example_pairs} shows how the action multipliers in Fig.~\ref{fig:sym_example_asym} could be reduced to functions of the three proportional relations between pairs $m_{0\rightarrow 1}$, $m_{0\rightarrow 2}$ and $m_{1\rightarrow 2}$. 

\subsection{Generate dataset for pairs} \label{sec:pairs}

After running the policy and collecting a batch of explored states, the symmetry fitting procedure can be executed. The first step is to generate a dataset to fit the user-defined function. This step is repeated for every pair of action elements, but for the sake of conciseness, only pair (0,1) will be analyzed. The objective is to fit an estimator $\hat{G}^\zeta_{0\rightarrow 1}:{\rm I\!R} \rightarrow {\rm I\!R}$, parameterized by $\zeta$, to the local transformation, such that $a'_t[1] = G_{0\rightarrow 1}(a_t[0])$ for $t \in \{0,1,...,B\}$, where $B$ is the batch size. The direction of such relation is $0\rightarrow 1$ because it transforms the action associated with limb 0 to that of limb 1.

The dataset does not depend on actions sampled during exploration. Instead, in line with previously introduced symmetry loss functions, it uses the distribution mean to reduce noise. It is limited to states that were explored because the process of fitting symmetry transformations relies on what the policy has learned. Including rare or unreachable states would be counterproductive since the policy could produce unreliable results.

The dataset is generated as seen in Table~\ref{tab:dataset}, with one column for input values and another one for output values. There are four variations of the same relation in symmetry transformations $g_\textbf{c}(a)$, $g_\textbf{d}(a)$ and $g_\textbf{e}(a)$. Relation (I) in $g_\textbf{d}(a)$ (see Fig.~\ref{fig:sym_example_pairs}) follows the desired direction $0\rightarrow 1$ (because under $g_\textbf{d}(a)$, $m_{0\rightarrow 1}$ is the value we multiply by $a[0]$ to obtain $a'[1]$). So, the inputs are directly obtained from the distribution mean for every state $s_t$ in the latest batch ($\overline{a}_0[0]$), and the outputs are obtained from the distribution mean for all $f_\textbf{d}(s_t)$ ($\overline{a}'_{\mathbf{d},t}[1]$).

\begin{table}[ht]
\caption{Extraction of dataset from policy. A batch of explored states of size $B$, such that time step $t \in \{0,1,...,B\}$, is fed into the policy to generate a dataset of actions. This table shows the dataset used to learn symmetry transformations for action pair (0,1).}
\begin{center}
\begin{tabular}{|c|c|c|}
\hline
\textbf{Local relation} & \textbf{Input} ($x$) & \textbf{Output} ($y$) \\ \hline
\multirow{3}{*}{\begin{tabular}[c]{@{}c@{}}$0\rightarrow 1$\\ ( \textcolor[rgb]{0.6,0,0}{I} )\end{tabular}} & $\overline{a}_0[0]$ & $\overline{a}'_{\mathbf{d},0}[1]$ \\ \cline{2-3} 
 & ... & ... \\ \cline{2-3} 
 & $\overline{a}_B[0]$ & $\overline{a}'_{\mathbf{d},B}[1]$ \\ \hline
\multirow{3}{*}{\begin{tabular}[c]{@{}c@{}}$0\rightarrow 1$, reflection\\ ( \textcolor[rgb]{0.6,0,0}{II} )\end{tabular}} & $-\overline{a}_0[0]$ & $\overline{a}'_{\mathbf{c},0}[1]$ \\ \cline{2-3} 
 & ... & ... \\ \cline{2-3} 
 & $-\overline{a}_B[0]$ & $\overline{a}'_{\mathbf{c},B}[1]$ \\ \hline
\multirow{3}{*}{\begin{tabular}[c]{@{}c@{}}$1\rightarrow 0$\\ ( \textcolor[rgb]{0.6,0,0}{III} )\end{tabular}} & $\overline{a}'_{\mathbf{e},0}[0]$ & $\overline{a}_0[1]$ \\ \cline{2-3} 
 & ... & ... \\ \cline{2-3} 
 & $\overline{a}'_{\mathbf{e},B}[0]$ & $\overline{a}_B[1]$ \\ \hline
\multirow{3}{*}{\begin{tabular}[c]{@{}c@{}}$1\rightarrow 0$, reflection\\ ( \textcolor[rgb]{0.6,0,0}{IV} )\end{tabular}} & $-\overline{a}'_{\mathbf{c},0}[0]$ & $\overline{a}_0[1]$ \\ \cline{2-3} 
 & ... & ... \\ \cline{2-3} 
 & $-\overline{a}'_{\mathbf{c},B}[0]$ & $\overline{a}_B[1]$ \\ \hline
\end{tabular}
\label{tab:dataset}
\end{center}
\end{table}

For relation (II), symmetry plane \textbf{c} requires limb 1 to mirror limb 0 (inverting the rotation direction), yielding $a'_t[1] = G_{0\rightarrow 1}(-a_t[0])$. Note the minus sign, which is applied in Table~\ref{tab:dataset} for all inputs of the second local relation.

Relation (III) is similar to relation (I) except that the direction is $1\rightarrow 0$ and thus, the inputs and outputs are swapped. Finally, relation (IV) requires an inversion of inputs and outputs and a sign reversal for the inputs due to the local reflection.

\subsection{Generate dataset for singles} \label{sec:singles}

Symmetry transformations $g_\textbf{a}(a)$, $g_\textbf{b}(a)$ and $g_\textbf{c}(a)$ also contain reflexive relations (singles), i.e., relations from and to the same action element. In Fig.~\ref{fig:sym_example_pairs}, singles have an action multiplier of $-1$, which means that in the symmetric state, the symmetric action consists in rotating the same limb but in the opposite direction. One limitation of reflexive relations is that they must be characterized by involutory functions to avoid the complexity of state dependency. This notion is better understood with a practical example.

\begin{figure}[htbp]
\centering
\includegraphics[trim={0 2 17 9},clip,width=0.74\columnwidth]{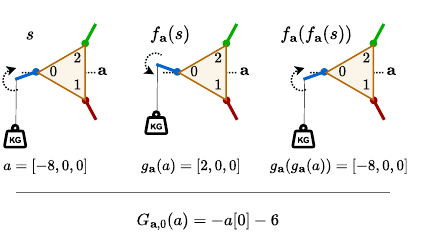}
\caption{Practical use case for symmetry fitting, where the symmetric action transformation can be approximated by a linear involutory function, that does not depend on the environment state}
\label{fig:sym_example_self1}
\end{figure}

Fig.~\ref{fig:sym_example_self1} denotes a practical use case for symmetry fitting, where limb 0 of the previously introduced 2D robot is subjected to a constant force pointing to the weight. Note that the weight is just a representation of the perturbation, which can have multiple sources, including the environment or robot defects. 

On the left side of Fig.~\ref{fig:sym_example_self1} is represented state $s$. Assume that limb 0 has a relatively short range of motion and the perturbation manifests as a constant torque of +3 being applied to limb 0 (positive torque means clockwise direction). Let $a[0]=-8$ be the optimal action in $s$ for limb 0 and $a[0]=2$ the optimal action in the symmetric state $f_\textbf{a}(s)$. Function $G_{\textbf{a},0}(a)=-a[0]-6$ is a linear approximation of the symmetry transformation for the corresponding action element. This function is symmetric across the line $y=x$ and, consequently, involutory. This means that for any symmetry plane and any state $G_{\textbf{a},0}(G_{\textbf{a},0}(a))=a[0]$.

\begin{figure}[htbp]
\centering
\includegraphics[trim={0 2 20 7},clip,width=0.74\columnwidth]{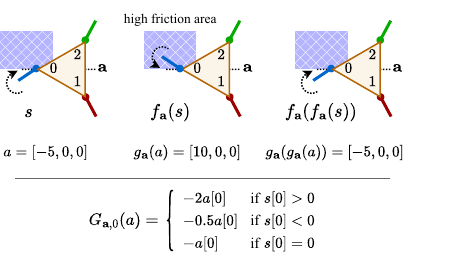}
\caption{Practical use case for symmetry fitting, where the symmetric action transformation is dependent on the environment state because it is not characterized by an involutory function}
\label{fig:sym_example_self2}
\end{figure}

A linear functions of the form $y=mx+b$ is involutory if $m=-1$ or $m=1 \wedge b=0$. Fig.~\ref{fig:sym_example_self2} presents another practical use case for symmetry fitting, where the symmetric action transformation depends on the position of limb 0 ($s[0]$). The diagonal grid in Fig.~\ref{fig:sym_example_self2} is a high friction area relative to the robot's body, where limb 0 must apply twice as much torque to overcome the extra friction. Therefore, when the limb is below symmetry plane \textbf{a}, or more formally, the angular position of limb 0 is positive ($s[0]>0$), the symmetry transformation is given by $G_{\textbf{a},0}(a)=-2a[0]$. The rotation direction is inverted and the torque is multiplied by 2 to counteract the friction and obtain a symmetric outcome. However, when $s[0]<0$, the inverse function is needed $G_{\textbf{a},0}(a)=-0.5a[0]$. 

Since $y=-2x$ is not involutory, a 3D piecewise function is required, as defined in the bottom of Fig.~\ref{fig:sym_example_self2}. Recall that $g$ is a function of state-action pairs $(s,a)$, even though the implicit state $s$ is omitted, as mentioned in Section \ref{sec:prelim_MDP_trans}. Considering a problem where the ground truth is unknown (e.g. a high friction area of unknown location or shape), fitting a symmetry transformation would require the algorithm to learn the number of sub-functions, the parameters of each sub-function and the associated subdomains. The possible ramifications of this problem are outside the scope of this paper. 

Therefore, to avoid the complexity of state dependency, symmetry transformations for singles are limited to involutions. The dataset for singles is a simplified version of Table~\ref{tab:dataset} since reflexive relations have a unique direction (e.g. $0\rightarrow0$). Moreover, linear transformations are a special case that reduces the number of local relations to one. As aforementioned, involutory linear functions of the form $y=mx+b$ are constrained by $m=1 \wedge b=0$ or $m=-1$. The former constraint has no learnable parameter, so it is not included in the dataset. The latter defines a negative relation and can be included in the dataset to learn parameter $b$. Since there are no positive learnable relations, there is no need to perform reflections about the vertical axis (reversing the sign of inputs), thus reducing the effective number of local relations to one. 

\subsection{Optimization} \label{sec:optimization}

A possible solution for this problem is to use the method of least squares to fit an arbitrary local estimator $\hat{G}_\zeta$ parameterized by $\zeta$ to the dataset
\begin{equation}
    \label{eq:minimize_F}
    \argmin_\zeta \sum_{i=1}^U \big(y_i - \hat{G}_\zeta(x_i)\big)^2,
\end{equation}

\noindent where $U$ is the number of inputs and outputs extracted from the policy. If the local estimator is a linear equation $\hat{G}_\zeta=\hat{m}x+\hat{b}$, the method formulation can be reduced to ordinary least squares, to which there is a constant-time analytical solution
\begin{align}
    \label{eq:minimize_mxb}
    \hat{m}, \hat{b} &= \argmin_{m, b} \sum_{i=1}^U \big(y_i - m x_i - b \big)^2,
    \\[3pt]
    \hat{m} &= \frac{U (\sum_i x_iy_i) - (\sum_i x_i)(\sum_i y_i)}{U (\sum_i x_i^2)-(\sum_i x_i)^2},
    \\[3pt]
    \hat{b} &= \frac{(\sum_i y_i) - m(\sum_i x_i)}{U}.
\end{align}

For singles, only $\hat{b}$ can be estimated, simplifying the solution

\begin{align}
    \label{eq:minimize_b}
    \hat{b} &= \argmin_{b} \sum_{i=1}^U \big(y_i + x_i - b\big)^2 \nonumber
    \\[3pt]
      &= \frac{(\sum_i x_i) + (\sum_i y_i)}{U}.
\end{align}

For pairs, having a bias component $b$ can be useful in certain scenarios, as will be discussed in Section \ref{sec:results}. However, for the sake of completeness, fixing $b$ reduces \eqref{eq:minimize_mxb} to
\begin{align}
    \label{eq:minimize_mx}
    \hat{m} &= \argmin_{m} \sum_{i=1}^U \big(y_i - m x_i \big)^2 \nonumber
    \\[3pt]
      &= \frac{(\sum_i x_iy_i)}{(\sum_i x_i^2)}.
\end{align}

\subsubsection{Cycles} \label{sec:cycles}

The described optimization problem can have additional restrictions if there are circular relations of 3 or more pairs (hereinafter referred to as cycles). In the 2D robot example, the pairs of symmetric action elements \{(0,1),(0,2),(1,2)\} form a single cycle (0 to 1 to 2), although it can also be represented as (2 to 1 to 0) since all relation pairs are bidirectional. One important property of cycles is that the composition of all functions in a cycle is an identity transformation $\text{id}:{\rm I\!R} \rightarrow {\rm I\!R}$, such that $\text{id}(k)=k, \forall\, k \in \rm I\!R$, e.g., $\hat{G}^\zeta_{0\rightarrow 1}\circ \hat{G}^\zeta_{1\rightarrow 2} \circ \hat{G}^\zeta_{2\rightarrow 0} = \text{id}$. Plugging this equality constraint into \eqref{eq:minimize_F}, and generalizing to an example with $C$ cycles and a set of pairs $\mathcal{P}$ yields 

\begin{align}
    \label{eq:minimize_F_cycle}
    \argmin_\zeta &\sum_{p}^\mathcal{P} \sum_{i=1}^{U_p} \big(y_i - \hat{G}^\zeta_{p}(x_i)\big)^2,
    \\[3pt]
    \text{subject to} \; &(\hat{G}^\zeta_{T_0^0}\circ \hat{G}^\zeta_{T_1^0} \circ \hat{G}^\zeta_{T_2^0} [\circ \; \dots])(k) = k, \nonumber
    \\[3pt]
    &\dots \nonumber
    \\[3pt]
    &(\hat{G}^\zeta_{T_0^C}\circ \hat{G}^\zeta_{T_1^C} \circ \hat{G}^\zeta_{T_2^C} [\circ \; \dots])(k) = k, \nonumber
\end{align}

\noindent where $T_a^b$ is a jagged array indicating the $a$-th pair of cycle $b$. Using cycle-based constraints increases the optimization complexity due to the constraint incorporation and the fact that pair relations contained in cycles can no longer be computed independently of each other. Moreover, there are typically numerous local minima that obey the constraints, making the optimization more sensitive to small differences in the dataset. In practice, during training, this technique introduced a substantial amount of variance in the estimator $\hat{G}_\zeta$. 

An alternative approach is to optimize each transformation independently using \eqref{eq:minimize_F}, and then compute an update weight based on the cycle error $\epsilon$. For pair $(q,u)$, the update weight $\text{\textit{w}}_{U_{q\leftrightarrow u}}$ can be computed through a user-defined function $\text{H}_\text{U}:{\rm I\!R}^+_0 \rightarrow [0,1]$, such that

\begin{align}
    \label{eq:cycle_error}
    \text{\textit{w}}_{U_{q\leftrightarrow u}} &= \min(\text{H}_\text{U}(\epsilon_0), \dots, \text{H}_\text{U}(\epsilon_C)),
    \\[3pt]
    \text{with} \quad \epsilon_i &= \left|(\hat{G}^\zeta_{T_0^i}\circ \hat{G}^\zeta_{T_1^i} \circ \hat{G}^\zeta_{T_2^i} [\circ \; \dots])(k) - k\right|, \nonumber
\end{align}

\noindent where $C$ is the number of cycles that includes pair $(q,u)$. An example for function $\text{H}_\text{U}$ is $\text{H}_\text{U}(x)=0.05 \times 0.01/(0.01 + x^4)$, where $0.05$ is the maximum update weight, and $0.01/(0.01 + x^4)$ is a decaying factor dependent on the cycle error. 

\subsubsection{Transformation Update} \label{sec:transf_update}

During the optimization stage, the algorithm finds the best parameter vector $\zeta$ for each local estimator function $\hat{G}_\zeta$, considering only the last batch of explored states. To reduce variance, this estimator is used as a target for the global estimator $\hat{G}_\nu$, introduced in the beginning of Section \ref{sec:sym_learning}. Using the weight $\text{\textit{w}}_U$ computed in the previous section, the update rule for each pair (q,u) is defined as an exponential moving average

\begin{equation}
	\label{eq:local_update}
	\nu_{q\rightarrow u} \leftarrow \nu_{q\rightarrow u} + \text{\textit{w}}_{U_{q\leftrightarrow u}}  (\zeta_{q\rightarrow u} - \nu_{q\rightarrow u}).
\end{equation}

For singles, the update rule is similar, but instead of $\text{\textit{w}}_U$, we use $\text{max}(\text{H}_\text{U})$, which is the maximum update weight mentioned at the end of the previous section. Assuming relations of the form $y=mx+b$, each independent update $i$ becomes
\begin{align}
    \label{eq:pair_update}
    \nu_{m_i} &\leftarrow \nu_{m_i} + \text{\textit{w}}_{U}  (\zeta_{m_i} - \nu_{m_i}),
    \\[3pt]
    \nu_{b_i} &\leftarrow \,\nu_{b_i} \,+ \text{\textit{w}}_{U}  (\zeta_{b_i} \;- \nu_{b_i}\;).
\end{align}

To summarize, there are three notations for transformation estimators:

\begin{enumerate}
	\item $\hat{G}_{j,i}^\nu$ is a global estimator parameterized by $\nu$  for every symmetry transformation $j$ and every action element $i$. As shown in Fig.~\ref{fig:sym_example}, the example robot model has 5 symmetry transformations ($j \in \{\textbf{a},\textbf{b},\textbf{c},\textbf{d},\textbf{e}\}$) and an action vector with 3 elements, resulting in 15 estimators. This notation is important when extracting data from the policy to create training datasets and later in Section \ref{sec:ASL_loss_function} when using the learned estimators to influence the policy.
	\item $\hat{G}^\zeta_q$ and $\hat{G}^\zeta_{q \rightarrow u}$ are two notations for a local estimator parameterized by $\zeta$ that is fitted from scratch after every policy exploration phase. We only need to learn 6 independent estimator (instead of 15), including 3 pairs \{$\hat{G}^\zeta_{0 \rightarrow 1}$,$\hat{G}^\zeta_{0 \rightarrow 2}$,$\hat{G}^\zeta_{1 \rightarrow 2}$\} and 3 reflexive relations or singles \{$\hat{G}^\zeta_0$,$\hat{G}^\zeta_1$,$\hat{G}^\zeta_2$\}.
	\item $\hat{G}^\nu_q$ and $\hat{G}^\nu_{q \rightarrow u}$ denote the exponential moving average of $\hat{G}_\zeta$. $\hat{G}^\nu_q$ and $\hat{G}^\nu_{q \rightarrow u}$ represent the independent variables of $\hat{G}_{j,i}^\nu$, i.e., each of the 15 global estimators represented by $\hat{G}_{j,i}^\nu$ can be written as a combination of 6 independent estimators \{$\hat{G}^\nu_{0 \rightarrow 1}$,$\hat{G}^\nu_{0 \rightarrow 2}$,$\hat{G}^\nu_{1 \rightarrow 2}$,$\hat{G}^\nu_0$,$\hat{G}^\nu_1$,$\hat{G}^\nu_2$\} as shown in Table~\ref{tab:global_vs_local}.
\end{enumerate}

\begin{table}[ht]
	\caption{Global symmetry transformation estimators can be written as combinations of independent estimators. This table shows the equivalences for the example robot from Fig.~\ref{fig:sym_example}. The third column is a specialization of the second column where every independent estimator is represented by a linear function.}
	\begin{center}
		\def\arraystretch{1.3}%  1 is the default, change whatever you need
		\small
		\begin{tabular}{|ll|l|l|}
			\hline
			\multicolumn{2}{|l|}{\textbf{Global estimator}} & \textbf{Indep. estimator} & \textbf{Indep. estimator (if linear)} \\ \hline
			\multirow{3}{*}{$\hat{g}_\textbf{a}^\nu(a)\;=\!\!\!$} & [\,$\hat{G}_{\textbf{a},0}^\nu(a[0])$, & $\hat{G}_{0}^\nu(a[0])$ & $\nu_{m_0}\cdot a[0]+\nu_{b_0}$ \\ \cline{3-4} 
			 & \phantom{[}\,$\hat{G}_{\textbf{a},1}^\nu(a[2])$, & $\hat{G}_{1}^\nu(\hat{G}_{1\rightarrow 2}^\nu(a[2])^{-1})$ & $\nu_{m_1}((a[2]-\nu_{b_{1\rightarrow 2}}) / \nu_{m_{1\rightarrow 2}})+\nu_{b_1}$ \\ \cline{3-4} 
			 & \phantom{[}\,$\hat{G}_{\textbf{a},2}^\nu(a[1])$\,] & $\hat{G}_{2}^\nu(\hat{G}_{1\rightarrow 2}^\nu(a[1]))$ &  $\nu_{m_2}(\nu_{m_{1\rightarrow 2}}\cdot a[1]+\nu_{b_{1\rightarrow 2}})+\nu_{b_2}$ \\ \hline
			\multirow{3}{*}{$\hat{g}_\textbf{b}^\nu(a)\;=\!\!\!$} & [\,$\hat{G}_{\textbf{b},0}^\nu(a[2])$, & $\hat{G}_{0}^\nu(\hat{G}_{0\rightarrow 2}^\nu(a[2])^{-1})$ & $\nu_{m_0}((a[2]-\nu_{b_{0\rightarrow 2}}) / \nu_{m_{0\rightarrow 2}})+\nu_{b_0}$ \\ \cline{3-4} 
			& \phantom{[}\,$\hat{G}_{\textbf{b},1}^\nu(a[1])$, & $\hat{G}_{1}^\nu(a[1])$ &  $\nu_{m_1}\cdot a[1]+\nu_{b_1}$ \\ \cline{3-4} 
			& \phantom{[}\,$\hat{G}_{\textbf{b},2}^\nu(a[0])$\,] & $\hat{G}_{2}^\nu(\hat{G}_{0\rightarrow 2}^\nu(a[0]))$ & $\nu_{m_2}(\nu_{m_{0\rightarrow 2}}\cdot a[0]+\nu_{b_{0\rightarrow 2}})+\nu_{b_2}$  \\ \hline
			\multirow{3}{*}{$\hat{g}_\textbf{c}^\nu(a)\;=\!\!\!$} & [\,$\hat{G}_{\textbf{c},0}^\nu(a[1])$, & $\hat{G}_{0}^\nu(\hat{G}_{0\rightarrow 1}^\nu(a[1])^{-1})$ & $\nu_{m_0}((a[1]-\nu_{b_{0\rightarrow 1}}) / \nu_{m_{0\rightarrow 1}})+\nu_{b_0}$ \\ \cline{3-4} 
			& \phantom{[}\,$\hat{G}_{\textbf{c},1}^\nu(a[0])$, & $\hat{G}_{1}^\nu(\hat{G}_{0\rightarrow 1}^\nu(a[0]))$ & $\nu_{m_1}(\nu_{m_{0\rightarrow 1}}\cdot a[0]+\nu_{b_{0\rightarrow 1}})+\nu_{b_1}$  \\ \cline{3-4} 
			& \phantom{[}\,$\hat{G}_{\textbf{c},2}^\nu(a[2])$\,] & $\hat{G}_{2}^\nu(a[2])$ & $\nu_{m_02}\cdot a[2]+\nu_{b_2}$ \\ \hline
			\multirow{3}{*}{$\hat{g}_\textbf{d}^\nu(a)\;=\!\!\!$} & [\,$\hat{G}_{\textbf{d},0}^\nu(a[2])$, & $\hat{G}_{0\rightarrow 2}^\nu(a[2])^{-1}$ & $(a[2]-\nu_{b_{0\rightarrow 2}}) / \nu_{m_{0\rightarrow 2}}$ \\ \cline{3-4} 
			& \phantom{[}\,$\hat{G}_{\textbf{d},1}^\nu(a[0])$, & $\hat{G}_{0\rightarrow 1}^\nu(a[0])$ & $\nu_{m_{0\rightarrow 1}}\cdot a[0]+\nu_{b_{0\rightarrow 1}}$ \\ \cline{3-4} 
			& \phantom{[}\,$\hat{G}_{\textbf{d},2}^\nu(a[1])$\,] & $\hat{G}_{1\rightarrow 2}^\nu(a[1])$ & $\nu_{m_{1\rightarrow 2}}\cdot a[1]+\nu_{b_{1\rightarrow 2}}$ \\ \hline
			\multirow{3}{*}{$\hat{g}_\textbf{e}^\nu(a)\;=\!\!\!$} & [\,$\hat{G}_{\textbf{e},0}^\nu(a[1])$, & $\hat{G}_{0\rightarrow 1}^\nu(a[1])^{-1}$ & $(a[1]-\nu_{b_{0\rightarrow 1}}) / \nu_{m_{0\rightarrow 1}}$ \\ \cline{3-4} 
			& \phantom{[}\,$\hat{G}_{\textbf{e},1}^\nu(a[2])$, & $\hat{G}_{1\rightarrow 2}^\nu(a[2])^{-1}$ & $(a[2]-\nu_{b_{1\rightarrow 2}}) / \nu_{m_{1\rightarrow 2}}$  \\ \cline{3-4} 
			& \phantom{[}\,$\hat{G}_{\textbf{e},2}^\nu(a[0])$\,] & $\hat{G}_{0\rightarrow 2}^\nu(a[0])$ & $\nu_{m_{0\rightarrow 2}}\cdot a[0]+\nu_{b_{0\rightarrow 2}}$ \\ \hline
		\end{tabular}
		\label{tab:global_vs_local}
	\end{center}
\end{table}

\subsection{Function weight} \label{sec:f_weight}

For the 2D robot example, there are 5 symmetric action transformations, each representing a vector of 3 functions (1 per limb). At an early optimization stage, the policy is mostly random and, consequently, the symmetry fitting targets tend to have high standard deviation $\sigma$. As the convergence starts, $\sigma$ is progressively reduced, albeit at different rates for distinct functions. Furthermore, a constant high $\sigma$ (in successive batches of explored states) may indicate that a given function is not a good estimator for the symmetry transformation. For these reasons, when targets have high $\sigma$, the respective function estimator has less importance in the symmetry loss.

For a given estimator function parameterized by vector \textbf{x}, let $\mu_{x_i}$ and $\sigma_{x_i}$ be the mean and standard deviation of the target value for parameter ${x_i}$, respectively, in the last $k_\text{I}$ iterations, where $k_\text{I}$ is a hyperparameter. The metric used to assess the recent target spread is the absolute value of the coefficient of variation, based on the definition of Everitt and Skrondal \cite{statsdict}, $c_v=\sigma/|\mu|$. This metric is important to reduce the relative standard deviation as the mean increases. However, to avoid instability near $\mu=0$, a small constant (0.1) is added to the function weight $\text{\textit{w}}_G$ equation

\begin{equation}
    \label{eq:func_weight}
    \text{\textit{w}}_G = \text{H}_G(\frac{\sigma}{|\mu|+0.1}),
\end{equation}

\noindent where $\text{H}_G:{\rm I\!R}^+_0 \rightarrow [0,1]$ is a user-defined function that transforms the spread metric into a weight between 0 and 1. The function weight $\text{\textit{w}}_G$ is used later by the symmetry loss in \eqref{eq:ASL_impl_r_with_func_w}.

\section{Adaptive Symmetry Learning: loss function} \label{sec:ASL_loss_function}

The Adaptive Symmetry Learning loss function computes a novel loss that tackles problems such as neutral states or disadvantageous updates. Additionally, symmetry transformation estimators are used according to their reliability, as defined above by $\text{\textit{w}}_G$. The algorithm is characterized by the following equation:

\begin{align}
    L^{ASL}(\theta,\omega) &= \hat{\mathbb{E}}_{t} \left[\text{\textit{w}}_\pi \cdot L_t^{\pi}(\theta) + \text{\textit{w}}_V \cdot L_t^{V}(\omega)\right], \label{eq:ASL}
    \\[10pt]
    \text{with} \quad
    L^{\pi}(\theta) &= -\hat{\mathbb{E}}_{t} \left[  r_t(\theta) \cdot \psi_t \cdot \phi_t \right], \label{eq:ASL_pi_loss}
    \\[3pt]
    L^{V}(\omega) &= \hat{\mathbb{E}}_{t} \left[(V_\omega(f(s_t)) - V_t^\text{targ})^2 \cdot \psi_t \cdot \phi_t \right], \label{eq:ASL_v_loss}
\end{align}

\noindent where $L^{\pi}$ and $L^{V}$ are the policy and value losses for symmetry, and $\text{\textit{w}}^\pi$ and $\text{\textit{w}}^V$ are the respective weight factors. The value loss reduces the difference between the value function (for the symmetric state), and the same value target used by PPO. The policy loss promotes a policy symmetrization with a restricted update. This behavior is partly achieved through a dead zone gate $\psi$ and a value gate $\phi$, both based on user-defined thresholds. The former rejects neutral states or those that are near neutrality (see Section~\ref{sec:neutral_exclusion}), and the latter avoids changing the symmetric action if it is more valuable than the current action (see Section~\ref{sec:value_gate}). 

However, the main progression restriction is imposed by the ratio $r_t$, which works similarly to PPO's ratio \eqref{eq:ppo_ratio}, except that the restriction is applied on the target action $\tau_t$ instead of the ratio itself, and the policy's standard deviation $\sigma$ is not actively modified:

\begin{align}
    r_t(\theta) &= \frac{\pi_{\theta}(\tau_t \mid f(s_t), \mu_\theta, \sigma )\;}{\pi_{\theta_\text{old}}(\tau_t \mid f(s_t), \mu_{\theta_\text{old}}, \sigma  )}, \label{eq:ASL_r}
    \\[3pt]
    \tau_t &= \text{clip}(\hat{g}(\mu_{\theta_\text{last}}(s_t)), \overline{a}'_t-\Delta\mu,\overline{a}'_t+\Delta\mu), \label{eq:ASL_t}
\end{align}

\noindent where $\overline{a}'_t$ \eqref{eq:mean_a't} is the mean of the policy distribution for state $f(s_t)$ when $a_t$ was sampled; $\mu_{\theta_\text{last}}(s_t)$ is the mean of the distribution after the last mini-batch update; $\hat{g}$ is a symmetry transformation estimator optimized during the symmetry fitting stage; and $\Delta\mu$ is a user-defined restriction threshold explained in the following subsection. The estimator $\hat{g}$ omits the parameterization variable $\nu$ mentioned in Section~\ref{sec:sym_learning} because it is frozen in the loss function.

The mean $\mu_{\theta_\text{last}}(s_t)$ represents the most recent action for state $s_t$. It is similar to $\mu_{\theta_\text{old}}(s_t)$ in the sense that it is a non-trainable constant, but it is updated during the optimization stage, at every mini-batch update of every train epoch. 
In \eqref{eq:ASL_t}, $\mu_{\theta_\text{last}}(s_t)$ could be replaced by the explored action $a_t$ in the cases where $a_t$ has a positive advantage estimate ($\hat{A}_t \geq 0$), since PPO would try to increase $a_t$'s relative probability. However, doing that would subvert PPO's trust region by allowing the symmetry update to exploit the sampled action in a greedy way. Moreover, the objective of the symmetry update is not to explore, but rather to motivate symmetry in a nondestructive way. 

As an example, consider a bidirectional variable-speed conveyor belt that is trying to center an item, and the state indicates whether the item is on the left or the right side. Assume that initially the neural network outputs a speed of -5 for both cases, leading every item to eventually fall to the left. After some exploration, the policy finds that for an item on the right, a speed of $a_t\sim \pi_\theta(s_t)=-2$ is advantageous (as it increases the time the item spends near the center). Let $\Delta\mu=0.5$. Before the update, the target $\tau_t$ is $\text{clip}(5, -5-0.5,-5+0.5)=-4.5$. Ideally, after every mini-batch update, $\mu_{\theta_\text{last}}(s_t)$ gets closer to -2 due to PPO, leading to a target of $\text{clip}(2, -5-0.5,-5+0.5)=-4.5$. In this scenario the target does not change because the action for the symmetric state is too far from $g(\mu_{\theta_\text{last}}(s_t))$. Large symmetry updates could harm the policy by changing the neural network too much. 

%in the explanation above the reason why the target does not change is because the "-5" is not supposed to change during mini-batch updates bc it uses theta old, and that's why the action "2" being too far from g(mu(st)) doesnt change the target.

If the target is exceeded (e.g. $\mu_{\theta}(f(s_t))=-4.4$), the gradient of $r_t(\theta)$ will try to move the policy back to the target. In an analogous situation, PPO would only stop motivating further progress towards its sampled target. This harsh symmetry restriction has the purpose of breaking momentum build-ups, as explained in the next subsection. Now assume that after hours of training, the NN outputs -10 and 10 for an item on the right and left sides, respectively. Yet, it finds that 11 is advantageous for the left side item. The target $\tau_t$ becomes $\text{clip}(-10, -10-0.5,-10+0.5)=-10$, which produces no difference. However, as $\mu_{\theta}(s_t)$ increases to 10.1 along several mini-batch updates, the target decreases to $\tau_t=-10.1$. So, the symmetric action closely follows the sampled action until it reaches -10.5. If the symmetric limitation is small enough, PPO's loss function will dominate the update direction, allowing for asymmetric updates, if required.

The next subsection delves into the motivation and design of the restricted update, and one of its most critical parameters --- the distribution's mean shift $\Delta\mu$. It also provides context for the standard deviation preservation in \eqref{eq:ASL_r}, and the momentum build up issues. The following subsections derive the dead zone and value gates, $\psi$ and $\phi$ from \eqref{eq:ASL_pi_loss}; present a condensed implementation of the loss function and summarize all ASL's hyperparameters.

\subsection{Restricted update} \label{sec:restricted_upd}

Both PPO and TRPO \cite{trpo} apply the trust region concept as depicted in Fig.~\ref{fig:ppo_plots}. During policy updates, the target relative probability of a certain action is restricted by a soft limit imposed by the clipping parameter $\epsilon$. As an example, consider an explored action with a positive estimated advantage ($\hat{A}_t>0$), and $\epsilon=0.3$. The loss function gradient will motivate a relative probability growth of 30\%, although the new relative probability can surpass that soft limit due to the influence of other actions in the update batch, or the optimizers used to update network weights.

As the policy follows a normal distribution, there are 2 parameters that can influence the relative probability for a given action: the mean $\mu_\theta$ and the standard deviation $\sigma$. The mean is parameterized by a neural network and altering it means shifting the preferred action for a given state, while the standard deviation expands or narrows the exploration around said action. For symmetry, there are three key differences in relation to the original concept of trust region:

\begin{enumerate}
    \item For a given time step $t$ there is always exactly one desired action, so the objective is to always increase the relative probability of its symmetric counterpart (except to enforce the trust region as explained in point 3);
    \item Since the update direction is known a priori, there is no need for exploration, so it could be risky for the symmetry loss to modify the policy's standard deviation $\sigma$, as it could negatively affect the main algorithm. The Proximal Symmetry Loss's term $\pi_{\theta_{old}}(a_t \mid s_t)$ in \eqref{eq:sym_ratio} attempts to mitigate this issue by limiting the relative probability of the symmetric action. However, ideally, the symmetry loss should not actively modify $\sigma$;
    \item While using common momentum-based optimizers such as Adaptive Moment Estimation (Adam) \cite{kingma2014adam}, the trust region applied to constant targets is inherently hard to control, as the constant update direction can build up momentum and easily override asymmetric exploration. There are at least three ways of overcoming this issue: not using momentum-based optimizers, which comes at a heavy cost; reduce the symmetry weight so that the noise created by the total loss is enough to prevent cyclic patterns (as applied by the PSL); or create a fixed target at the trust region border, which actively reduces the relative probability of the symmetric action if it exceeds the region, instead of simply not motivating further progress, as does PPO. This last solution is used by ASL in \eqref{eq:ASL_r} and \eqref{eq:ASL_t}.
\end{enumerate}

\begin{figure}[!ht]
\centering
\includegraphics[scale = 1.12]{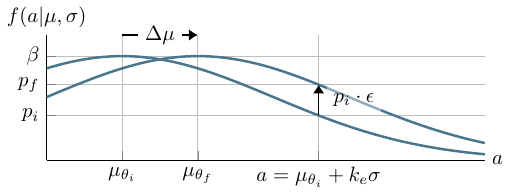}

\caption{Relation between: (1) distance of an advantageous explored action ($a_t$) to the distribution mean before the update ($\mu_{\theta_i}$) and (2) the progress of the distribution. Assume an ideal update with the trust region and a policy that follows a univariate normal distribution with fixed standard deviation. }
\label{fig:exp_delta}
    
\end{figure}

It is important to control the influence of symmetry in exploration and harmonize the final loss function. To understand the best approach, we need to analyze the average PPO update for a single time step, considering the above-mentioned restrictions. Fig. \ref{fig:exp_delta} shows the relation between exploration and target policy update for a single time step. Action $a=\mu_{\theta_i} + k_\text{e}\sigma$ has a positive estimated advantage and follows a univariate normal distribution with fixed standard deviation. Exploration parameter $k_\text{e}$ defines the distance of the sampled action to the initial mean of the distribution $\mu_{\theta_i}$ in relation to $\sigma$. The gradient of PPO's loss function motivates the initial relative probability $p_i$ of action $a_t$ to increase until $p_i\cdot\epsilon$. Assuming an ideal update, the neural network shifts the distribution's mean to $\mu_{\theta_f}$. The relation between $k_\text{e}$ and the distribution mean shift $\Delta\mu$ will allow us to replicate that behavior for the symmetry update. 

However, this example deals with univariate distributions, and typical reinforcement learning environments have high-dimensional action spaces. In PPO, the clipping parameter $\epsilon$ limits the relative probability of the action as a whole. Considering the notion of joint probability for independent events, each action element has an average soft limit 

\begin{equation}
    \label{eq:xi}
    \xi=\sqrt[n]{1+\epsilon}-1,
\end{equation}

\noindent where $n$ is the action size. Although the PPO loss function does not require this dimension-level specification, the explored action uses an independent standard deviation value for each element, so it makes sense to also establish individual targets for each element. 

\begin{figure}[!ht]
    \centering
    \includegraphics[scale = 1.08]{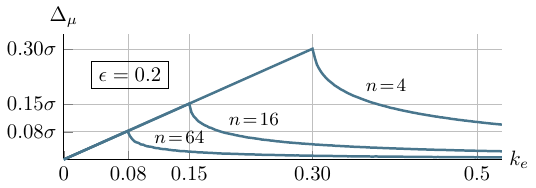}
    \caption{Plot of the distribution's mean shift $\Delta\mu$ as a function of $k_\text{e}$ with $\xi=\sqrt[n]{1+0.2}-1$ and $n\in\{4,16,64\}$}
    \label{fig:exp_curve}
\end{figure}

The following derivation considers the relation between $k_\text{e}$ and $\Delta\mu$ for a single element. Let the probability density function (PDF) for a univariate normal distribution be $f(x|\mu,\sigma)=\beta e^{-\frac{1}{2}(\frac{x-\mu}{\sigma})^2}, \; \text{with} \quad \beta = 1/(\sigma\sqrt{2\pi})$. The distribution mean shift $\Delta\mu$ can be obtained as the distance from $a_t$ to the largest $a$ for which the PDF of ${\mathcal {N}}(\mu_{\theta_i} ,\sigma)$ is $p_f$:

\begin{align}
    p_i&=f(\mu_{\theta_i}+\text{e}\sigma\,|\,\mu_{\theta_i},\sigma)=\beta e^{-\frac{1}{2}k_\text{e}^2},
    \\[3pt]
    p_f&= \min((1+\xi)p_i, \beta), 
    \\[3pt]
    \Delta\mu &= \mu_{\theta_i} + k_\text{e} \sigma - f^{-1}(p_f|\mu_{\theta_i},\sigma)  \nonumber
    \\[3pt]
     &= \cancel{\mu_{\theta_i}} + k_\text{e} \sigma - \cancel{\mu_{\theta_i}} - \sigma \sqrt{-2\cdot\ln \frac{\min((1+\xi)\cancel{\beta} e^{-\frac{1}{2}k_\text{e}^2},\cancel{\beta})}{\cancel{\beta}} }  \nonumber
    \\[3pt]
     &= \sigma \left(k_\text{e} - \sqrt{-2\cdot\ln \min((1+\xi)e^{-\frac{1}{2}k_\text{e}^2},1)}\right), \label{eq:deltamu}
\end{align}

\noindent where $f^{-1}(x|\mu,\sigma)=\mu + \sigma \sqrt{-2\ln(x/\beta)}$ is the largest solution for the inverse PDF. Fig.~\ref{fig:exp_curve} shows the plot of $\Delta\mu$ as a function of $k_\text{e}$, for a typical clipping value $\epsilon=0.2$ \cite{schulman2017ppo} and 3 action space dimensions $n\in\{4,16,64\}$. For each function represented in Fig.~\ref{fig:exp_curve}, the maximum at $k_\text{e}=\sqrt{-2\ln(1/(1+\xi))}$, separates the linear part on the left, from the nonlinear part on the right (that corresponds to Fig.~\ref{fig:exp_delta}). The linear case happens when $\mu_{\theta_f}=a_t$.

From an empirical standpoint, restricting the symmetry update such that $\Delta\mu\geq\sigma\sqrt{-2\ln(1/(1+\xi))}$ serves no purpose other than to accelerate the symmetry convergence at the beginning of training. Therefore, for the sake of simplicity, the update restriction can be reduced to 

\begin{equation}
    \label{eq:delta_mu_2}
    \Delta\mu=k_\text{s}\sigma\sqrt{-2\ln\frac{1}{1+\xi}}.
\end{equation}

\begin{figure}[!ht]
    \centering
    \includegraphics[scale = 1.1]{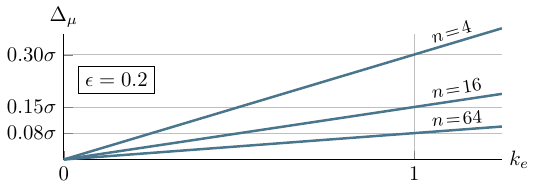}
    \caption{Plot of the distribution's mean shift $\Delta\mu$ as a function of $k_\text{s}$ with $\xi=\sqrt[n]{1+0.2}-1$ and $n\in\{4,16,64\}$}
    \label{fig:exp_curve_2}
\end{figure}

In both cases, the restriction is given in relation to $\sigma$ and $\xi$. However, instead of defining the distance of the artificially explored symmetric action through $k_\text{e}$, and then compute the maximum distribution shift, we are defining that shift directly through $k_\text{s}$. Consequently, the stationary point from Fig.~\ref{fig:exp_curve} is removed and $\Delta\mu(k_\text{s}|\sigma,\xi)$ becomes a linear function, as shown in Fig.~\ref{fig:exp_curve_2}. 

The unit value for $k_\text{e}$ and $k_\text{s}$ also has a different meaning. Choosing $k_\text{e}=1$ means adopting PPO's update behavior for an action sampled at 1 standard deviation from the distribution's mean ($a=\mu_{\theta_i}\pm\sigma$), assuming a constant $\sigma$. Alternatively, defining $k_\text{s}=1$ means adopting PPO's update behavior for the action that would yield the largest update, also assuming a constant $\sigma$. Moreover, defining $k_\text{s}>1$ has an intuitive meaning, since the update magnitude increases monotonically.

\subsection{Exclusion of neutral states} \label{sec:neutral_exclusion}

As introduced in Section~\ref{sec:prelim_neutral}, neutral states and neighboring areas in the state space pose a significant challenge for learning efficient behaviors through symmetric policies. ASL provides the option to reject regions that surround neutral states through a dead zone gate $\psi_t$:

\begin{align}
    &
    \psi_t = 
    \begin{cases}
        1,  & \text{if } \frac{1}{n} \sum_{i=1}^n \big(\delta_{t,i}\big)   > k_\text{d} \\
        0,  & \text{otherwise}
    \end{cases},
    \\[3pt]
    &\text{with}\quad
    \delta_{t,i} = \frac{|s_{t,i} - f(s_{t,i})|}{\text{MAD}_i},
    \\[3pt]
    &\text{and}\quad
    \text{MAD}_i = \frac{\sum_{t=1}^{k_\text{t}} |s_{t,i} - \overline{s}_{i}|}{k_\text{t}},
\end{align}

\noindent where $\text{MAD}_i$ is the mean absolute deviation of  state $s$ for state element $i$, for the last $k_\text{t}$ time steps (typically $k_\text{t}=10\cdot B$, where $B$ is the batch size). Parameter $\delta_{t,i}$ represents the relative distance between state $s$ and its symmetric counterpart $f(s)$, for a specific time step $t$ and state element $i$.

The dead zone gate $\psi_t$ is 0 if $s_t$ is relatively close to a neutral state, and 1 otherwise. When $\psi_t=0$ in \eqref{eq:ASL_pi_loss} and \eqref{eq:ASL_v_loss}, the gradients of $L^{ASL}$ w.r.t. $\theta$ and $\omega$ ignore time step $t$. In practice, the policy is free to learn asymmetric behaviors near neutral states, thus circumventing the problem stated in Section \ref{sec:prelim_neutral}. 

\subsection{Value gate} \label{sec:value_gate}

Analogously to PSL's unidirectional update (see Section \ref{sec:unidir_update}), ratio $r_t$ in \eqref{eq:ASL_r} promotes an adjustment to $\pi_\theta(f(s_t))$ without changing $\pi_\theta(s_t)$. As already concluded, adapting the symmetric action is better than the explored action because the explored action is more likely to be optimal than the symmetric action.

This approach can be improved by comparing the estimated value of the explored and symmetric actions, and then blocking updates when the symmetric action is considerably more valuable. To achieve that purpose, a value gate $\phi_t$ yields 1 to allow an update, or 0 to block it, based on a user defined parameter $k_\text{v} \in (1,+\infty)$, such that

\begin{align}
    \phi_t &= 
    \begin{cases}
        1,  & \text{if } v_t > V_\omega(f(s_t)) \\
        0,  & \text{otherwise}
    \end{cases}, \label{eq:value_gate}
    \\[3pt]
    \text{with}\quad
    v_t &= 
    \begin{cases}
        k_\text{v} \cdot V_\omega(s_t),  & \text{if } V_\omega(s_t)\geq 0 \\
        \frac{V_\omega(s_t)}{k_\text{v}} ,  & \text{otherwise}
    \end{cases}. \label{eq:value_gate_vt}
\end{align}

\begin{figure}[!ht]
    \centering
    \includegraphics[scale = 1.2]{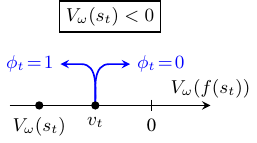}
    \includegraphics[scale = 1.2]{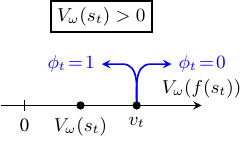}
    \caption{ Value gate $\phi_t$ behavior when the explored state value $V_\omega(s_t)$ is negative (left) and positive (right). The difference lies in how the auxiliary variable $v_t$ is obtained.}
    \label{fig:value_gate}
\end{figure}

Although setting $k_\text{v}=1$ would obviate the need for \eqref{eq:value_gate_vt}, simplifying the condition in \eqref{eq:value_gate} to $V_\omega(s_t) > V_\omega(f(s_t))$, it revealed to be counterproductive in early tests, since many time steps were being blocked due to inaccuracies in the value function estimator. Therefore, to reduce the gate sensitivity, it is recommended to use values for $k_\text{v}$ well above 1 (e.g. 1.5). Since $k_\text{v}$ is a proportional factor, it is required to distinguish when $V_\omega(s_t)$ is positive and negative, as illustrated in Fig.~\ref{fig:value_gate}. 

In both cases, the update is blocked ($\phi_t=0$) when the value of the symmetric state is larger than $v_t$, being the only difference the way $v_t$ is calculated. Using a proportional factor $v_t$ instead of adding a constant is important to deal with outlier states with unusually small or large values.

\subsection{Implementation} \label{sec:ASL_impl}

This subsection presents a condensed implementation of ASL's loss function, compiling the final form of all previously explained modules. Rewriting \eqref{eq:ASL} as an empirical average over a finite batch of $B$ samples for $N$ symmetries yields:

\begin{align}
    \label{eq:ASL_impl}
    L^{ASL}(\theta,\omega) &= \sum_{j=1}^N \frac{1}{B}\sum_{t=1}^B \left[\text{\textit{w}}^\pi_j \cdot L_{j,t}^{\pi}(\theta) + \text{\textit{w}}^V_j \cdot L_{j,t}^{V}(\omega)\right],
    \\%[3pt]
    \text{with} \quad
    L_{j,t}^{\pi}(\theta) &= -r_{j,t}(\theta) \cdot \psi_{j,t} \cdot \phi_{j,t},
    \\[3pt]
    L_{j,t}^{V}(\omega) &= (V_\omega(f_j(s_t)) - V_t^\text{targ})^2 \cdot \psi_{j,t} \cdot \phi_{j,t}.
\end{align}

The ratio $r_{j,t}$ between two probability densities can be simplified, considering that both deal with the same standard deviation variable $\sigma$, and by applying the quotient rule for exponents with the same base:

\begin{align}
    \label{eq:ASL_impl_r}
    r_{j,t}(\theta) &= \frac{\pi_{\theta}(\tau_{j,t} \mid f_j(s_t), \mu_\theta, \sigma )\;}{\pi_{\theta_\text{old}}(\tau_{j,t} \mid f_j(s_t), \mu_{\theta_{old}}, \sigma  )} \nonumber
    \\[3pt]
    &= \frac{\cancel{\frac{1}{(2\pi)^{n/2}\prod_i\sigma_i}}
    \cdot\exp\left(-\frac{1}{2}\sum_{i=1}^n
    \frac{\left(\tau_{j,t,i}-{\mu_{\theta}}_i(f_j(s_t))\right)^2}{\sigma_i^2}\right) }
    {\cancel{\frac{1}{(2\pi)^{n/2}\prod_i\sigma_i}}
    \cdot\exp\left(-\frac{1}{2}\sum_{i=1}^n
    \frac{\left(\tau_{j,t,i}-\overline{a}'_{j,t,i}(f_j(s_t))\right)^2}{\sigma_i^2}\right) } \nonumber
    \\[3pt]
    &= \exp\left(\sum_{i=1}^n\frac{(\tau_{j,t,i} -\overline{a}'_{j,t,i})^2-(\tau_{j,t,i} -{\mu_{\theta}}_i(f_j(s_t)))^2}{2 \cdot\sigma_i^2}\right),
\end{align}

\noindent where $\overline{a}'_{j,t} = \mu_{\theta_\text{old}}(f_j(s_t))$ and $n$ is the number of elements of each action. Plugging in the function weight $\text{\textit{w}}_G$ defined in \eqref{eq:func_weight} for symmetry fitting:

\begin{equation}
    \label{eq:ASL_impl_r_with_func_w}
    r_{j,t}(\theta) = \exp\left(\sum_{i=1}^n\frac{(\tau_{j,t,i} -\overline{a}'_{j,t,i})^2-(\tau_{j,t,i} -{\mu_{\theta}}_i(f_j(s_t)))^2}{2 \cdot\sigma_i^2 / \text{\textit{w}}^G_{j,i}}\right).
\end{equation}

Target $\tau_t$, distribution mean shift $\Delta\mu$ and dead zone gate $\psi_t$ can be rewritten for a specific symmetry index $j$ with minimum changes:

\begin{align}
    \tau_{j,t} &= \text{clip}(\hat{g}_j(\mu_{\theta_\text{last}}(s_t)), \overline{a}'_{j,t}-\Delta\mu_j,\overline{a}'_{j,t}+\Delta\mu_j), \label{eq:ASL_impl_t}
    \\[3pt]
    \Delta\mu_j&=k_j^\text{s}\sigma\sqrt{-2\ln\frac{1}{1+\xi}}, \label{eq:ASL_impl_delta_mu}
    \\[3pt]
    \psi_{j,t} &= 
    \begin{cases}
        1,  & \text{if } \frac{1}{n} \sum_{i=1}^n \big(\delta_{j,t,i}\big)  > k_j^\text{d} \\
        0,  & \text{otherwise}
    \end{cases},
    \\[3pt]
    &\text{with}\quad
    \delta_{j,t,i} = \frac{|s_{t,i} - f_j(s_{t,i})|}{\text{MAD}_i},
    \\[3pt]
    &\text{and}\quad
    \text{MAD}_i = \frac{\sum_{t=1}^{k_\text{t}} |s_{t,i} - \overline{s}_{i}|}{k_\text{t}},
\end{align}

\noindent where $\xi$ is defined in \eqref{eq:xi}. Value gate $\phi_t$ can also be easily adapted to a specific symmetry index $j$ but the auxiliary variable $v_t$ can be improved for parallel computation by replacing the branch with an equivalent equation:

\begin{align}
    \phi_{j,t} &= 
    \begin{cases}
        1,  & \text{if } v_t > V_\omega(f_j(s_t)) \\
        0,  & \text{otherwise}
    \end{cases}, 
    \\[3pt]
    \text{with}\quad
    v_t &= \alpha \cdot V_\omega(s_t) + (k_\text{v}-\alpha)|V_\omega(s_t)|,
\end{align}

\noindent where $\alpha$ is an auxiliary constant equal to $({k_\text{v}}^2+1)/(2k_\text{v})$.

\subsection{Hyperparameters}

The hyperparameters described in previous subsections are summarized in Table \ref{tab:asl_hyperparameters}. Most parameters are vectors with one scalar per symmetry operation. The exceptions are system-wide parameters that include $\boldsymbol{\hat{g}}(a)$ form, $\text{H}_\text{U}$, $\text{H}_\text{G}$ and $k_\text{t}$.

\begin{table}[ht]
\caption{ASL Hyperparameters}
\begin{center}
\begin{tabular}{|r|l|c|}
\hline
\textbf{Symbol} & \textbf{Description} & \textbf{Section} \\ \hline
$\boldsymbol{f}(s)$ & symmetric state transformation & \ref{sec:prelim_MDP_trans}  \\ \hline
$\text{\textit{\textbf{w}}}_\pi$ & policy symmetry loss weight & \ref{sec:ASLearn}  \\ \hline
$\text{\textit{\textbf{w}}}_\text{V}$ & value symmetry loss weight & \ref{sec:ASLearn}  \\ \hline
$\boldsymbol{\hat{g}}(a)$ & sym. action transformation estimator & \ref{sec:sym_learning}  \\ \hline
$\boldsymbol{\hat{g}}(a)$ form & function form of $\boldsymbol{\hat{g}}(a)$ & \ref{sec:sym_learning} \\ \hline
$\text{H}_\text{U}$ & cycle penalty function & \ref{sec:cycles}  \\ \hline
$\text{H}_\text{G}$ & action transformation penalty function & \ref{sec:f_weight}  \\ \hline
$\boldsymbol{k}_\text{s}$ & maximum distribution shift & \ref{sec:restricted_upd} \\ \hline
$\boldsymbol{k}_\text{d}$ & dead zone around neutral state & \ref{sec:neutral_exclusion} \\ \hline
$k_\text{t}$ & time steps for neutral state rejection & \ref{sec:neutral_exclusion}  \\ \hline
$\boldsymbol{k}_\text{v}$ & value gate proportional factor & \ref{sec:value_gate}  \\ \hline
\end{tabular}
\label{tab:asl_hyperparameters}
\end{center}
\end{table}

\section{Evaluation Environment}\label{sec:eval_env}

The experiments for this paper were executed in PyBullet --- a physics simulator based on the Bullet Physics SDK \cite{coumans2020}. An ant robot based on the AntBulletEnv-v0 environment was chosen as the base to develop several learning scenarios. This robot model was selected due to the type and number of symmetry transformations under which it is invariant.  

\begin{figure}[!h]
	\centering
	\begin{minipage}{0.52\textwidth} % Adjust width as needed
		\centering
		\includegraphics[width=\columnwidth]{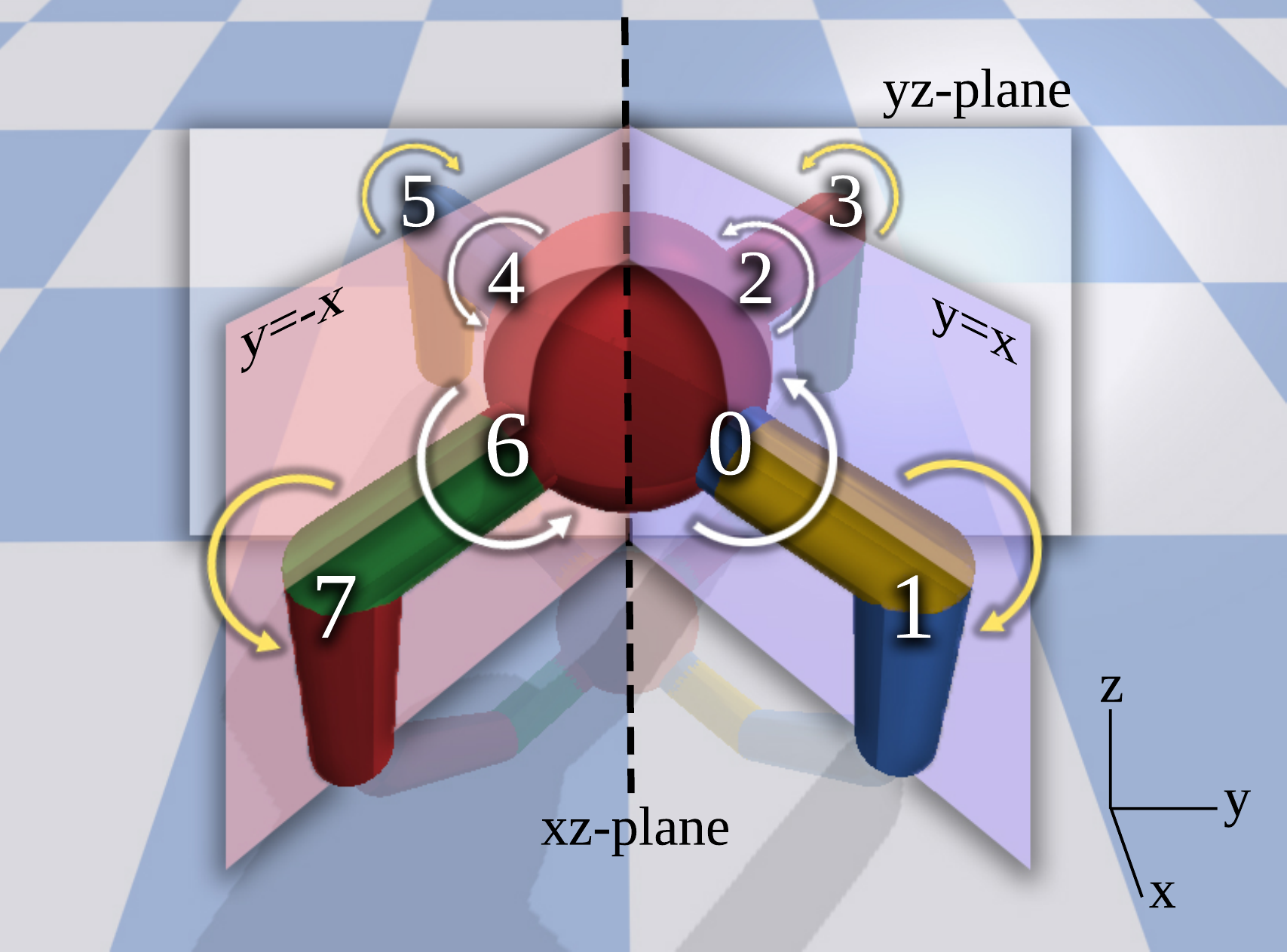}
		\caption{AntBulletEnv-v0's robot has 8 joints/actuators. Even joints rotate around the robot's z-axis, and odd joints (ant's knees) rotate around a vector orthogonal to the z-axis. Arrows show the positive rotation direction for each joint.  The robot has 4 planes of symmetry, xy-plane, yz-plane, y=x and y=-x, and 4-fold rotational symmetry around the z-axis.}
		\label{fig:ant}
	\end{minipage}
	\hfill
	\begin{minipage}{0.44\textwidth}
		\centering
		\small
		\captionof{table}{AntBulletEnv-v0's state space}
		\begin{center}
			\begin{tabular}{|l|l|}
				\hline
				\textbf{Index} & \textbf{Description} \\ \hline
				0 & height - initial height \\ \hline
				1-2 & unit goal vector (y,x) \\ \hline
				3-5 & linear velocity (x,y,z) \\ \hline
				6-7 & roll (x), pitch (y) \\ \hline
				8-23 & position/speed of joints 0-7 \\ \hline
				24-27 & feet contact (boolean) \\ \hline
			\end{tabular}
			\label{tab:state_space}
		\end{center}
	\end{minipage}
\end{figure}

The ant robot, shown in Fig.~\ref{fig:ant}, is a four-legged robot with 8 joints and actuators. Joint 0 controls the rotation of one leg around the robot's z-axis, and joint 1 controls the knee rotation for the same leg. The remaining joints control the other legs in an equivalent way. The robot has 4 planes of symmetry: xy-plane, yz-plane, y=x and y=-x. Additionally, it has 4-fold rotational symmetry around the z-axis. More specifically, the robot is invariant under rotations of 90, 180, 270 or 360 deg around the z-axis, although the last rotation is an identity operation.

The state space for all ant-based scenarios is listed in Table~\ref{tab:state_space}. The first observation indicates the relative height of the robot in comparison to the height at the beginning of an episode, which is fixed. The second observation is a unit vector parallel to the ground that indicates the goal direction, relative to the robot. After the relative linear velocity of the robot, roll and pitch, comes the vector $[p_0,s_0,p_1,s_1,...,p_7,s_7]$, where $p_x$ and $s_x$ are the position and speed of joint $x$. Finally, a binary value indicates whether each foot is touching the ground.

\begin{figure}[!t]
	\centering
\includegraphics[width=0.95\columnwidth]{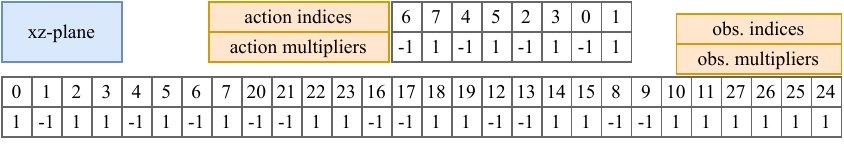}
\caption{xz-plane symmetry transformation for AntBulletEnv-v0}
\label{fig:ant_sym_1}
\end{figure}

The symmetry transformations are determined as explained in Section~\ref{sec:fitting_setup}. Consider the xz-plane depicted in Fig.~\ref{fig:ant}. The respective action transformation is composed of 8 \textit{action indices} and 8 \textit{action multipliers}, as listed in the top two rows of Fig.~\ref{fig:ant_sym_1}. Note that a positive action across all elements will cause knee flexion at joints 1 and 7, and knee extension at joints 3 and 5. Although for the xz-plane this results in direct proportion, for the yz-plane the relation is inverted.

The last two rows in Fig.~\ref{fig:ant_sym_1}, correspond to the \textit{observation indices} and \textit{observation multipliers}. The first column corresponds to the robot height as presented in Table~\ref{tab:state_space}. The height is invariant to any of the symmetry planes or rotations, so its index corresponds to itself, and the multiplier is 1. For the unit goal vector $(y,x)$, $y$ must be inverted and $x$ is invariant. Therefore, the index is kept unchanged (1 and 2), and the multiplier is -1 and 1, respectively. This logic can be applied to all the elements of Table~\ref{tab:state_space} to fill the remaining columns of \textit{observation indices} and \textit{observation multipliers}. See \ref{app:A} for the symmetry transformations concerning the remaining symmetry transformations.

\section{Evaluation Scenarios} \label{sec:eval_scenarios}

In all scenarios, the main objective for the robot is to move as far as possible towards the goal, set at an unreachable distance, while minimizing sideways deviation. The original reward function from AntBulletEnv-v0 was not changed, and it encompasses: stay-alive bonus, progress towards the objective, electricity cost, joints-at-limit cost to discourage stuck joints, and foot collision cost. An episode ends if the ant's torso makes contact with the ground or after 1000 steps. 

\subsection{Main objective}

\begin{figure}[htbp]
\centering
\includegraphics[width=0.4\columnwidth]{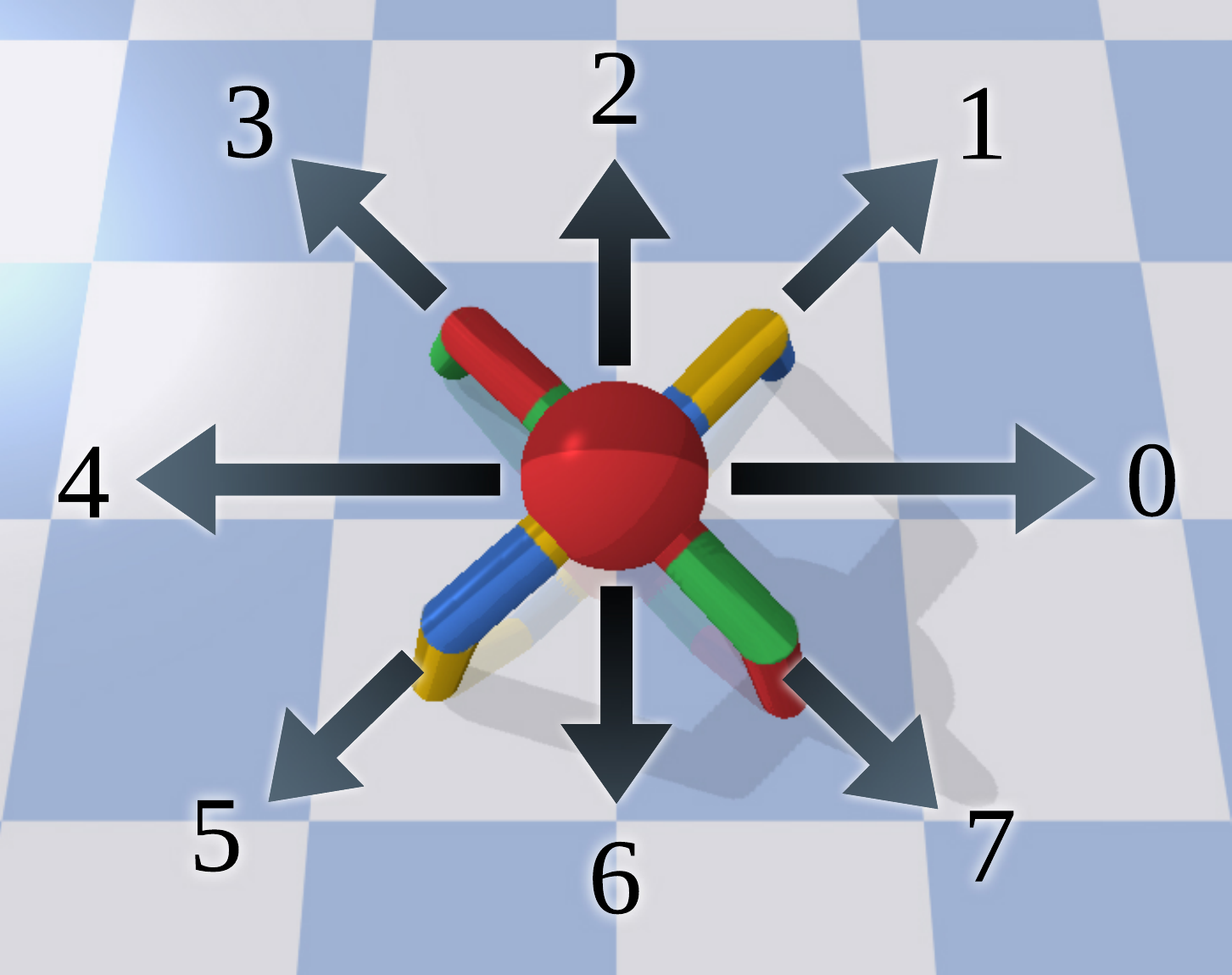}
\caption{Sequence of goals for 8-directional ant locomotion}
\label{fig:ant_goals}
\end{figure}

Contrary to training a single goal direction like the original AntBulletEnv-v0, the proposed scenarios define 8 possible directions with a fixed angular distance of 45 deg as shown in Fig.~\ref{fig:ant_goals}. For a batch of episodes, the goal generation is sequential (0 to 7) and not random, in order to reduce the variability while training and also to reduce biases when periodically evaluating the policy. In our environment implementation (see Section \ref{sec:codebase}), to minimize code modifications, we changed the initial robot orientation instead of the goal location.

\begin{figure*}[htbp]
\centering
\includegraphics[width=0.8\columnwidth]{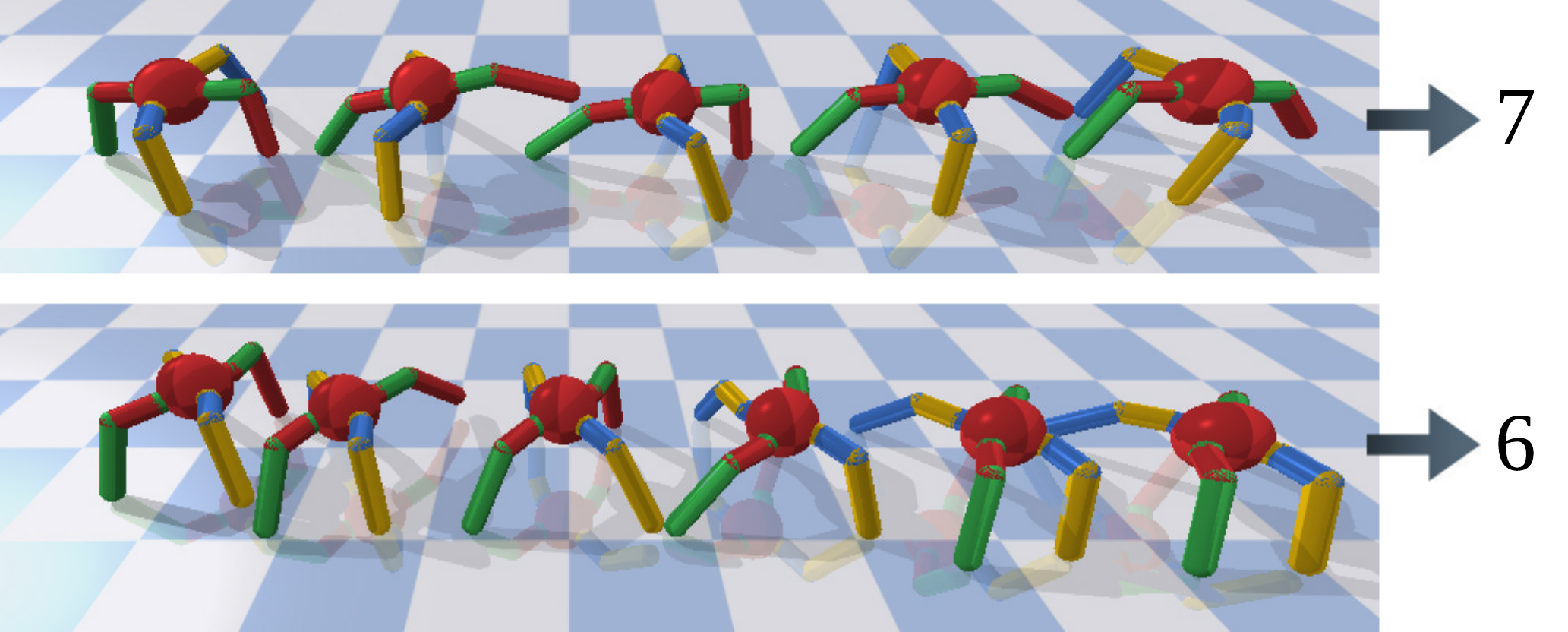}
\caption{Sequential frames of the ant robot moving to the right, in pursuit of goal 6 (bottom) and 7 (top)}
\label{fig:ant_snap}
\end{figure*}

During preliminary experiments the robot had a tendency to move towards the goal with one front leg, one back leg, and 2 side legs. In practice, if the goals were 1, 3, 5 or 7, the robot would move without rotating its body, while for other goals, it would first rotate until one of the legs was pointing to the goal. To prevent this bias, in addition to the terminal conditions applied by the original AntBulletEnv-v0 environment, the episode is terminated if the robot's orientation deviates more than 25 deg from its initial value. Fig.~\ref{fig:ant_snap} presents sequential frames of example runs where the robot is pursuing goal 6 (bottom) and 7 (top). 

\subsection{Controlled scenarios}

The first set of proposed scenarios is listed in Table~\ref{tab:eval_scenarios_1}. Scenario \textbf{A1.1} assumes that no perturbation is introduced by the robot, environment, reward function or control mechanism. The action modifier vector from the second column is multiplied by the policy's output (a vector of ones means no modification). The third column indicates the range to which the action is clipped after being modified. For Scenario \textbf{A1.1}, all action elements are clipped to [-1,1] before being sent to the robot. The last column indicates the goals used during training, where the index corresponds to the goal numbering denoted in Fig.~\ref{fig:ant_goals}. During evaluation, all scenarios presented in Table~\ref{tab:eval_scenarios_1} are tested for goals 0 to 7.

\begin{table}[t]
\caption{Evaluation scenarios I}
\begin{center}
\begin{tabular}{|l|c|c|c|}
\hline
\textbf{Scenario} & \textbf{Action Mod. (AM)} & \textbf{Action Range} & \textbf{Train goals}\\ \hline
\textbf{A1.1} & {[}1,1,1,1,1,1,1,1{]} & -1 to 1 & 0-7 \\ \hline
\textbf{A1.2} & {[}1,1,1,1,1,1,1,1{]} & -1 to 1 & 0,1 \\ \hline
\textbf{A2.1} & \multirow{2}{*}{{[}0.65,0.75,...,1.35{]}} & -1 to 1 & \multirow{4}{*}{0-7}  \\ \cline{1-1} \cline{3-3} 
\textbf{A2.2} &  & -AM to AM &  \\ \cline{1-3}
\textbf{A3.1} & \multirow{2}{*}{{[}0.30,0.50,...,1.70{]}} & -1 to 1 &  \\ \cline{1-1} \cline{3-3} 
\textbf{A3.2} &  & -AM to AM &  \\ \hline
\end{tabular}
\label{tab:eval_scenarios_1}
\end{center}
\end{table}

Scenario \textbf{A1.2} is similar in all aspects, except the training goals. While training, the robot is only exposed to goals 0 and 1 but, during evaluation, it is still tested on all goals (0-7). In theory, according to the ant robot's symmetry planes and rotations, symmetry-aware algorithms should be able to transpose the knowledge from goal 0 to (2, 4, 6) and from goal 1 to (3, 5, 7).

The remaining scenarios from Table~\ref{tab:eval_scenarios_1} simulate an artificial perturbation in the action controller. In \textbf{A2}, each action element is multiplied by a value in the vector [0.65,0.75,...,1.35]. \textbf{A2.1} then clips the values between -1 and 1. In this case all motors are equivalent, there is only a control imbalance. This means that a symmetric outcome is still possible if the perturbation is compensated by the policy. 

However, \textbf{A2.2} clips the action values between [-0.65,-0.75,...,-1.35] and [0.65,0.75,...,1.35]. Now, some actuators have different maximum torque values (e.g. the maximum torque for the knee at joint 1 in relation to the knee at joint 7 is 55\%). This problem is not trivial and requires asymmetric learning, independently of perturbation corrections. In this example, the ability to perform symmetric actions in symmetric states is not guaranteed. Scenarios \textbf{A3} are a harder version of \textbf{A2} in terms of action modification.

\subsection{Realistic scenarios} \label{sec:realistic_scenarios}

A second set of scenarios listed in Table~\ref{tab:eval_scenarios_2} was devised to test realistic robot and environment perturbations. As mentioned in the beginning of Section \ref{sec:eval_scenarios}, AntBulletEnv-v0's reward function has a stay-alive bonus. Its purpose is to reward the agent for lifting its torso above the ground at all times. Yet, in harsh conditions, the agent may learn to exploit the stay-alive bonus by not moving, since any wrong movement can end the episode earlier. To avoid changing the reward function, a new condition was added to the environment for the second set of scenarios --- the episode ends if no progress was made towards the goal in the last 30 steps.

\begin{table}[ht]
\caption{Evaluation scenarios II}
\begin{center}
\begin{tabular}{|l|c|c|cc|}
\hline
\textbf{Scenario} & \textbf{Feet Mass} & \textbf{Surface} & \multicolumn{1}{c|}{\textbf{Train Goals}} & \textbf{Eval. Goals} \\ \hline
\textbf{A4.1} & \multirow{4}{*}{{[}40,20,$\kappa$,$\kappa${]}} & \multirow{8}{*}{flat} & \multicolumn{2}{c|}{0-7}\\ \cline{1-1} \cline{4-5} 
\textbf{A4.2} &  &  & \multicolumn{2}{c|}{1,3,5,7} \\ \cline{1-1} \cline{4-5} 
\textbf{A4.3} &  &  & \multicolumn{2}{c|}{0,2,4,6} \\ \cline{1-1} \cline{4-5} 
\textbf{A5.1} &  &  & \multicolumn{1}{c|}{0,1,2,3} & 0-7 \\ \cline{1-2} \cline{4-5}
\textbf{A4.4} & \multirow{4}{*}{{[}40,$\kappa$,20,$\kappa${]}} &  & \multicolumn{2}{c|}{0-7} \\ \cline{1-1} \cline{4-5} 
\textbf{A4.5} &  &  & \multicolumn{2}{c|}{1,3,5,7} \\ \cline{1-1} \cline{4-5} 
\textbf{A4.6} &  &  & \multicolumn{2}{c|}{0,2,4,6} \\ \cline{1-1} \cline{4-5}
\textbf{A5.2} &  &  & \multicolumn{1}{c|}{0,1,2,3} & 0-7 \\ \hline
\textbf{A6.1} & \multirow{6}{*}{{[}$\kappa$,$\kappa$,$\kappa$,$\kappa${]}} & \multirow{3}{*}{\begin{tabular}[c]{@{}l@{}}5\degree\,tilt \\ Dir: 0\end{tabular}} & \multicolumn{2}{c|}{0-7}\\ \cline{1-1} \cline{4-5} 
\textbf{A6.2} &  &  & \multicolumn{2}{c|}{1,3,5,7} \\ \cline{1-1} \cline{4-5} 
\textbf{A6.3} &  &  & \multicolumn{2}{c|}{0,2,4,6} \\ \cline{1-1} \cline{3-5} 
\textbf{A6.4} &  & \multirow{3}{*}{\begin{tabular}[c]{@{}l@{}}5\degree\,tilt \\ Dir: 1\end{tabular}} & \multicolumn{2}{c|}{0-7}     \\ \cline{1-1} \cline{4-5} 
\textbf{A6.5} &  &  & \multicolumn{2}{c|}{1,3,5,7} \\ \cline{1-1} \cline{4-5} 
\textbf{A6.6} &  &  & \multicolumn{2}{c|}{0,2,4,6} \\ \hline
\end{tabular}
\label{tab:eval_scenarios_2}
\end{center}
\end{table}

In Table~\ref{tab:eval_scenarios_2}, the second column specifies the mass of each foot, where the first vector element corresponds to the foot connected to joint 1 (see Fig.~\ref{fig:ant}), followed by joints 3, 5 and 7. The default mass $\kappa$ is 13.5. Scenarios \textbf{A4.1} to \textbf{A4.3} modify the mass of the first and second feet. \textbf{A4.1} is trained and evaluated on all goals. \textbf{A4.2} handles only odd goals (1,3,5,7) while \textbf{A4.3} handles only even goals (0,2,4,6). Scenarios \textbf{A4.4} to \textbf{A4.6} work in a similar way, except that they modify the first and third feet.

An explanation is required for some of the design choices. First, the different mass configurations are an attempt to create a large amount of perturbation patterns across all symmetry planes and rotations. Second, the modified mass values (40, 20) are a trade-off between disrupting existing models (i.e. models trained without the modification have no control over the robot after the modifications) and allowing training (since joint actuators have limited torque). Third, the training and evaluation goals are separated into groups to allow the additional assessment of 2 types of locomotion (associated with the odd and even goals).

Scenarios \textbf{A5.1} and \textbf{A5.2} are a complement to the perturbations tested in \textbf{A4.1} and \textbf{A4.4}, respectively, but the agent is only able to train on half of the goals. The objective is for the algorithm to learn the perturbations with the training goals and generalize that knowledge to the remaining goals. In this experiment, allowing only two training goals, like in \textbf{A1.2}, would make the symmetry fitting unfeasible, since some symmetric states would never be experienced --- hence the 4 training goals.

Finally, scenarios \textbf{A6} follow a similar setup to \textbf{A4} but instead of modifying the feet mass, the surface is tilted 5 deg. For \textbf{A6.1} to \textbf{A6.3}, the tilt is in the direction of goal 0 (i.e. gravity pulls the robot towards goal 0), and for \textbf{A6.4} to \textbf{A6.6}, the tilt is in the direction of goal 1. It is important to note that if the agent was able to know the tilt direction through observations, there would be no need for symmetry fitting. If for instance, the roll and pitch observations (see Table~\ref{tab:state_space}) were relative to the ground and not the tilted surface, the agent would know the tilt direction. Consider the example presented as follows.

\begin{figure}[t]
\centering
\includegraphics[width=0.80\columnwidth]{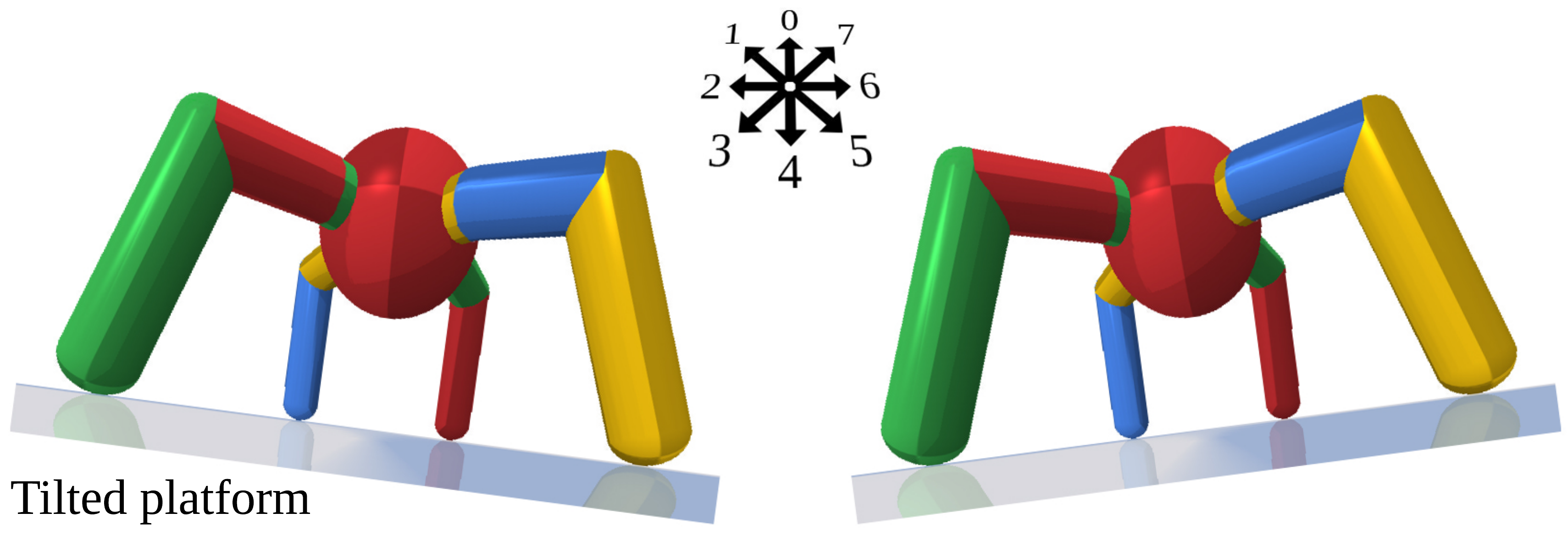}
\caption{Experienced tilted state (left) and its symmetric state in relation to the xz-plane (right)}
\label{fig:ant_tilt}
\end{figure}

Fig.~\ref{fig:ant_tilt} on the left, shows a scenario where the platform is tilted in the direction of goal 6. In each episode, the ant would chase a different goal direction, but the green and blue feet would always be stepping on higher ground, when directly compared to the other two feet (note that the robot cannot rotate more than 25 deg or the episode is terminated). So, for instance, considering the xz-plane, the symmetric state (on the right) is unreachable, and the algorithm will not leverage symmetry for this symmetry plane. This is the case for all symmetry rotations and symmetry planes, except for the yz-plane because it is aligned with the tilt direction. In conclusion, this is not a scenario where a symmetry loss function would be very useful.

With this in mind, it is possible to make use of symmetric states and symmetry loss functions when the perturbation is not part of the state space (i.e. it is not observable). In this way, the optimization algorithm will be able to generalize, to a certain extent, the learned behavior to the symmetric states, even if in practice the initial symmetry transformations do not accurately represent reality.

To make sure the roll and pitch observations were not affected, instead of tilting the surface, we changed the gravity vector to achieve the equivalent effect.

\section{Evaluation Methodology} \label{sec:methodology}

The performance in each scenario was evaluated for four algorithms: PPO (without any extension), PPO+ASL (PPO extended with Adaptive Symmetry Learning), PPO+PSL (PPO extended with Proximal Symmetry Loss), and PPO+MSL (PPO extended with Mirror Symmetry Loss). MSL and PSL employ the improved versions introduced in Section \ref{sec:MSL} and \ref{sec:PSL}, respectively.

All experiments were conducted on a single machine with two Intel Xeon 6258R with 28 cores. In total, 908 distinct hand-tuned configurations were tested. Each configuration was used to train 12 models. Typically, between 48 and 96 models were learned in parallel, requiring between 12h and 24h. A new set of hand-tuned configurations was designed after each learning round. 

Automatic tuning approaches were not used for two main reasons. First, the ASL algorithm was tweaked during the experiments to fix errors and test new methods and implementation techniques. This process requires constant human intervention since auto-tuning methods do not detect problems in the algorithm. Second, due to time and computational constraints, hand-tuning was the only viable option in terms of efficiency.

\subsection{Codebase} \label{sec:codebase}

The implementation of each extension was made in Python on top of a Stable Baselines 3 repository fork \cite{stable_baselines}. This codebase was chosen due to its high-quality implementation, considering how specific code-optimizations can have a deep impact on PPO's performance \cite{ppo_implementation}. The final code allows the user to select any of the aforementioned extensions per symmetry transformation, allowing for versatile combinations of symmetry extensions and parameters, although in the scope of this work, symmetry extensions were evaluated independently to allow for a fair comparison. All the algorithm implementations tested in this work, as well as code modification to the ant environment are available at our GitHub repository\footnote{\url{https://github.com/m-abr/Adaptive-Symmetry-Learning}}.

\subsection{Hyperparameters}

The choice of hyperparameters for PPO is crucial to enable successful applications, and should be tuned case by case \cite{ppo_hypers}. However, for consistency among scenarios and algorithms, a fixed set of hyperparameters was used, as listed in Table \ref{tab:ppo_hyperparameters}. 

\begin{table}[t]
	\caption{PPO Hyperparameters}
	\begin{center}
		\begin{tabular}[t]{|l|c|}
			\hline
			\textbf{Parameter} & \textbf{Value} \\ \hline
			batch steps & 4096  \\ \hline
			clip range & 0.4 \\ \hline
			entropy coefficient & 0 \\ \hline
			environments & 1 \\ \hline
			epochs & 20 \\ \hline
			GAE $\lambda$ (lambda) & 0.9 \\ \hline
			$\gamma$ (gamma) & 0.99 \\ \hline
			learning rate & \num{3e-5} \\ \hline
			max. gradient norm & 0.5 \\ \hline
		\end{tabular}
		\begin{tabular}[t]{|l|c|}
			\hline
			\textbf{Parameter} & \textbf{Value}  \\ \hline
			mini-batch size & 64  \\ \hline
			NN activation & ReLU  \\ \hline
			NN log std init & -1 \\ \hline
			NN orthogonal init & false  \\ \hline
			normalize advantage & true \\ \hline
			policy NN & [256,256] \\ \hline
			value coefficient & 0.5 \\ \hline
			value NN & [256,256] \\ \hline
		\end{tabular}
		\label{tab:ppo_hyperparameters}
	\end{center}
\end{table}

The chosen values were adapted from the RL Baselines3 Zoo training framework 
\cite{rl-zoo3}. The number of environments was reduced from 16 to 1 to improve 
performance while training parallel models (by eliminating the thread 
synchronization overhead). Considering this modification and the increased 
complexity imposed by the proposed scenarios, in comparison with the original 
AntBulletEnv-v0 environment, some other parameter values were changed: batch 
steps, mini-batch size and the total time steps (which is defined per 
scenario). 

Regarding the evaluated symmetry algorithms, their hyperparameters were tuned 
per scenario. Nonetheless, the values presented in 
Table~\ref{tab:common_asl_hyperparameters} were common to all cases. Symmetry 
transformations $\boldsymbol{f}(s)$ and $\boldsymbol{g}(a)$ represent the 
initial knowledge about the ant robot, assuming a perturbation-free system. The 
remaining hyperparameter vectors in Table~\ref{tab:common_asl_hyperparameters} 
were assigned a scalar as a slight abuse of notation to convey that the same 
scalar was assigned to all symmetry transformations, with the exception of 
$\boldsymbol{k}_\text{d}$[rot], which concerns only the subset of rotations.

\begin{table}[ht]
	\caption{Common ASL Hyperparameters}
	\begin{center}
		\begin{tabular}{|c|r|l|c|}
			\hline
			\textbf{Method} & \textbf{Symbol} & \textbf{Short Description} & 
			\textbf{Value} \\ \hline
			all & $\boldsymbol{f}(s)$ & sym. state transf. & 
			\multirow{2}{*}{Fig.~\ref{fig:ant_sym_1} and \ref{fig:ant_sym_2}} 
			\\ \cline{1-3}
			all & $\boldsymbol{g}(a)$ & sym. action transf. &  \\ \hline
			all & $\text{\textit{\textbf{w}}}_\text{V}$ & value sym. loss 
			weight & 0.5  \\ \hline
			ASL & $\text{H}_\text{U}$ & cycle penalty & 
			$0.05\cdot\frac{10^{-2}}{(10^{-2} + x^4)}$  \\ \hline
			ASL & $\text{H}_\text{G}$ & action transf. penalty & $1.1^{-x}$  \\ 
			\hline
			ASL & $\boldsymbol{k}_\text{d}$[rot] & dead zone (rotation) & 0 \\ 
			\hline
			ASL & $\boldsymbol{k}_\text{t}$ & time steps (neutral state) & 10 
			$\!\times\!$ (batch steps)\\ \hline
			ASL & $\boldsymbol{k}_\text{v}$ & value gate factor & 1.5  \\ \hline
		\end{tabular}
		\label{tab:common_asl_hyperparameters}
	\end{center}
\end{table}

The value symmetry loss function is the same in all evaluated symmetry algorithms. Since $\text{\textit{\textbf{w}}}_\text{V}^{PPO}=\text{\textit{\textbf{w}}}_\text{V}^{ASL}=\text{\textit{\textbf{w}}}_\text{V}^{PSL}=\text{\textit{\textbf{w}}}_\text{V}^{MSL}$ typically produces good results, and $\text{\textit{\textbf{w}}}_\text{V}^{PPO}=0.5$ (see Table~\ref{tab:ppo_hyperparameters}), parameter $\text{\textit{\textbf{w}}}_\text{V}$ was also set to 0.5 for all symmetry algorithms.

In the scenarios tested in this paper, neutral states affect reflections but not rotations. Every time a rotation transformation is applied, the unit goal vector from the state space (see Table~\ref{tab:state_space}) is also rotated, so the symmetric state can never be the same as the initial state, which is a necessary condition for neutral states. For that reason $k_\text{d}=0$ for all rotations. For reflections, $\boldsymbol{k}_\text{d}$ was set independently for each scenario. The remaining parameters from Table~\ref{tab:common_asl_hyperparameters} were found to be consistent across 
all experiments. 

\subsection{Metrics}

\begin{figure}[!t]
	\centering
	\includegraphics[width=0.60\columnwidth]{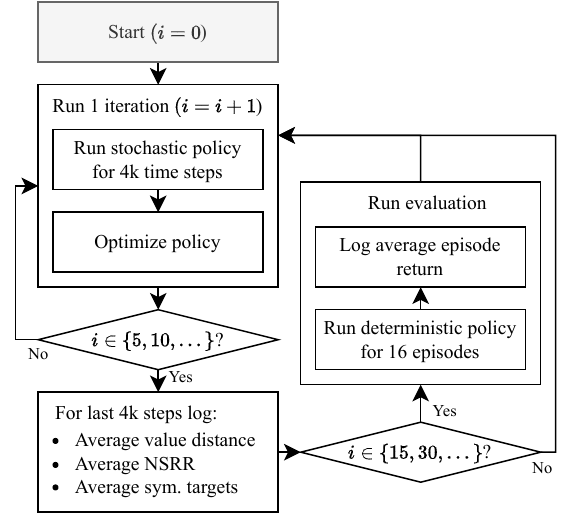}
	\caption{Reinforcement learning flowchart with integrated performance evaluation every 15 iterations. Some variables are logged every 5 iterations.}
	\label{fig:metrics}
\end{figure}

All algorithms were evaluated and compared in terms of performance. Additional metrics were used to analyze symmetric properties. The reinforcement learning process alternates between training for 15 iterations ($15 \times 4096$ batch steps), and evaluating the policy for 16 episodes (using the deterministic neural network output), as depicted in Fig.~\ref{fig:metrics}. Some variables are logged every 5 iterations, depending on the chosen symmetry extension, and are averaged over the last 4096 time steps. For all metrics, each data point is then averaged again for 12 independent instances. In total, four variables were generated: 

\begin{enumerate}
	\item Average episode return --- the undiscounted return provides a raw measure of performance that takes into consideration all the implicit objectives defined in the reward function;
	\item Average value distance --- absolute difference between the value estimates of the explored state and the corresponding symmetric state. This metric assesses how accurately the current symmetry transformations describe the policy for explored states; % note: these were not displayed for PPO-baseline as the data is not meaningful (high variance for highly random policies)
	\item Average neutral state rejection ratio (NSRR) --- ratio of states that were rejected by the Adaptive Symmetry Learning algorithm (see Section~\ref{sec:neutral_exclusion}). The NSRR measures to what extent the neutral state rejection module was actually applied to each symmetry plane: xz-plane, yz-plane, y=x plane, and y=-x plane.
	\item Average symmetry transformation targets --- target values for the current action transformation approximation to a user-defined parameterized function. For linear functions, the ant model has 12 action multipliers (e.g. $m_{0\rightarrow 2}$) and 16 actions biases (12 based on symmetric pairs, such as $b_{0\rightarrow 2}$, and 4 based on reflexive relations such as $b_{0\rightarrow 0}$ or $b_{0}$). This metric evaluates the repeatability of results while comparing independent optimizations, and the accuracy of the learned action transformations if the ground truth is known. A detailed explanation for linear approximations is given in the next paragraph.
\end{enumerate}

The problem of learning action transformations (see their initial values in Fig.~\ref{fig:ant_sym_1} and \ref{fig:ant_sym_2}) can be simplified due to the existence of redundant symmetry relations. The ant model has 8 actions and 7 non-identity symmetry transformations, totaling 56 action transformations. Consequently, there are 112 variables if the operations can be defined as linear equations of the form $a'[y]=m_{x\rightarrow y}\cdot a[x] + b_{x\rightarrow y}$, or 56 variables if $b$ is not considered. Analogously to the process described in Fig.~\ref{fig:sym_example_pairs}, the ant's action transformations can be written as a function of 12 pairs of symmetric action elements \{(0,2),(0,4),(0,6),(1,3),\hspace{0pt}(1,5),(1,7),\hspace{0pt}(2,4),(2,6),\hspace{0pt}(3,5),\hspace{0pt}(3,7),\hspace{0pt}(4,6),(5,7)\} and 4 singles \{0,2,4,6\}, yielding 12 actions multipliers and 16 biases (totaling 28 independent variables for $y=mx+b$ operations).

\section{Evaluation Results} \label{sec:results}

The hyperparameters used for each scenario are listed in \ref{app:B} and the evaluated metrics are plotted in   \ref{app:C}. Below are discussed the results for each scenario. For the purpose of this analysis, PPO without any extension is considered the baseline. 

Due to lacking a mandatory progression condition, the controlled scenarios (\textbf{A1.1} to \textbf{A3.2}) exhibit a similar behavior in the beginning of training (see Fig.~\ref{fig:res_return}). The local peak at approximately 0.3 million time steps corresponds to the algorithm learning how to exploit the stay-alive bonus by not moving to avoid irrecoverable mistakes, as explained in Section~\ref{sec:realistic_scenarios}. Then, the slight drop coincides with a more exploratory phase, where the agent starts taking risks.

The average symmetry transformation target error in Fig.~\ref{fig:res_target1} was only plotted for scenarios \textbf{A2} and \textbf{A3}. For \textbf{A2.1} and \textbf{A3.1} the ground truth is known. Let $\boldsymbol{x}$ be the action multiplier vector shown in the second column of Table~\ref{tab:eval_scenarios_1}. Consider for instance the symmetric relation between two action elements $a_0$ and $a_2$. To achieve a symmetric outcome, the equality $a_0x_0 \!=\! a_2x_2$ must hold. Therefore, the transformation parameterized by $m_{0\rightarrow 2}$ can be obtained such that

\begin{equation}
	\label{eq:ground_truth}
	\begin{cases}
		a_0x_0 \!=\! a_2x_2 \\
		a_2 \!=\! a_0m_{0\rightarrow 2}
	\end{cases}\hspace{-6pt}\Rightarrow\;\,
	a_0x_0 \!=\! a_0m_{0\rightarrow 2}x_2 \\
	\;\,\Rightarrow\;\,
	m_{0\rightarrow 2} \!=\! \frac{x_0}{x_2}.
\end{equation}

So, in Fig.~\ref{fig:res_target1}, the ground truth for $[m_{0\rightarrow 2}$,$m_{0\rightarrow 4}$,\dots,$m_{5\rightarrow 7}]$ is given by $[0.76,0.62,0.52,\allowbreak0.79,0.65,0.56,\allowbreak0.81,0.68,0.83,\allowbreak0.70,0.84,0.85]$ for \textbf{A2.1}, and $[0.43,0.27,0.20,0.56,0.38,\allowbreak0.29,\allowbreak0.64,0.47,0.69,0.53,0.73,0.76]$ for \textbf{A3.1}. Since the ground truth for \textbf{A2.2} and \textbf{A3.2} is unknown, for comparison, we reuse the values used for \textbf{A2.1} and \textbf{A3.1}, respectively.

The remainder of this section delves into the discussion of specific results for each scenario. For the sake of simplicity, when mentioning ASL, it should be read as PPO+ASL, and the same is true for the other symmetry extensions.

\begin{table}[!b]
	\caption{Statistics for scenarios \textbf{A1.1} (top) and \textbf{A1.2} (bottom) extracted from average episode return (see Fig.~\ref{fig:res_return}) and average value distance (see Fig.~\ref{fig:res_valdiff}) }
	
	\footnotesize
	\begin{center}
		\begin{tabular}{|x{1.1cm}|ccc|x{1.6cm}|x{1.55cm}|cc|}
			\hline
			\multirow{3}{*}{\parbox{1cm}{\textbf{Algo-rithm}}} & \multicolumn{3}{x{4.0cm}|}{\textbf{Maximum Return \; (5pt window)}} & \textbf{TS when Return } & \textbf{Value Distance} & \multicolumn{2}{x{3.1cm}|}{\textbf{Value Distance (last 5pt window)}} \\ \cline{2-4} \cline{6-8} 
			& \multicolumn{1}{c|}{\textbf{Mean} $\boldsymbol{\mu}$} & \multicolumn{1}{c|}{\textbf{SD}} & \textbf{T.Step} & $\boldsymbol{\geq 0.9\mu}$ & \textbf{Mean} & \multicolumn{1}{c|}{\textbf{Mean}} & \textbf{SD} \\ \hline
			PPO & \multicolumn{1}{c|}{892} & \multicolumn{1}{c|}{12.6} & 0.68M & 0.37M & - & \multicolumn{1}{c|}{-} & - \\ \hline
			+ASL & \multicolumn{1}{c|}{2183} & \multicolumn{1}{c|}{29.7} & 3.82M & 2.65M & 1.01 & \multicolumn{1}{c|}{0.96} & 0.06 \\ \hline
			+PSL & \multicolumn{1}{c|}{2157} & \multicolumn{1}{c|}{20.2} & 3.88M & 2.65M & 1.08 & \multicolumn{1}{c|}{1.27} & 0.14 \\ \hline
			+MSL & \multicolumn{1}{c|}{2082} & \multicolumn{1}{c|}{53.5} & 3.63M & 2.77M & 1.10 & \multicolumn{1}{c|}{1.26} & 0.04 \\ \hline
		\end{tabular}
		\vspace{0.2cm}
		
		\begin{tabular}{|x{1.1cm}|ccc|x{1.6cm}|x{1.55cm}|cc|}
			\hline
			\multirow{3}{*}{\parbox{1cm}{\textbf{Algo-rithm}}} & \multicolumn{3}{x{4.0cm}|}{\textbf{Maximum Return \; (5pt window)}} & \textbf{TS when Return } & \textbf{Value Distance} & \multicolumn{2}{x{3.1cm}|}{\textbf{Value Distance (last 5pt window)}} \\ \cline{2-4} \cline{6-8} 
			& \multicolumn{1}{c|}{\textbf{Mean} $\boldsymbol{\mu}$} & \multicolumn{1}{c|}{\textbf{SD}} & \textbf{T.Step} & $\boldsymbol{\geq 0.9\mu}$ & \textbf{Mean} & \multicolumn{1}{c|}{\textbf{Mean}} & \textbf{SD} \\ \hline
			PPO & \multicolumn{1}{c|}{824} & \multicolumn{1}{c|}{19.4} & 3.57M & 1.72M & - & \multicolumn{1}{c|}{-} & - \\ \hline
			+ASL & \multicolumn{1}{c|}{2580} & \multicolumn{1}{c|}{41.6} & 3.88M & 1.91M & 1.04 & \multicolumn{1}{c|}{1.07} & 0.06 \\ \hline
			+PSL & \multicolumn{1}{c|}{2040} & \multicolumn{1}{c|}{49.1} & 3.63M & 2.15M & 1.20 & \multicolumn{1}{c|}{1.46} & 0.15 \\ \hline
			+MSL & \multicolumn{1}{c|}{2078} & \multicolumn{1}{c|}{21.1} & 3.88M & 2.46M & 1.03 & \multicolumn{1}{c|}{1.27} & 0.11 \\ \hline
		\end{tabular}
		\label{tab:stats_A1}
	\end{center}
\end{table}

\bigskip
\noindent\textbf{A1.1}
\smallskip

Table~\ref{tab:stats_A1} summarizes relevant statistics for the two evaluation metrics that are applied to all symmetric algorithms: the average episode return shown in Fig.~\ref{fig:res_return} and the average value distance shown in Fig.~\ref{fig:res_valdiff}. The details of each column are provided below:

\begin{itemize}
	\item \textbf{Algorithm} --- Baselines (PPO) and symmetry extended algorithms (PPO+ASL, PPO+PSL, and PPO+MSL).
	\item \textbf{Maximum Return} — This metric is calculated by sliding a 5-point window (equivalent to 75 learning iterations or 307,200 time steps) over the plots shown in Fig.~\ref{fig:res_return}. We identify the window with the highest mean return $\boldsymbol{\mu}$, reporting its standard deviation (SD) and the midpoint (the time step at the center of the window).
	\item \textbf{TS when Return} $\boldsymbol{\geq 0.9\mu}$ --- First time step when the return is greater than or equal to 90\% of the highest mean return $\boldsymbol{\mu}$.
	\item \textbf{Value Distance} --- Mean value for the plots shown in Fig.~\ref{fig:res_valdiff}.
	\item \textbf{Value Distance (last 5pt window)} --- Reports the mean and standard deviation (SD) of the last 5-point window in the plots shown in Fig.~\ref{fig:res_valdiff}. In other words, this metric represents the Value Distance at the end of the learning process.
\end{itemize}

All symmetric algorithms performed similarly in this scenario over 4 million time steps, with a maximum return increase between 133\% and 145\% in relation to the baseline, which plateaued early, reaching 90\% of maximum return by 0.37 million time steps. ASL performed best without symmetry fitting, which indicates that linear modifications to the symmetry transformations yield no advantage in this context.

In terms of average value distance (see Fig.~\ref{fig:res_valdiff}), all the symmetry extensions follow a common pattern of allowing asymmetric exploration until around \num{2.2e6} time steps, before the model starts to converge to the provided symmetry transformations. As seen in Table~\ref{tab:stats_A1}, the value distance at the end of the learning process is similar across methods, with ASL showing a marginally lower value. This suggests a slightly higher preference for symmetry in the final policy, although it does not translate into a performance gain when compared to PSL.

The average NSRR for ASL shown in Fig.~\ref{fig:res_nsrr} reveals the contrast between both gaits depicted in Fig.~\ref{fig:ant_snap} in terms of neutral state incidence. When walking with one front leg, a symmetric policy is a good policy if and only if the initial position is perfectly aligned with the objective, and there is no perturbation in the policy or environment. These conditions are impractical, even more so due to the initial random component introduced by AntBulletEnv-v0. Nonetheless, walking with one front leg results in an increasing number of neutral states as the policy converges to local optima, resulting in an increasing NSRR for planes y=x and y=-x. In contrast, walking with two front legs rarely generates neutral states, hence the low NSRR for planes $xz$ and $yz$.

\bigskip
\noindent\textbf{A1.2}
\smallskip

In comparison with the previous scenario, the policy was only able to train on goals 0 and 1, having to generalize its behavior to the other 6 goals. In this setting, ASL outperformed the other three methods, and itself in \textbf{A1.1}. Both symmetry fitting and the value gate were disabled. Since most value estimates were being modified by the value loss and not by exploration, the value gate lost its purpose and became counterproductive, consequently being disabled. Therefore, the only modules that contributed to this performance were the ratio with fixed standard deviation in \eqref{eq:ASL_impl_r}, and neutral state rejection (Section \ref{sec:neutral_exclusion}). 

The value distance plots show that the policies under all symmetry methods converged earlier to the symmetry transformations, which led to a steeper learning curve in all three cases. The NSRR of planes yz and y=-x was zero during training because the agents only trained on goals 0 and 1. Yet, the NSRR for plane y=x indicates that convergence to the symmetry transformations was much faster in comparison with \textbf{A1.1}, and for the xz-plane it was constantly higher. These results show that symmetry in gaits is correlated with performance in this scenario, and that generalizing to symmetric states can be more efficient when training only in a subset of non-symmetrically-redundant states. Note that in \textbf{A1.1}, the policy is also being generalized to symmetric states. The only difference is that states similar to those generalized also occur during training in \textbf{A1.1}.

\bigskip
\noindent\textbf{A2}
\smallskip

From this point forward, symmetry fitting was enabled on all experiments. In \textbf{A2.1}, the introduced modifiers are proportional to the action, and in \textbf{A2.2} there is an additional nonlinear component. In both cases, as supported by the Value Distance in Table~\ref{tab:stats_A2}, the ASL policies are following the learned symmetry transformation closer, in comparison with the other methods. In \textbf{A2.1}, the maximum return is identical to \textbf{A1.1}, indicating that symmetry fitting was able to compensate for the action modifiers without performance loss, while PSL and MSL experienced some decay by having the extra effort of learning an asymmetric policy. In \textbf{A2.2}, as expected, symmetry fitting could only handle linear perturbations, resulting in less advantage over the other approaches. This difference is also evident from the NSRR data, when comparing both scenarios with \textbf{A1.1}.

\begin{table}[!t]
	\caption{Statistics for scenarios \textbf{A2.1} (top) and \textbf{A2.2} (bottom) extracted from average episode return (see Fig.~\ref{fig:res_return}) and average value distance (see Fig.~\ref{fig:res_valdiff}) }
	\footnotesize
	\begin{center}
		\begin{tabular}{|x{1.1cm}|ccc|x{1.6cm}|x{1.55cm}|cc|}
			\hline
			\multirow{3}{*}{\parbox{1cm}{\textbf{Algo-rithm}}} & \multicolumn{3}{x{4.0cm}|}{\textbf{Maximum Return \; (5pt window)}} & \textbf{TS when Return } & \textbf{Value Distance} & \multicolumn{2}{x{3.1cm}|}{\textbf{Value Distance (last 5pt window)}} \\ \cline{2-4} \cline{6-8} 
			& \multicolumn{1}{c|}{\textbf{Mean} $\boldsymbol{\mu}$} & \multicolumn{1}{c|}{\textbf{SD}} & \textbf{T.Step} & $\boldsymbol{\geq 0.9\mu}$ & \textbf{Mean} & \multicolumn{1}{c|}{\textbf{Mean}} & \textbf{SD} \\ \hline
			PPO & \multicolumn{1}{c|}{891} & \multicolumn{1}{c|}{19.1} & 0.49M & 0.31M & - & \multicolumn{1}{c|}{-} & - \\ \hline
			+ASL & \multicolumn{1}{c|}{2180} & \multicolumn{1}{c|}{33.5} & 3.69M & 2.65M & 0.95 & \multicolumn{1}{c|}{1.08} & 0.10 \\ \hline
			+PSL & \multicolumn{1}{c|}{2049} & \multicolumn{1}{c|}{22.0} & 3.45M & 2.83M & 1.24 & \multicolumn{1}{c|}{1.66} & 0.12 \\ \hline
			+MSL & \multicolumn{1}{c|}{1975} & \multicolumn{1}{c|}{25.9} & 3.88M & 2.65M & 1.25 & \multicolumn{1}{c|}{1.64} & 0.08 \\ \hline
		\end{tabular}
		\vspace{0.2cm}
		
		\begin{tabular}{|x{1.1cm}|ccc|x{1.6cm}|x{1.55cm}|cc|}
			\hline
			\multirow{3}{*}{\parbox{1cm}{\textbf{Algo-rithm}}} & \multicolumn{3}{x{4.0cm}|}{\textbf{Maximum Return \; (5pt window)}} & \textbf{TS when Return } & \textbf{Value Distance} & \multicolumn{2}{x{3.1cm}|}{\textbf{Value Distance (last 5pt window)}} \\ \cline{2-4} \cline{6-8} 
			& \multicolumn{1}{c|}{\textbf{Mean} $\boldsymbol{\mu}$} & \multicolumn{1}{c|}{\textbf{SD}} & \textbf{T.Step} & $\boldsymbol{\geq 0.9\mu}$ & \textbf{Mean} & \multicolumn{1}{c|}{\textbf{Mean}} & \textbf{SD} \\ \hline
			PPO & \multicolumn{1}{c|}{880} & \multicolumn{1}{c|}{21.5} & 0.55M & 0.31M & - & \multicolumn{1}{c|}{-} & - \\ \hline
			+ASL & \multicolumn{1}{c|}{2087} & \multicolumn{1}{c|}{18.8} & 3.88M & 2.71M & 1.02 & \multicolumn{1}{c|}{1.17} & 0.07 \\ \hline
			+PSL & \multicolumn{1}{c|}{1954} & \multicolumn{1}{c|}{54.1} & 3.88M & 2.95M & 1.21 & \multicolumn{1}{c|}{1.45} & 0.06 \\ \hline
			+MSL & \multicolumn{1}{c|}{1932} & \multicolumn{1}{c|}{37.1} & 3.82M & 2.95M & 1.39 & \multicolumn{1}{c|}{1.83} & 0.18 \\ \hline
		\end{tabular}
		\label{tab:stats_A2}
	\end{center}
\end{table}

By observing Fig.~\ref{fig:res_target1}, it is possible to conclude that the learned symmetry transformations are very similar in both \textbf{A2.1} and \textbf{A2.2}, and the respective error relative to the ground truth of \textbf{A2.1} converges to the vicinity of zero. Note that the initial symmetry transformation had an average error of 0.282 and, after training, that value dropped to 0.031 on \textbf{A2.1} and 0.049 on \textbf{A2.2}.

\bigskip
\noindent\textbf{A3}
\smallskip

Given the complexity of this scenario and the ones that followed, the training was prolonged to 5 million time steps. \textbf{A3.1} and \textbf{A3.2} are harder versions of the previous scenarios, with larger action modifiers. The maximum return in Table~\ref{tab:stats_A3} reveals a disproportionate effect of this additional difficulty on the extended methods, being PSL and MSL considerably more affected. In relation to \textbf{A2.1} and \textbf{A2.2}, ASL decayed 22\% and 19\%, PSL 29\% and 33\%, and MSL 42\% and 41\%, respectively.

\begin{table}[!b]
	\caption{Statistics for scenarios \textbf{A3.1} (top) and \textbf{A3.2} (bottom) extracted from average episode return (see Fig.~\ref{fig:res_return}) and average value distance (see Fig.~\ref{fig:res_valdiff}) }
	\footnotesize
	\begin{center}
		\begin{tabular}{|x{1.1cm}|ccc|x{1.6cm}|x{1.55cm}|cc|}
			\hline
			\multirow{3}{*}{\parbox{1cm}{\textbf{Algo-rithm}}} & \multicolumn{3}{x{4.0cm}|}{\textbf{Maximum Return \; (5pt window)}} & \textbf{TS when Return } & \textbf{Value Distance} & \multicolumn{2}{x{3.1cm}|}{\textbf{Value Distance (last 5pt window)}} \\ \cline{2-4} \cline{6-8} 
			& \multicolumn{1}{c|}{\textbf{Mean} $\boldsymbol{\mu}$} & \multicolumn{1}{c|}{\textbf{SD}} & \textbf{T.Step} & $\boldsymbol{\geq 0.9\mu}$ & \textbf{Mean} & \multicolumn{1}{c|}{\textbf{Mean}} & \textbf{SD} \\ \hline
			PPO & \multicolumn{1}{c|}{920} & \multicolumn{1}{c|}{10.8} & 1.17M & 0.43M & - & \multicolumn{1}{c|}{-} & - \\ \hline
			+ASL & \multicolumn{1}{c|}{1705} & \multicolumn{1}{c|}{17.5} & 4.75M & 3.15M & 1.45 & \multicolumn{1}{c|}{1.88} & 0.09 \\ \hline
			+PSL & \multicolumn{1}{c|}{1448} & \multicolumn{1}{c|}{55.9} & 4.88M & 3.83M & 1.60 & \multicolumn{1}{c|}{2.67} & 0.08 \\ \hline
			+MSL & \multicolumn{1}{c|}{1149} & \multicolumn{1}{c|}{32.8} & 3.70M & 3.09M & 1.45 & \multicolumn{1}{c|}{2.44} & 0.15 \\ \hline
		\end{tabular}
		\vspace{0.2cm}
		
		\begin{tabular}{|x{1.1cm}|ccc|x{1.6cm}|x{1.55cm}|cc|}
			\hline
			\multirow{3}{*}{\parbox{1cm}{\textbf{Algo-rithm}}} & \multicolumn{3}{x{4.0cm}|}{\textbf{Maximum Return \; (5pt window)}} & \textbf{TS when Return } & \textbf{Value Distance} & \multicolumn{2}{x{3.1cm}|}{\textbf{Value Distance (last 5pt window)}} \\ \cline{2-4} \cline{6-8} 
			& \multicolumn{1}{c|}{\textbf{Mean} $\boldsymbol{\mu}$} & \multicolumn{1}{c|}{\textbf{SD}} & \textbf{T.Step} & $\boldsymbol{\geq 0.9\mu}$ & \textbf{Mean} & \multicolumn{1}{c|}{\textbf{Mean}} & \textbf{SD} \\ \hline
			PPO & \multicolumn{1}{c|}{918} & \multicolumn{1}{c|}{9.4} & 0.86M & 0.37M & - & \multicolumn{1}{c|}{-} & - \\ \hline
			+ASL & \multicolumn{1}{c|}{1686} & \multicolumn{1}{c|}{50.7} & 4.63M & 3.33M & 1.36 & \multicolumn{1}{c|}{2.14} & 0.07 \\ \hline
			+PSL & \multicolumn{1}{c|}{1307} & \multicolumn{1}{c|}{51.2} & 4.63M & 3.15M & 1.68 & \multicolumn{1}{c|}{3.00} & 0.05 \\ \hline
			+MSL & \multicolumn{1}{c|}{1140} & \multicolumn{1}{c|}{31.3} & 3.89M & 2.28M & 1.54 & \multicolumn{1}{c|}{2.70} & 0.24 \\ \hline
		\end{tabular}
		\label{tab:stats_A3}
	\end{center}
\end{table}

\begin{table}[!b]
	\caption{Statistics for scenarios \textbf{A4.1} (top) to \textbf{A4.3} (bottom) extracted from average episode return (see Fig.~\ref{fig:res_return}) and average value distance (see Fig.~\ref{fig:res_valdiff}) }
	\footnotesize
	\begin{center}
		\begin{tabular}{|x{1.1cm}|ccc|x{1.6cm}|x{1.55cm}|cc|}
			\hline
			\multirow{3}{*}{\parbox{1cm}{\textbf{Algo-rithm}}} & \multicolumn{3}{x{4.0cm}|}{\textbf{Maximum Return \; (5pt window)}} & \textbf{TS when Return } & \textbf{Value Distance} & \multicolumn{2}{x{3.1cm}|}{\textbf{Value Distance (last 5pt window)}} \\ \cline{2-4} \cline{6-8} 
			& \multicolumn{1}{c|}{\textbf{Mean} $\boldsymbol{\mu}$} & \multicolumn{1}{c|}{\textbf{SD}} & \textbf{T.Step} & $\boldsymbol{\geq 0.9\mu}$ & \textbf{Mean} & \multicolumn{1}{c|}{\textbf{Mean}} & \textbf{SD} \\ \hline
			PPO & \multicolumn{1}{c|}{449} & \multicolumn{1}{c|}{10.8} & 3.58M & 2.72M & - & \multicolumn{1}{c|}{-} & - \\ \hline
			+ASL & \multicolumn{1}{c|}{1689} & \multicolumn{1}{c|}{29.5} & 4.51M & 2.90M & 1.77 & \multicolumn{1}{c|}{1.57} & 0.05 \\ \hline
			+PSL & \multicolumn{1}{c|}{1160} & \multicolumn{1}{c|}{32.6} & 4.81M & 3.95M & 2.36 & \multicolumn{1}{c|}{2.68} & 0.08 \\ \hline
			+MSL & \multicolumn{1}{c|}{1265} & \multicolumn{1}{c|}{26.2} & 4.44M & 2.78M & 2.16 & \multicolumn{1}{c|}{2.51} & 0.05 \\ \hline
		\end{tabular}
		\vspace{0.2cm}
		
		\begin{tabular}{|x{1.1cm}|ccc|x{1.6cm}|x{1.55cm}|cc|}
			\hline
			\multirow{3}{*}{\parbox{1cm}{\textbf{Algo-rithm}}} & \multicolumn{3}{x{4.0cm}|}{\textbf{Maximum Return \; (5pt window)}} & \textbf{TS when Return } & \textbf{Value Distance} & \multicolumn{2}{x{3.1cm}|}{\textbf{Value Distance (last 5pt window)}} \\ \cline{2-4} \cline{6-8} 
			& \multicolumn{1}{c|}{\textbf{Mean} $\boldsymbol{\mu}$} & \multicolumn{1}{c|}{\textbf{SD}} & \textbf{T.Step} & $\boldsymbol{\geq 0.9\mu}$ & \textbf{Mean} & \multicolumn{1}{c|}{\textbf{Mean}} & \textbf{SD} \\ \hline
			PPO & \multicolumn{1}{c|}{587} & \multicolumn{1}{c|}{26.4} & 4.51M & 3.46M & - & \multicolumn{1}{c|}{-} & - \\ \hline
			+ASL & \multicolumn{1}{c|}{1220} & \multicolumn{1}{c|}{36.6} & 4.88M & 3.58M & 1.67 & \multicolumn{1}{c|}{1.66} & 0.05 \\ \hline
			+PSL & \multicolumn{1}{c|}{1057} & \multicolumn{1}{c|}{40.3} & 4.88M & 4.26M & 1.95 & \multicolumn{1}{c|}{2.41} & 0.07 \\ \hline
			+MSL & \multicolumn{1}{c|}{1011} & \multicolumn{1}{c|}{36.7} & 4.75M & 4.07M & 1.81 & \multicolumn{1}{c|}{2.26} & 0.10 \\ \hline
		\end{tabular}
		\vspace{0.2cm}
		
		\begin{tabular}{|x{1.1cm}|ccc|x{1.6cm}|x{1.55cm}|cc|}
			\hline
			\multirow{3}{*}{\parbox{1cm}{\textbf{Algo-rithm}}} & \multicolumn{3}{x{4.0cm}|}{\textbf{Maximum Return \; (5pt window)}} & \textbf{TS when Return } & \textbf{Value Distance} & \multicolumn{2}{x{3.1cm}|}{\textbf{Value Distance (last 5pt window)}} \\ \cline{2-4} \cline{6-8} 
			& \multicolumn{1}{c|}{\textbf{Mean} $\boldsymbol{\mu}$} & \multicolumn{1}{c|}{\textbf{SD}} & \textbf{T.Step} & $\boldsymbol{\geq 0.9\mu}$ & \textbf{Mean} & \multicolumn{1}{c|}{\textbf{Mean}} & \textbf{SD} \\ \hline
			PPO & \multicolumn{1}{c|}{687} & \multicolumn{1}{c|}{20.1} & 3.33M & 2.96M & - & \multicolumn{1}{c|}{-} & - \\ \hline
			+ASL & \multicolumn{1}{c|}{1731} & \multicolumn{1}{c|}{41.5} & 3.77M & 2.59M & 1.22 & \multicolumn{1}{c|}{1.64} & 0.05 \\ \hline
			+PSL & \multicolumn{1}{c|}{1390} & \multicolumn{1}{c|}{27.5} & 4.88M & 3.27M & 1.70 & \multicolumn{1}{c|}{1.90} & 0.05 \\ \hline
			+MSL & \multicolumn{1}{c|}{1384} & \multicolumn{1}{c|}{57.5} & 4.32M & 2.41M & 1.46 & \multicolumn{1}{c|}{1.83} & 0.09 \\ \hline
		\end{tabular}
		\label{tab:stats_A4a}
	\end{center}
\end{table}

The final value distance shows that ASL learned valuable symmetry transformations but not as reliably as in \textbf{A2}, resulting in a more asymmetric policy. The average target error confirms this efficiency decline even though the initial symmetry transformation had an average error of 0.503 and, after training, that value dropped to 0.128 on \textbf{A2.1} and 0.106 on \textbf{A2.2}. As expected, due to some transformation targets still having a considerable error, the gait was more affected for some goals, resulting in an NSRR imbalance between plane y=x and y=-x.

\clearpage
%\bigskip
\noindent\textbf{A4}
\smallskip

In \textbf{A4.1} to \textbf{A4.3}, additional weight is placed on the first and second feet, while in \textbf{A4.4} to \textbf{A4.6}, the extra weight is placed on the first and third feet. This realistic scenario takes advantage of symmetry fitting to adjust the linear imbalance in symmetry transformations, while the nonlinear adjustments are handled by PPO. For this reason, in general, ASL learns faster and reaches higher episode returns in the end, with performance gains between 3\% and 46\% in relation to PSL and MSL (see Table~\ref{tab:stats_A4a} and Table~\ref{tab:stats_A4b}).

\begin{table}[t]
	\caption{Statistics for scenarios \textbf{A4.4} (top) to \textbf{A4.6} (bottom) extracted from average episode return (see Fig.~\ref{fig:res_return}) and average value distance (see Fig.~\ref{fig:res_valdiff}) }
	\footnotesize
	\begin{center}
		\begin{tabular}{|x{1.1cm}|ccc|x{1.6cm}|x{1.55cm}|cc|}
			\hline
			\multirow{3}{*}{\parbox{1cm}{\textbf{Algo-rithm}}} & \multicolumn{3}{x{4.0cm}|}{\textbf{Maximum Return \; (5pt window)}} & \textbf{TS when Return } & \textbf{Value Distance} & \multicolumn{2}{x{3.1cm}|}{\textbf{Value Distance (last 5pt window)}} \\ \cline{2-4} \cline{6-8} 
			& \multicolumn{1}{c|}{\textbf{Mean} $\boldsymbol{\mu}$} & \multicolumn{1}{c|}{\textbf{SD}} & \textbf{T.Step} & $\boldsymbol{\geq 0.9\mu}$ & \textbf{Mean} & \multicolumn{1}{c|}{\textbf{Mean}} & \textbf{SD} \\ \hline
			PPO & \multicolumn{1}{c|}{449} & \multicolumn{1}{c|}{14.4} & 3.89M & 2.72M & - & \multicolumn{1}{c|}{-} & - \\ \hline
			+ASL & \multicolumn{1}{c|}{1784} & \multicolumn{1}{c|}{21.2} & 4.44M & 2.90M & 1.85 & \multicolumn{1}{c|}{1.71} & 0.05 \\ \hline
			+PSL & \multicolumn{1}{c|}{1560} & \multicolumn{1}{c|}{54.3} & 4.51M & 3.64M & 2.24 & \multicolumn{1}{c|}{2.30} & 0.11 \\ \hline
			+MSL & \multicolumn{1}{c|}{1557} & \multicolumn{1}{c|}{23.0} & 4.14M & 2.78M & 2.12 & \multicolumn{1}{c|}{2.39} & 0.14 \\ \hline
		\end{tabular}
		\vspace{0.2cm}
		
		\begin{tabular}{|x{1.1cm}|ccc|x{1.6cm}|x{1.55cm}|cc|}
			\hline
			\multirow{3}{*}{\parbox{1cm}{\textbf{Algo-rithm}}} & \multicolumn{3}{x{4.0cm}|}{\textbf{Maximum Return \; (5pt window)}} & \textbf{TS when Return } & \textbf{Value Distance} & \multicolumn{2}{x{3.1cm}|}{\textbf{Value Distance (last 5pt window)}} \\ \cline{2-4} \cline{6-8} 
			& \multicolumn{1}{c|}{\textbf{Mean} $\boldsymbol{\mu}$} & \multicolumn{1}{c|}{\textbf{SD}} & \textbf{T.Step} & $\boldsymbol{\geq 0.9\mu}$ & \textbf{Mean} & \multicolumn{1}{c|}{\textbf{Mean}} & \textbf{SD} \\ \hline
			PPO & \multicolumn{1}{c|}{661} & \multicolumn{1}{c|}{12.2} & 3.95M & 2.59M & - & \multicolumn{1}{c|}{-} & - \\ \hline
			+ASL & \multicolumn{1}{c|}{1544} & \multicolumn{1}{c|}{42.0} & 4.51M & 2.78M & 2.31 & \multicolumn{1}{c|}{2.49} & 0.10 \\ \hline
			+PSL & \multicolumn{1}{c|}{1374} & \multicolumn{1}{c|}{35.0} & 4.81M & 3.77M & 2.38 & \multicolumn{1}{c|}{2.72} & 0.05 \\ \hline
			+MSL & \multicolumn{1}{c|}{1079} & \multicolumn{1}{c|}{30.7} & 4.38M & 3.33M & 2.51 & \multicolumn{1}{c|}{3.15} & 0.11 \\ \hline
		\end{tabular}
		\vspace{0.2cm}
		
		\begin{tabular}{|x{1.1cm}|ccc|x{1.6cm}|x{1.55cm}|cc|}
			\hline
			\multirow{3}{*}{\parbox{1cm}{\textbf{Algo-rithm}}} & \multicolumn{3}{x{4.0cm}|}{\textbf{Maximum Return \; (5pt window)}} & \textbf{TS when Return } & \textbf{Value Distance} & \multicolumn{2}{x{3.1cm}|}{\textbf{Value Distance (last 5pt window)}} \\ \cline{2-4} \cline{6-8} 
			& \multicolumn{1}{c|}{\textbf{Mean} $\boldsymbol{\mu}$} & \multicolumn{1}{c|}{\textbf{SD}} & \textbf{T.Step} & $\boldsymbol{\geq 0.9\mu}$ & \textbf{Mean} & \multicolumn{1}{c|}{\textbf{Mean}} & \textbf{SD} \\ \hline
			PPO & \multicolumn{1}{c|}{751} & \multicolumn{1}{c|}{17.8} & 3.89M & 2.53M & - & \multicolumn{1}{c|}{-} & - \\ \hline
			+ASL & \multicolumn{1}{c|}{1893} & \multicolumn{1}{c|}{29.7} & 3.95M & 3.09M & 1.44 & \multicolumn{1}{c|}{1.75} & 0.07 \\ \hline
			+PSL & \multicolumn{1}{c|}{1834} & \multicolumn{1}{c|}{38.4} & 4.88M & 3.46M & 1.39 & \multicolumn{1}{c|}{1.48} & 0.06 \\ \hline
			+MSL & \multicolumn{1}{c|}{1592} & \multicolumn{1}{c|}{60.2} & 4.32M & 3.09M & 1.42 & \multicolumn{1}{c|}{1.55} & 0.10 \\ \hline
		\end{tabular}
		\label{tab:stats_A4b}
	\end{center}
\end{table}

The best results for ASL were obtained for transformations modeled as linear operations of the form $y=mx+b$. MSL and PSL performed similarly, with MSL learning faster in 3 scenarios, while PSL achieved a higher final performance in the last 2 scenarios. As for ASL, the transformation targets' standard deviation across 12 instances, averaged over all multipliers and biases, was [0.05,0.06,0.06,0.04,0.13,0.08] for each scenario, indicating that symmetry fitting was consistent in independent optimizations. 

A noteworthy result was that simultaneously training on all goals (\textbf{A4.1} and \textbf{A4.4}) did not appear to harm the final performance. This implies that in some cases, the performance on odd goals actually improved, in comparison with \textbf{A4.2} and \textbf{A4.5}. The scenario with lowest gain for ASL was \textbf{A4.6} in terms of episode return. This might be explained by the value distance plots, since its the only scenario where ASL does not take the lead, hinting that symmetry fitting did not sufficiently improve results in this configuration. Finally, the NSRR plots show an imbalance towards y=x except in \textbf{A4.3} and \textbf{A4.6}, since these do not contain any odd goals (where one front leg is required).

\bigskip
\noindent\textbf{A5}
\smallskip

\textbf{A5.1} and \textbf{A5.2} are equivalent to \textbf{A4.1} and \textbf{A4.4}, respectively, except that the policies could only train on the first 4 goals, having to generalize that knowledge to the remaining ones. In theory, the first 4 goals have enough information to learn the linear part of these modifications. However, when training with 4 goals, the policy explores only 4 different gaits, yielding a more biased linear transformation estimate, despite having lower average standard deviation, at [0.03,0.03] per scenario.

Regardless of the overall loss of performance in comparison with \textbf{A4}, ASL generates better policies than MSL or PSL, with performance gains between  37\% and 65\% (see Table~\ref{tab:stats_A5}). 

\begin{table}[t]
	\caption{Statistics for scenarios \textbf{A5.1} (top) and \textbf{A5.2} (bottom) extracted from average episode return (see Fig.~\ref{fig:res_return}) and average value distance (see Fig.~\ref{fig:res_valdiff}) }
	\footnotesize
	\begin{center}
		\begin{tabular}{|x{1.1cm}|ccc|x{1.6cm}|x{1.55cm}|cc|}
			\hline
			\multirow{3}{*}{\parbox{1cm}{\textbf{Algo-rithm}}} & \multicolumn{3}{x{4.0cm}|}{\textbf{Maximum Return \; (5pt window)}} & \textbf{TS when Return } & \textbf{Value Distance} & \multicolumn{2}{x{3.1cm}|}{\textbf{Value Distance (last 5pt window)}} \\ \cline{2-4} \cline{6-8} 
			& \multicolumn{1}{c|}{\textbf{Mean} $\boldsymbol{\mu}$} & \multicolumn{1}{c|}{\textbf{SD}} & \textbf{T.Step} & $\boldsymbol{\geq 0.9\mu}$ & \textbf{Mean} & \multicolumn{1}{c|}{\textbf{Mean}} & \textbf{SD} \\ \hline
			PPO & \multicolumn{1}{c|}{373} & \multicolumn{1}{c|}{12.6} & 4.38M & 3.33M & - & \multicolumn{1}{c|}{-} & - \\ \hline
			+ASL & \multicolumn{1}{c|}{1157} & \multicolumn{1}{c|}{16.9} & 4.44M & 2.78M & 2.23 & \multicolumn{1}{c|}{2.07} & 0.10 \\ \hline
			+PSL & \multicolumn{1}{c|}{847} & \multicolumn{1}{c|}{13.1} & 4.75M & 3.46M & 2.39 & \multicolumn{1}{c|}{2.40} & 0.09 \\ \hline
			+MSL & \multicolumn{1}{c|}{702} & \multicolumn{1}{c|}{17.6} & 4.26M & 2.96M & 2.16 & \multicolumn{1}{c|}{2.31} & 0.03 \\ \hline
		\end{tabular}
		\vspace{0.2cm}
		
		\begin{tabular}{|x{1.1cm}|ccc|x{1.6cm}|x{1.55cm}|cc|}
			\hline
			\multirow{3}{*}{\parbox{1cm}{\textbf{Algo-rithm}}} & \multicolumn{3}{x{4.0cm}|}{\textbf{Maximum Return \; (5pt window)}} & \textbf{TS when Return } & \textbf{Value Distance} & \multicolumn{2}{x{3.1cm}|}{\textbf{Value Distance (last 5pt window)}} \\ \cline{2-4} \cline{6-8} 
			& \multicolumn{1}{c|}{\textbf{Mean} $\boldsymbol{\mu}$} & \multicolumn{1}{c|}{\textbf{SD}} & \textbf{T.Step} & $\boldsymbol{\geq 0.9\mu}$ & \textbf{Mean} & \multicolumn{1}{c|}{\textbf{Mean}} & \textbf{SD} \\ \hline
			PPO & \multicolumn{1}{c|}{504} & \multicolumn{1}{c|}{16.7} & 4.69M & 3.02M & - & \multicolumn{1}{c|}{-} & - \\ \hline
			+ASL & \multicolumn{1}{c|}{1400} & \multicolumn{1}{c|}{28.7} & 4.32M & 2.90M & 2.25 & \multicolumn{1}{c|}{2.06} & 0.10 \\ \hline
			+PSL & \multicolumn{1}{c|}{1009} & \multicolumn{1}{c|}{18.1} & 4.75M & 3.46M & 2.57 & \multicolumn{1}{c|}{2.75} & 0.12 \\ \hline
			+MSL & \multicolumn{1}{c|}{940} & \multicolumn{1}{c|}{28.3} & 3.70M & 2.65M & 2.69 & \multicolumn{1}{c|}{3.06} & 0.17 \\ \hline
		\end{tabular}
		\label{tab:stats_A5}
	\end{center}
\end{table}

\bigskip
\noindent\textbf{A6}
\smallskip

In \textbf{A6} there are no feet mass modifications, but the floor surface is tilted 5 degrees in a specific direction. The results were underwhelming but also thought-provoking. For instance, the gait used to go down a small slope is different from the one used to go up, but we expected to find similarities that could be used to linearly model part of that relationship. In fact, the symmetry transformation targets are almost as stable as in \textbf{A4}, with an average standard deviation of [0.06,0.08,0.07,0.06,0.08,0.06]. Moreover, these transformations were not discarded by the policy, as demonstrated by the lower value distance in comparison with other methods. 

However, in general, the results were mostly on par with MSL and PSL, both in terms of learning speed and final accuracy (see Table~\ref{tab:stats_A6a} and Table~\ref{tab:stats_A6b}). 

\begin{table}[!t]
	\caption{Statistics for scenarios \textbf{A6.1} (top) to \textbf{A6.3} (bottom) extracted from average episode return (see Fig.~\ref{fig:res_return}) and average value distance (see Fig.~\ref{fig:res_valdiff}) }
	\vspace{-0.3cm}
	\footnotesize
	\begin{center}
		\begin{tabular}{|x{1.1cm}|ccc|x{1.6cm}|x{1.55cm}|cc|}
			\hline
			\multirow{3}{*}{\parbox{1cm}{\textbf{Algo-rithm}}} & \multicolumn{3}{x{4.0cm}|}{\textbf{Maximum Return \; (5pt window)}} & \textbf{TS when Return } & \textbf{Value Distance} & \multicolumn{2}{x{3.1cm}|}{\textbf{Value Distance (last 5pt window)}} \\ \cline{2-4} \cline{6-8} 
			& \multicolumn{1}{c|}{\textbf{Mean} $\boldsymbol{\mu}$} & \multicolumn{1}{c|}{\textbf{SD}} & \textbf{T.Step} & $\boldsymbol{\geq 0.9\mu}$ & \textbf{Mean} & \multicolumn{1}{c|}{\textbf{Mean}} & \textbf{SD} \\ \hline
			PPO & \multicolumn{1}{c|}{764} & \multicolumn{1}{c|}{14.9} & 3.21M & 2.78M & - & \multicolumn{1}{c|}{-} & - \\ \hline
			+ASL & \multicolumn{1}{c|}{1333} & \multicolumn{1}{c|}{24.3} & 4.20M & 2.04M & 2.55 & \multicolumn{1}{c|}{2.73} & 0.06 \\ \hline
			+PSL & \multicolumn{1}{c|}{1658} & \multicolumn{1}{c|}{44.1} & 4.88M & 3.09M & 2.90 & \multicolumn{1}{c|}{3.10} & 0.08 \\ \hline
			+MSL & \multicolumn{1}{c|}{1442} & \multicolumn{1}{c|}{20.0} & 3.64M & 2.35M & 2.90 & \multicolumn{1}{c|}{3.36} & 0.10 \\ \hline
		\end{tabular}
		\vspace{0.2cm}
		
		\begin{tabular}{|x{1.1cm}|ccc|x{1.6cm}|x{1.55cm}|cc|}
			\hline
			\multirow{3}{*}{\parbox{1cm}{\textbf{Algo-rithm}}} & \multicolumn{3}{x{4.0cm}|}{\textbf{Maximum Return \; (5pt window)}} & \textbf{TS when Return } & \textbf{Value Distance} & \multicolumn{2}{x{3.1cm}|}{\textbf{Value Distance (last 5pt window)}} \\ \cline{2-4} \cline{6-8} 
			& \multicolumn{1}{c|}{\textbf{Mean} $\boldsymbol{\mu}$} & \multicolumn{1}{c|}{\textbf{SD}} & \textbf{T.Step} & $\boldsymbol{\geq 0.9\mu}$ & \textbf{Mean} & \multicolumn{1}{c|}{\textbf{Mean}} & \textbf{SD} \\ \hline
			PPO & \multicolumn{1}{c|}{1042} & \multicolumn{1}{c|}{33.8} & 2.72M & 2.16M & - & \multicolumn{1}{c|}{-} & - \\ \hline
			+ASL & \multicolumn{1}{c|}{2033} & \multicolumn{1}{c|}{62.5} & 4.88M & 3.46M & 1.76 & \multicolumn{1}{c|}{1.88} & 0.02 \\ \hline
			+PSL & \multicolumn{1}{c|}{1904} & \multicolumn{1}{c|}{47.4} & 4.44M & 3.02M & 2.54 & \multicolumn{1}{c|}{2.53} & 0.08 \\ \hline
			+MSL & \multicolumn{1}{c|}{1879} & \multicolumn{1}{c|}{59.9} & 4.51M & 2.72M & 2.73 & \multicolumn{1}{c|}{3.32} & 0.13 \\ \hline
		\end{tabular}
		\vspace{0.2cm}
		
		\begin{tabular}{|x{1.1cm}|ccc|x{1.6cm}|x{1.55cm}|cc|}
			\hline
			\multirow{3}{*}{\parbox{1cm}{\textbf{Algo-rithm}}} & \multicolumn{3}{x{4.0cm}|}{\textbf{Maximum Return \; (5pt window)}} & \textbf{TS when Return } & \textbf{Value Distance} & \multicolumn{2}{x{3.1cm}|}{\textbf{Value Distance (last 5pt window)}} \\ \cline{2-4} \cline{6-8} 
			& \multicolumn{1}{c|}{\textbf{Mean} $\boldsymbol{\mu}$} & \multicolumn{1}{c|}{\textbf{SD}} & \textbf{T.Step} & $\boldsymbol{\geq 0.9\mu}$ & \textbf{Mean} & \multicolumn{1}{c|}{\textbf{Mean}} & \textbf{SD} \\ \hline
			PPO & \multicolumn{1}{c|}{1097} & \multicolumn{1}{c|}{35.4} & 3.09M & 2.16M & - & \multicolumn{1}{c|}{-} & - \\ \hline
			+ASL & \multicolumn{1}{c|}{2098} & \multicolumn{1}{c|}{51.8} & 4.57M & 2.47M & 2.09 & \multicolumn{1}{c|}{2.32} & 0.07 \\ \hline
			+PSL & \multicolumn{1}{c|}{2158} & \multicolumn{1}{c|}{39.8} & 4.69M & 3.09M & 2.41 & \multicolumn{1}{c|}{2.32} & 0.12 \\ \hline
			+MSL & \multicolumn{1}{c|}{1782} & \multicolumn{1}{c|}{31.7} & 4.44M & 2.84M & 2.95 & \multicolumn{1}{c|}{3.51} & 0.17 \\ \hline
		\end{tabular}
		\label{tab:stats_A6a}
	\end{center}
	\vspace{-0.4cm}
\end{table}

\begin{table}[!h]
	\caption{Statistics for scenarios \textbf{A6.4} (top) to \textbf{A6.6} (bottom) extracted from average episode return (see Fig.~\ref{fig:res_return}) and average value distance (see Fig.~\ref{fig:res_valdiff}) }
	\vspace{-0.3cm}
	\footnotesize
	\begin{center}
		\begin{tabular}{|x{1.1cm}|ccc|x{1.6cm}|x{1.55cm}|cc|}
			\hline
			\multirow{3}{*}{\parbox{1cm}{\textbf{Algo-rithm}}} & \multicolumn{3}{x{4.0cm}|}{\textbf{Maximum Return \; (5pt window)}} & \textbf{TS when Return } & \textbf{Value Distance} & \multicolumn{2}{x{3.1cm}|}{\textbf{Value Distance (last 5pt window)}} \\ \cline{2-4} \cline{6-8} 
			& \multicolumn{1}{c|}{\textbf{Mean} $\boldsymbol{\mu}$} & \multicolumn{1}{c|}{\textbf{SD}} & \textbf{T.Step} & $\boldsymbol{\geq 0.9\mu}$ & \textbf{Mean} & \multicolumn{1}{c|}{\textbf{Mean}} & \textbf{SD} \\ \hline
			PPO & \multicolumn{1}{c|}{706} & \multicolumn{1}{c|}{20.7} & 3.83M & 3.15M & - & \multicolumn{1}{c|}{-} & - \\ \hline
			+ASL & \multicolumn{1}{c|}{1379} & \multicolumn{1}{c|}{17.9} & 4.32M & 2.41M & 2.47 & \multicolumn{1}{c|}{2.81} & 0.09 \\ \hline
			+PSL & \multicolumn{1}{c|}{1589} & \multicolumn{1}{c|}{49.8} & 4.88M & 3.15M & 2.94 & \multicolumn{1}{c|}{3.10} & 0.12 \\ \hline
			+MSL & \multicolumn{1}{c|}{1485} & \multicolumn{1}{c|}{43.8} & 4.20M & 3.02M & 3.05 & \multicolumn{1}{c|}{3.56} & 0.12 \\ \hline
		\end{tabular}
		\vspace{0.2cm}
		
		\begin{tabular}{|x{1.1cm}|ccc|x{1.6cm}|x{1.55cm}|cc|}
			\hline
			\multirow{3}{*}{\parbox{1cm}{\textbf{Algo-rithm}}} & \multicolumn{3}{x{4.0cm}|}{\textbf{Maximum Return \; (5pt window)}} & \textbf{TS when Return } & \textbf{Value Distance} & \multicolumn{2}{x{3.1cm}|}{\textbf{Value Distance (last 5pt window)}} \\ \cline{2-4} \cline{6-8} 
			& \multicolumn{1}{c|}{\textbf{Mean} $\boldsymbol{\mu}$} & \multicolumn{1}{c|}{\textbf{SD}} & \textbf{T.Step} & $\boldsymbol{\geq 0.9\mu}$ & \textbf{Mean} & \multicolumn{1}{c|}{\textbf{Mean}} & \textbf{SD} \\ \hline
			PPO & \multicolumn{1}{c|}{993} & \multicolumn{1}{c|}{27.0} & 2.96M & 2.59M & - & \multicolumn{1}{c|}{-} & - \\ \hline
			+ASL & \multicolumn{1}{c|}{1479} & \multicolumn{1}{c|}{41.0} & 3.70M & 2.10M & 1.88 & \multicolumn{1}{c|}{2.22} & 0.05 \\ \hline
			+PSL & \multicolumn{1}{c|}{1930} & \multicolumn{1}{c|}{56.3} & 4.88M & 3.21M & 2.43 & \multicolumn{1}{c|}{2.45} & 0.09 \\ \hline
			+MSL & \multicolumn{1}{c|}{1659} & \multicolumn{1}{c|}{39.1} & 4.38M & 2.35M & 2.45 & \multicolumn{1}{c|}{2.94} & 0.06 \\ \hline
		\end{tabular}
		\vspace{0.2cm}
		
		\begin{tabular}{|x{1.1cm}|ccc|x{1.6cm}|x{1.55cm}|cc|}
			\hline
			\multirow{3}{*}{\parbox{1cm}{\textbf{Algo-rithm}}} & \multicolumn{3}{x{4.0cm}|}{\textbf{Maximum Return \; (5pt window)}} & \textbf{TS when Return } & \textbf{Value Distance} & \multicolumn{2}{x{3.1cm}|}{\textbf{Value Distance (last 5pt window)}} \\ \cline{2-4} \cline{6-8} 
			& \multicolumn{1}{c|}{\textbf{Mean} $\boldsymbol{\mu}$} & \multicolumn{1}{c|}{\textbf{SD}} & \textbf{T.Step} & $\boldsymbol{\geq 0.9\mu}$ & \textbf{Mean} & \multicolumn{1}{c|}{\textbf{Mean}} & \textbf{SD} \\ \hline
			PPO & \multicolumn{1}{c|}{933} & \multicolumn{1}{c|}{18.6} & 3.77M & 2.35M & - & \multicolumn{1}{c|}{-} & - \\ \hline
			+ASL & \multicolumn{1}{c|}{2072} & \multicolumn{1}{c|}{47.2} & 4.26M & 2.47M & 2.27 & \multicolumn{1}{c|}{2.53} & 0.10 \\ \hline
			+PSL & \multicolumn{1}{c|}{1965} & \multicolumn{1}{c|}{36.8} & 4.88M & 3.58M & 3.08 & \multicolumn{1}{c|}{3.19} & 0.05 \\ \hline
			+MSL & \multicolumn{1}{c|}{1804} & \multicolumn{1}{c|}{50.9} & 4.51M & 2.84M & 3.05 & \multicolumn{1}{c|}{3.52} & 0.06 \\ \hline
		\end{tabular}
		\label{tab:stats_A6b}
	\end{center}
\end{table}

Two potential conclusions can be derived from this outcome. First, the nonlinear part of the symmetric relation can monopolize the optimization, leaving most of the adaptation work to PPO. Second, the benefit of adapting the symmetry transformations in \textbf{A6} might be canceled out by the additional variance introduced by modifying such transformations during optimization.

\clearpage
\section{Conclusion}

There are multiple approaches to model minimization that are concerned with reducing redundant symmetries. Loss functions are the most versatile way of achieving this objective, due to their ability to represent complex concepts but also because they can directly control the optimization direction by producing a gradient. For actor-critic methods such as PPO, their loss functions can be extended to introduce symmetry concepts that will influence the optimization of the policy and value function.

This work presents two existing symmetry loss functions --- Mirror Symmetry Loss (MSL) and Proximal Symmetry Loss (PSL) --- extending them with value losses and, for MSL, the capacity to handle involutory transformations, such as most rotations. A novel Adaptive Symmetry Learning (ASL) method is presented, where the main focus is to handle incomplete or inexact symmetry descriptions by learning adaptations. There are multiple sources for potential imperfections, including perturbations in the robot, environment, reward function, or agent (control process).

ASL is composed of a symmetry fitting component and a modular loss function with useful gates to exclude neutral states and disadvantageous updates. While the policy is learning the best action for each state, ASL's loss function is enforcing a common symmetric linear relation across all states and, at the same time, adapting that same relation to match what the policy is learning. Adapting nonlinear relations is also an option, although it can be counterproductive due to the additional learning time. In the end, it is a trade-off between the convergence efficiency of symmetry transformations and the policy itself.

The case study presented in this work explored a four-legged ant model adapted from PyBullet's AntBulletEnv-v0. This is an interesting model to learn multidirectional locomotion tasks as it admits 7 non-identity symmetry transformations --- 4 reflections and 3 rotations. In all the proposed scenarios, vanilla PPO was not enough to learn solid policies. By contrast, PPO+ASL was able to equal or outperform the alternative symmetry-enhanced methods in most situations.

The main strengths that differentiate ASL are its performance in recovering from large linearly modelable perturbations, and generalizing knowledge to hidden symmetric states. The former characteristic was observed in a controlled setting (scenario \textbf{A3}) and in a realistic context (\textbf{A4}). The ability to successfully cope with hidden symmetric states was verified in configurations with symmetry fitting (\textbf{A5}) and without it (\textbf{A1.2}).

ASL was comparable to MSL and PSL in perfectly symmetric and fully observable conditions (\textbf{A1.1}), and slightly imperfect models (\textbf{A2}). Finally, in an inclined surface (\textbf{A6}), ASL exhibited a mixed performance, producing consistent adaptations but not translating that into better episode returns. In conclusion, the advantage of using ASL in relation to other symmetry loss functions depends on the context. However, in general, its efficiency is at least as good as the other methods, being in some circumstances considerably better.

Future research directions encompass automating hyperparameter optimization through meta-learning to mitigate setup complexity, enabling faster results with limited computational resources. Additionally, exploring symmetry perturbations that deteriorate over time as a result of wear and tear on robot components constitutes another promising avenue of investigation. While ASL can accommodate dynamic symmetry relationships without getting stuck in local minima, alternative approaches to PPO should be explored for handling changing environments.

Within the broader context of reinforcement learning, we focused on model-free methods due to the complexity and unpredictability of realistic robotic tasks, where model-based approaches often fall short due to oversimplified environment models. Policy optimization techniques like PPO are particularly suitable, as they allow direct manipulation of the policy, providing precise control over symmetry enforcement during learning.

ASL’s framework aligns with the actor-critic structure by modifying both the policy and value function through its extended loss function, making it straightforward to adapt to methods like A2C \cite{Mnih2016_A2C_A3C} and TRPO \cite{Schulman2015_TRPO}. However, while ASL’s loss function harmonizes well with the trust regions of PPO and TRPO, it may not harmonize with A2C, potentially affecting performance and control precision. For off-policy algorithms like SAC \cite{Haarnoja2018_SAC}, the same problem could occur, but ASL's symmetry fitting, which extracts actions from the current policy rather than past experiences, would still work effectively with observations from previous policies.

While value-based methods are theoretically compatible with ASL, they would require significant adaptations. The current approach focuses primarily on controlling the policy, with less emphasis on the value function. Extending ASL to methods like DQN \cite{mnih2015DQN} would require rethinking how symmetry is applied to the value function.

\section*{CRediT authorship contribution statement}

\textbf{Miguel Abreu:} Conceptualization, Investigation, Methodology, Validation, Software, Formal analysis, Visualization, Writing - Original Draft. \textbf{Lu\' {i}s Paulo Reis:} Writing - Review \& Editing, Supervision, Funding acquisition. \textbf{Nuno Lau:} Writing - Review \& Editing, Supervision, Funding acquisition. 

\section*{Declaration of competing interest}
	
The authors declare that they have no known competing financial interests or personal relationships that could have appeared to influence the work reported in this paper.

\section*{Data availability}

The algorithm implementations and environment configurations evaluated in this work are fully available at \url{https://github.com/m-abr/Adaptive-Symmetry-Learning}.

\section*{Acknowledgements}

The first author is supported by FCT --- Foundation for Science and Technology under grant SFRH/BD/139926/2018. This work was financially supported by FCT/MCTES (PIDDAC), under projects UIDB/00027/2020 (LIACC) and UIDB/00127/2020 (IEETA).

%\clearpage % flushes all floats before proceeding

\bibliographystyle{elsarticle-num}
\bibliography{bibliography}

\begin{thebibliography}{10}
\expandafter\ifx\csname url\endcsname\relax
  \def\url#1{\texttt{#1}}\fi
\expandafter\ifx\csname urlprefix\endcsname\relax\def\urlprefix{URL }\fi
\expandafter\ifx\csname href\endcsname\relax
  \def\href#1#2{#2} \def\path#1{#1}\fi

\bibitem{weyl}
H.~Weyl, Symmetry, Princeton University Press, Princeton, 1952.

\bibitem{mainzer2005}
K.~Mainzer, Symmetry and complexity: The spirit and beauty of nonlinear
  science, Vol.~51, World Scientific, 2005.

\bibitem{handedness}
M.~Papadatou-Pastou, E.~Ntolka, J.~Schmitz, M.~Martin, M.~R. Munaf{\`o},
  S.~Ocklenburg, S.~Paracchini, Human handedness: A meta-analysis,
  Psychological bulletin 146~(6) (2020) 481--524.

\bibitem{lin1992}
L.-J. Lin, Self-improving reactive agents based on reinforcement learning,
  planning and teaching, Machine learning 8~(3-4) (1992) 293--321.

\bibitem{browne21}
M.~G. Browne, C.~S. Smock, R.~T. Roemmich, The human preference for symmetric
  walking often disappears when one leg is constrained, The Journal of
  Physiology 599~(4) (2021) 1243--1260.

\bibitem{handvzic15}
I.~Hand{\v{z}}i{\'c}, K.~B. Reed, Perception of gait patterns that deviate from
  normal and symmetric biped locomotion, Frontiers in psychology 6 (2015).

\bibitem{yu2018learning}
W.~Yu, G.~Turk, C.~K. Liu, Learning symmetric and low-energy locomotion, ACM
  Transactions on Graphics 37~(4) (July 2018).

\bibitem{psl}
M.~Kasaei, M.~Abreu, N.~Lau, A.~Pereira, L.~P. Reis, A {CPG}-based agile and
  versatile locomotion framework using {Proximal Symmetry Loss}, arXiv preprint
  arXiv:2103.00928 (2021).

\bibitem{schulman2017ppo}
J.~Schulman, F.~Wolski, P.~Dhariwal, A.~Radford, O.~Klimov, Proximal policy
  optimization algorithms, arXiv preprint arXiv:1707.06347 (2017).

\bibitem{ravindran2001}
B.~Ravindran, A.~G. Barto, Symmetries and model minimization in {Markov}
  decision processes, Tech. rep., University of Massachusetts, USA (2001).

\bibitem{Zinkevich}
M.~Zinkevich, T.~R. Balch, Symmetry in {Markov} decision processes and its
  implications for single agent and multiagent learning, in: Proceedings of the
  Eighteenth International Conference on Machine Learning, ICML '01, Morgan
  Kaufmann Publishers Inc., San Francisco, CA, USA, 2001, p. 632.

\bibitem{agostini2009exploiting}
A.~Agostini, E.~Celaya, Exploiting domain symmetries in reinforcement learning
  with continuous state and action spaces, in: 2009 International Conference on
  Machine Learning and Applications, IEEE, 2009, pp. 331--336.

\bibitem{Zeng21sym}
K.~Zeng, M.~D. Graham, Symmetry reduction for deep reinforcement learning
  active control of chaotic spatiotemporal dynamics, Physical Review E 104
  (July 2021).

\bibitem{ildefonso21}
P.~Ildefonso, P.~Rem{\'e}dios, R.~Silva, M.~Vasco, F.~S. Melo, A.~Paiva,
  M.~Veloso, Exploiting symmetry in human robot-assisted dressing using
  reinforcement learning, in: Progress in Artificial Intelligence: 20th EPIA
  Conference on Artificial Intelligence, Vol. 12981 of Lecture Notes in
  Computer Science, Springer, 2021, pp. 405--417.

\bibitem{surovik2019adaptive}
D.~Surovik, K.~Wang, M.~Vespignani, J.~Bruce, K.~E. Bekris, Adaptive tensegrity
  locomotion: Controlling a compliant icosahedron with symmetry-reduced
  reinforcement learning, The International Journal of Robotics Research
  (2019).

\bibitem{xie2020learning}
Z.~Xie, P.~Clary, J.~Dao, P.~Morais, J.~Hurst, M.~van~de Panne, Learning
  locomotion skills for {Cassie}: Iterative design and sim-to-real, in:
  Proceedings of the Conference on Robot Learning, Vol. 100 of Proceedings of
  Machine Learning Research, PMLR, 2020, pp. 317--329.

\bibitem{hereid2018dynamic}
A.~Hereid, C.~M. Hubicki, E.~A. Cousineau, A.~D. Ames, Dynamic humanoid
  locomotion: A scalable formulation for {HZD} gait optimization, IEEE
  Transactions on Robotics 34~(2) (2018) 370--387.

\bibitem{peng2017deeploco}
X.~B. Peng, G.~Berseth, K.~Yin, M.~van~de Panne, Deeploco: Dynamic locomotion
  skills using hierarchical deep reinforcement learning, ACM Transactions on
  Graphics (Proc. SIGGRAPH 2017) 36~(4) (2017).

\bibitem{bree2021data}
R.~van Bree, Data augmentation for regularizing learned world models in
  reinforcement learning, Master's thesis, University of Twente (2021).

\bibitem{lin2020}
Y.~Lin, J.~Huang, M.~Zimmer, Y.~Guan, J.~Rojas, P.~Weng, Invariant transform
  experience replay: Data augmentation for deep reinforcement learning, IEEE
  Robotics and Automation Letters 5~(4) (2020) 6615--6622.

\bibitem{abdolhosseini2019learning}
F.~Abdolhosseini, H.~Y. Ling, Z.~Xie, X.~B. Peng, M.~van~de Panne, On learning
  symmetric locomotion, in: Proceedings of the 12th ACM SIGGRAPH Conference on
  Motion, Interaction and Games, ACM, 2019.

\bibitem{mishra2019augmenting}
S.~Mishra, A.~Abdolmaleki, A.~Guez, P.~Trochim, D.~Precup, Augmenting learning
  using symmetry in a biologically-inspired domain, arXiv preprint
  arXiv:1910.00528 (2019).

\bibitem{silver2016go}
D.~Silver, A.~Huang, C.~J. Maddison, A.~Guez, L.~Sifre, G.~Van Den~Driessche,
  J.~Schrittwieser, I.~Antonoglou, V.~Panneershelvam, M.~Lanctot, et~al.,
  Mastering the game of {Go} with deep neural networks and tree search, Nature
  529~(7587) (2016) 484--489.

\bibitem{Pol2020mdp}
E.~van~der Pol, D.~E. Worrall, H.~van Hoof, F.~A. Oliehoek, M.~Welling, {MDP}
  homomorphic networks: Group symmetries in reinforcement learning, in:
  Proceedings of the 34th International Conference on Neural Information
  Processing Systems, 2020, pp. 4199--4210.

\bibitem{encoding22}
A.~Bhattacharya, M.~Mattheakis, P.~Protopapas, Encoding involutory invariances
  in neural networks, in: 2022 International Joint Conference on Neural
  Networks (IJCNN), 2022.

\bibitem{mondal20}
A.~K. Mondal, P.~Nair, K.~Siddiqi, Group equivariant deep reinforcement
  learning, arXiv preprint arXiv:2007.03437.\,Presented at the ICML 2020
  Workshop on Inductive Biases, Invariances and Generalization in RL (2020).

\bibitem{rav2017}
S.~Ravanbakhsh, J.~Schneider, B.~P{\'o}czos, Equivariance through
  parameter-sharing, in: Proceedings of the 34th International Conference on
  Machine Learning, Vol.~70 of Proceedings of Machine Learning Research, PMLR,
  2017, pp. 2892--2901.

\bibitem{san2019}
A.~Sannai, Y.~Takai, M.~Cordonnier, Universal approximations of permutation
  invariant/equivariant functions by deep neural networks, arXiv preprint
  arXiv:1903.01939 (2019).

\bibitem{cohen2016}
T.~Cohen, M.~Welling, Group equivariant convolutional networks, in: Proceedings
  of The 33rd International Conference on Machine Learning, Vol.~48 of
  Proceedings of Machine Learning Research, PMLR, 2016, pp. 2990--2999.

\bibitem{Mahajan17_abstract}
A.~Mahajan, T.~Tulabandhula, Symmetry detection and exploitation for function
  approximation in deep {RL}, in: Proceedings of the 16th Conference on
  Autonomous Agents and MultiAgent Systems, AAMAS '17, International Foundation
  for Autonomous Agents and Multiagent Systems, 2017, p. 1619–1621.

\bibitem{mahajan2017symmetry}
A.~Mahajan, T.~Tulabandhula, Symmetry learning for function approximation in
  reinforcement learning, arXiv preprint arXiv:1706.02999 (2017).

\bibitem{watkins1992q}
C.~J. C.~H. Watkins, P.~Dayan, Q-learning, Machine learning 8~(3) (1992)
  279--292.

\bibitem{openAIbaselines}
P.~Dhariwal, C.~Hesse, O.~Klimov, A.~Nichol, M.~Plappert, A.~Radford,
  J.~Schulman, S.~Sidor, Y.~Wu, P.~Zhokhov, {OpenAI} baselines,
  \url{https://github.com/openai/baselines} (2017).

\bibitem{stable_baselines}
A.~Raffin, A.~Hill, A.~Gleave, A.~Kanervisto, M.~Ernestus, N.~Dormann,
  {Stable-Baselines3}: Reliable reinforcement learning implementations (2021).

\bibitem{McWeeny}
R.~McWeeny, Symmetry: an introduction to group theory and its applications,
  Pergamon Press, MacMillan, New York, 1963.

\bibitem{lectures1917}
F.~M. Jaeger, Lectures on the principle of symmetry and its applications in all
  natural sciences, Elsevier, Amsterdam, 1917.

\bibitem{statsdict}
B.~S. Everitt, A.~Skrondal, The Cambridge dictionary of statistics, 4th
  Edition, Cambridge University Press, Cambridge, UK, 2010.

\bibitem{trpo}
J.~Schulman, S.~Levine, P.~Abbeel, M.~Jordan, P.~Moritz, Trust region policy
  optimization, in: Proceedings of the 32nd International Conference on Machine
  Learning, Vol.~37 of Proceedings of Machine Learning Research, PMLR, 2015,
  pp. 1889--1897.

\bibitem{kingma2014adam}
D.~P. Kingma, J.~Ba, Adam: {A} method for stochastic optimization, in:
  Y.~Bengio, Y.~LeCun (Eds.), 3rd International Conference on Learning
  Representations ({ICLR}), San Diego, CA, USA, 2015.

\bibitem{coumans2020}
E.~Coumans, Y.~Bai, {PyBullet}, a {Python} module for physics simulation for
  games, robotics and machine learning, \url{http://pybullet.org} (2016--2024).

\bibitem{ppo_implementation}
E.~Logan, I.~Andrew, S.~Shibani, T.~Dimitris, J.~Firdaus, R.~Larry,
  M.~Aleksander, Implementation matters in deep {RL: A case study on PPO and
  TRPO}, in: International Conference on Learning Representations (ICLR), 2019.

\bibitem{ppo_hypers}
M.~Andrychowicz, A.~Raichuk, P.~Sta{\'n}czyk, M.~Orsini, S.~Girgin,
  R.~Marinier, L.~Hussenot, ..., O.~Bachem, What matters for on-policy deep
  actor-critic methods? a large-scale study, in: International Conference on
  Learning Representations (ICLR), 2021.

\bibitem{rl-zoo3}
A.~Raffin, {RL} baselines3 zoo,
  \url{https://github.com/DLR-RM/rl-baselines3-zoo} (2020).

\bibitem{Mnih2016_A2C_A3C}
V.~Mnih, A.~P. Badia, M.~Mirza, A.~Graves, T.~Lillicrap, T.~Harley, D.~Silver,
  K.~Kavukcuoglu, Asynchronous methods for deep reinforcement learning, in:
  International conference on machine learning, Vol.~48, 2016, pp. 1928--1937.

\bibitem{Schulman2015_TRPO}
J.~Schulman, S.~Levine, P.~Abbeel, M.~Jordan, P.~Moritz, Trust region policy
  optimization, in: Proceedings of the 32nd International Conference on Machine
  Learning, Vol.~37 of Proceedings of Machine Learning Research, PMLR, 2015,
  pp. 1889--1897.

\bibitem{Haarnoja2018_SAC}
T.~Haarnoja, A.~Zhou, P.~Abbeel, S.~Levine, Soft actor-critic: Off-policy
  maximum entropy deep reinforcement learning with a stochastic actor, in:
  International Conference on Machine Learning, PMLR, 2018, pp. 1861--1870.

\bibitem{mnih2015DQN}
V.~Mnih, K.~Kavukcuoglu, D.~Silver, A.~A. Rusu, J.~Veness, M.~G. Bellemare,
  A.~Graves, M.~Riedmiller, A.~K. Fidjeland, G.~Ostrovski, et~al., Human-level
  control through deep reinforcement learning, Nature 518~(7540) (2015)
  529--533.

\end{thebibliography}

\clearpage % flushes all floats before proceeding

\appendix

\section{Symmetry transformations}\label{app:A}

\begin{figure}[h!]
	\includegraphics[width=\columnwidth]{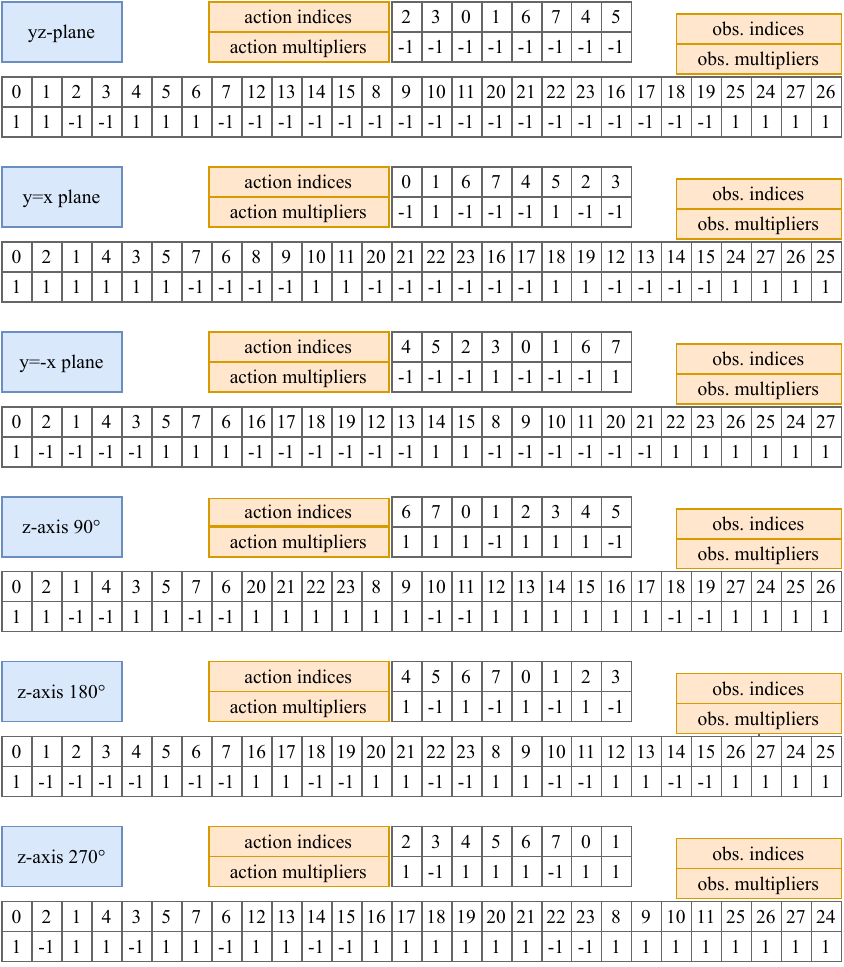}
	\caption{Remaining symmetry transformation for AntBulletEnv-v0, as a complement to Fig. \ref{fig:ant_sym_1}.}
	\label{fig:ant_sym_2}
\end{figure}

\clearpage % flushes all floats before proceeding

\section{Evaluation hyperparameters}\label{app:B}
\begin{table}[ht]
	\caption{Scenario \textbf{A1.1} --- ASL Hyperparameters}
	\begin{center}
		\begin{tabular}{|c|r|l|c|}
			\hline
			\textbf{Method} & \textbf{Symbol} & \textbf{Short Description} & \textbf{Value} \\ \hline
			all & TS & total time steps & \num{4e6} \\ \hline
			ASL & $\text{\textit{\textbf{w}}}_\pi$ & policy sym. loss weight & 0.05  \\ \hline
			ASL & $\text{H}_\text{U}$, $\text{H}_\text{G}$, $\boldsymbol{\hat{g}}(a)$ form  & 
			symmetry fitting & Disabled  \\ \hline
			ASL & $\boldsymbol{k}_\text{s}$ & max. distribution shift & 0.3 \\ \hline
			ASL & $\boldsymbol{k}_\text{d}$[reflec] & dead zone (sym. plane) & 0.1 \\ \hline
			MSL & $\text{\textit{\textbf{w}}}_\pi$ & policy sym. loss weight & 10  \\ \hline
			PSL & $\text{\textit{\textbf{w}}}_\pi$ & policy sym. loss weight & 0.008  \\ \hline
		\end{tabular}
		\label{tab_asl:asl_hyperparameters_A1.1}
	\end{center}
\end{table}

\begin{table}[ht]
	\caption{Scenario \textbf{A1.2} --- ASL Hyperparameters}
	\begin{center}
		\begin{tabular}{|c|r|l|c|}
			\hline
			\textbf{Method} & \textbf{Symbol} & \textbf{Short Description} & \textbf{Value} \\ \hline
			all & TS & total time steps & \num{4e6} \\ \hline
			ASL & $\text{\textit{\textbf{w}}}_\pi$ & policy sym. loss weight & 0.25  \\ \hline
			ASL & $\text{H}_\text{U}$, $\text{H}_\text{G}$, $\boldsymbol{\hat{g}}(a)$ form  & 
			symmetry fitting & Disabled  \\ \hline
			ASL & $\boldsymbol{k}_\text{s}$ & max. distribution shift & 1 \\ \hline
			ASL & $\boldsymbol{k}_\text{d}$[reflec] & dead zone (sym. plane) & 0.2 \\ \hline
			ASL & $\boldsymbol{k}_\text{v}$ & value gate factor & Disabled  \\ \hline
			MSL & $\text{\textit{\textbf{w}}}_\pi$ & policy sym. loss weight & 10  \\ \hline
			PSL & $\text{\textit{\textbf{w}}}_\pi$ & policy sym. loss weight & 0.08  \\ \hline
		\end{tabular}
		\label{tab_asl:asl_hyperparameters_A1.2}
	\end{center}
\end{table}

\begin{table}[!ht]
	\caption{Scenario \textbf{A2.1} --- ASL Hyperparameters}
	\begin{center}
		\begin{tabular}{|c|r|l|c|}
			\hline
			\textbf{Method} & \textbf{Symbol} & \textbf{Short Description} & \textbf{Value} \\ \hline
			all & TS & total time steps & \num{4e6} \\ \hline
			ASL & $\text{\textit{\textbf{w}}}_\pi$ & policy sym. loss weight & 0.05  \\ 
			\hline
			ASL & $\boldsymbol{k}_\text{s}$ & max. distribution shift & 0.3 \\ \hline
			ASL & $\boldsymbol{k}_\text{d}$[reflec] & dead zone (sym. plane) & 0.1 \\ \hline
			ASL & $\boldsymbol{\hat{g}}(a)$ form & function form of $\boldsymbol{g}(a)$ & $y=mx$ \\ \hline
			MSL & $\text{\textit{\textbf{w}}}_\pi$ & policy sym. loss weight & 3  \\ \hline
			PSL & $\text{\textit{\textbf{w}}}_\pi$ & policy sym. loss weight & 0.003  \\ \hline
		\end{tabular}
		\label{tab_asl:asl_hyperparameters_A2.1}
	\end{center}
\end{table}

\begin{table}[ht]
	\caption{Scenario \textbf{A2.2} --- ASL Hyperparameters}
	\begin{center}
		\begin{tabular}{|c|r|l|c|}
			\hline
			\textbf{Method} & \textbf{Symbol} & \textbf{Short Description} & \textbf{Value} \\ \hline
			all & TS & total time steps & \num{4e6} \\ \hline
			ASL & $\text{\textit{\textbf{w}}}_\pi$ & policy sym. loss weight & 0.05  \\ 
			\hline
			ASL & $\boldsymbol{k}_\text{s}$ & max. distribution shift & 0.2 \\ \hline
			ASL & $\boldsymbol{k}_\text{d}$[reflec] & dead zone (sym. plane) & 0.1 \\ \hline
			ASL & $\boldsymbol{\hat{g}}(a)$ form & function form of $\boldsymbol{g}(a)$ & $y=mx$ \\ \hline
			MSL & $\text{\textit{\textbf{w}}}_\pi$ & policy sym. loss weight & 2  \\ \hline
			PSL & $\text{\textit{\textbf{w}}}_\pi$ & policy sym. loss weight & 0.001  \\ \hline
		\end{tabular}
		\label{tab_asl:asl_hyperparameters_A2.2}
	\end{center}
\end{table}

\begin{table}[ht]
	\caption{Scenario \textbf{A3.1} --- ASL Hyperparameters}
	\begin{center}
		\begin{tabular}{|c|r|l|c|}
			\hline
			\textbf{Method} & \textbf{Symbol} & \textbf{Short Description} & \textbf{Value} \\ \hline
			all & TS & total time steps & \num{5e6} \\ \hline
			ASL & $\text{\textit{\textbf{w}}}_\pi$ & policy sym. loss weight & 0.1  \\ 
			\hline
			ASL & $\boldsymbol{k}_\text{s}$ & max. distribution shift & 0.5 \\ \hline
			ASL & $\boldsymbol{k}_\text{d}$[reflec] & dead zone (sym. plane) & 0.1 \\ \hline
			ASL & $\boldsymbol{\hat{g}}(a)$ form & function form of $\boldsymbol{g}(a)$ & $y=mx$ \\ \hline
			MSL & $\text{\textit{\textbf{w}}}_\pi$ & policy sym. loss weight & 1  \\ \hline
			PSL & $\text{\textit{\textbf{w}}}_\pi$ & policy sym. loss weight & 0.001  \\ \hline
		\end{tabular}
		\label{tab_asl:asl_hyperparameters_A3.1}
	\end{center}
\end{table}

\begin{table}[ht]
	\caption{Scenario \textbf{A3.2} --- ASL Hyperparameters}
	\begin{center}
		\begin{tabular}{|c|r|l|c|}
			\hline
			\textbf{Method} & \textbf{Symbol} & \textbf{Short Description} & \textbf{Value} \\ \hline
			all & TS & total time steps & \num{5e6} \\ \hline
			ASL & $\text{\textit{\textbf{w}}}_\pi$ & policy sym. loss weight & 0.1  \\ 
			\hline
			ASL & $\boldsymbol{k}_\text{s}$ & max. distribution shift & 0.25 \\ \hline
			ASL & $\boldsymbol{k}_\text{d}$[reflec] & dead zone (sym. plane) & 0.1 \\ \hline
			ASL & $\boldsymbol{\hat{g}}(a)$ form & function form of $\boldsymbol{g}(a)$ & $y=mx$ \\ \hline
			MSL & $\text{\textit{\textbf{w}}}_\pi$ & policy sym. loss weight & 0.1  \\ 
			\hline
			PSL & $\text{\textit{\textbf{w}}}_\pi$ & policy sym. loss weight & 0.001  \\ \hline
		\end{tabular}
		\label{tab_asl:asl_hyperparameters_A3.2}
	\end{center}
\end{table}

\begin{table}[ht]
	\caption{Scenario \textbf{A4.1} --- ASL Hyperparameters}
	\begin{center}
		\begin{tabular}{|c|r|l|c|}
			\hline
			\textbf{Method} & \textbf{Symbol} & \textbf{Short Description} & \textbf{Value} \\ \hline
			all & TS & total time steps & \num{5e6} \\ \hline
			ASL & $\text{\textit{\textbf{w}}}_\pi$ & policy sym. loss weight & $0.1$  \\ 
			\hline
			ASL & $\boldsymbol{k}_\text{s}$ & max. distribution shift & 0.1 \\ \hline
			ASL & $\boldsymbol{k}_\text{d}$[reflec] & dead zone (sym. plane) & 0.2 \\ \hline
			ASL & $\boldsymbol{\hat{g}}(a)$ form & function form of $\boldsymbol{g}(a)$ & $y=mx+b$ \\ \hline
			MSL & $\text{\textit{\textbf{w}}}_\pi$ & policy sym. loss weight & 0.05  \\ 
			\hline
			PSL & $\text{\textit{\textbf{w}}}_\pi$ & policy sym. loss weight & 0.0005  \\ \hline
		\end{tabular}
		\label{tab_asl:asl_hyperparameters_A4.1}
	\end{center}
\end{table}

\begin{table}[ht]
	\caption{Scenario \textbf{A4.2} --- ASL Hyperparameters}
	\begin{center}
		\begin{tabular}{|c|r|l|c|}
			\hline
			\textbf{Method} & \textbf{Symbol} & \textbf{Short Description} & \textbf{Value} \\ \hline
			all & TS & total time steps & \num{5e6} \\ \hline
			ASL & $\text{\textit{\textbf{w}}}_\pi$ & policy sym. loss weight & $0.1$  \\ 
			\hline
			ASL & $\boldsymbol{k}_\text{s}$ & max. distribution shift & 0.1 \\ \hline
			ASL & $\boldsymbol{k}_\text{d}$[reflec] & dead zone (sym. plane) & 0.2 \\ \hline
			ASL & $\boldsymbol{\hat{g}}(a)$ form & function form of $\boldsymbol{g}(a)$ & $y=mx+b$ \\ \hline
			MSL & $\text{\textit{\textbf{w}}}_\pi$ & policy sym. loss weight & 0.05  \\ 
			\hline
			PSL & $\text{\textit{\textbf{w}}}_\pi$ & policy sym. loss weight & 0.001  \\ \hline
		\end{tabular}
		\label{tab_asl:asl_hyperparameters_A4.2}
	\end{center}
\end{table}

\begin{table}[ht]
	\caption{Scenario \textbf{A4.3} --- ASL Hyperparameters}
	\begin{center}
		\begin{tabular}{|c|r|l|c|}
			\hline
			\textbf{Method} & \textbf{Symbol} & \textbf{Short Description} & \textbf{Value} \\ \hline
			all & TS & total time steps & \num{5e6} \\ \hline
			ASL & $\text{\textit{\textbf{w}}}_\pi$ & policy sym. loss weight & $0.1$  \\ 
			\hline
			ASL & $\boldsymbol{k}_\text{s}$ & max. distribution shift & 0.1 \\ \hline
			ASL & $\boldsymbol{k}_\text{d}$[reflec] & dead zone (sym. plane) & 0.2 \\ \hline
			ASL & $\boldsymbol{\hat{g}}(a)$ form & function form of $\boldsymbol{g}(a)$ & $y=mx+b$ \\ \hline
			MSL & $\text{\textit{\textbf{w}}}_\pi$ & policy sym. loss weight & 0.05  \\ 
			\hline
			PSL & $\text{\textit{\textbf{w}}}_\pi$ & policy sym. loss weight & 0.001  \\ \hline
		\end{tabular}
		\label{tab_asl:asl_hyperparameters_A4.3}
	\end{center}
\end{table}

\begin{table}[ht]
	\caption{Scenario \textbf{A4.4} --- ASL Hyperparameters}
	\begin{center}
		\begin{tabular}{|c|r|l|c|}
			\hline
			\textbf{Method} & \textbf{Symbol} & \textbf{Short Description} & \textbf{Value} \\ \hline
			all & TS & total time steps & \num{5e6} \\ \hline
			ASL & $\text{\textit{\textbf{w}}}_\pi$ & policy sym. loss weight & $0.1$  \\ 
			\hline
			ASL & $\boldsymbol{k}_\text{s}$ & max. distribution shift & 0.1 \\ \hline
			ASL & $\boldsymbol{k}_\text{d}$[reflec] & dead zone (sym. plane) & 0.2 \\ \hline
			ASL & $\boldsymbol{\hat{g}}(a)$ form & function form of $\boldsymbol{g}(a)$ & $y=mx+b$ \\ \hline
			MSL & $\text{\textit{\textbf{w}}}_\pi$ & policy sym. loss weight & 0.05  \\ 
			\hline
			PSL & $\text{\textit{\textbf{w}}}_\pi$ & policy sym. loss weight & 0.001  \\ \hline
		\end{tabular}
		\label{tab_asl:asl_hyperparameters_A4.4}
	\end{center}
\end{table}

\begin{table}[ht]
	\caption{Scenario \textbf{A4.5} --- ASL Hyperparameters}
	\begin{center}
		\begin{tabular}{|c|r|l|c|}
			\hline
			\textbf{Method} & \textbf{Symbol} & \textbf{Short Description} & \textbf{Value} \\ \hline
			all & TS & total time steps & \num{5e6} \\ \hline
			ASL & $\text{\textit{\textbf{w}}}_\pi$ & policy sym. loss weight & $0.05$  \\ 
			\hline
			ASL & $\boldsymbol{k}_\text{s}$ & max. distribution shift & 0.02 \\ \hline
			ASL & $\boldsymbol{k}_\text{d}$[reflec] & dead zone (sym. plane) & 0.2 \\ \hline
			ASL & $\boldsymbol{\hat{g}}(a)$ form & function form of $\boldsymbol{g}(a)$ & $y=mx+b$ \\ \hline
			MSL & $\text{\textit{\textbf{w}}}_\pi$ & policy sym. loss weight & 0.0005  \\ 
			\hline
			PSL & $\text{\textit{\textbf{w}}}_\pi$ & policy sym. loss weight & 0.0005  \\ \hline
		\end{tabular}
		\label{tab_asl:asl_hyperparameters_A4.5}
	\end{center}
\end{table}

\begin{table}[ht]
	\caption{Scenario \textbf{A4.6} --- ASL Hyperparameters}
	\begin{center}
		\begin{tabular}{|c|r|l|c|}
			\hline
			\textbf{Method} & \textbf{Symbol} & \textbf{Short Description} & \textbf{Value} \\ \hline
			all & TS & total time steps & \num{5e6} \\ \hline
			ASL & $\text{\textit{\textbf{w}}}_\pi$ & policy sym. loss weight & $0.05$  \\ 
			\hline
			ASL & $\boldsymbol{k}_\text{s}$ & max. distribution shift & 0.05 \\ \hline
			ASL & $\boldsymbol{k}_\text{d}$[reflec] & dead zone (sym. plane) & 0.2 \\ \hline
			ASL & $\boldsymbol{\hat{g}}(a)$ form & function form of $\boldsymbol{g}(a)$ & $y=mx+b$ \\ \hline
			MSL & $\text{\textit{\textbf{w}}}_\pi$ & policy sym. loss weight & 0.05  \\ 
			\hline
			PSL & $\text{\textit{\textbf{w}}}_\pi$ & policy sym. loss weight & 0.001  \\ \hline
		\end{tabular}
		\label{tab_asl:asl_hyperparameters_A4.6}
	\end{center}
\end{table}

\begin{table}[ht]
	\caption{Scenario \textbf{A5.1} --- ASL Hyperparameters}
	\begin{center}
		\begin{tabular}{|c|r|l|c|}
			\hline
			\textbf{Method} & \textbf{Symbol} & \textbf{Short Description} & \textbf{Value} \\ \hline
			all & TS & total time steps & \num{5e6} \\ \hline
			ASL & $\text{\textit{\textbf{w}}}_\pi$ & policy sym. loss weight & $0.2$  \\ 
			\hline
			ASL & $\boldsymbol{k}_\text{s}$ & max. distribution shift & 0.4 \\ \hline
			ASL & $\boldsymbol{k}_\text{d}$[reflec] & dead zone (sym. plane) & 0.2 \\ \hline
			ASL & $\boldsymbol{\hat{g}}(a)$ form & function form of $\boldsymbol{g}(a)$ & $y=mx+b$ \\ \hline
			MSL & $\text{\textit{\textbf{w}}}_\pi$ & policy sym. loss weight & 0.05  \\ 
			\hline
			PSL & $\text{\textit{\textbf{w}}}_\pi$ & policy sym. loss weight & 0.002  \\ \hline
		\end{tabular}
		\label{tab_asl:asl_hyperparameters_A5.1}
	\end{center}
\end{table}

\begin{table}[ht]
	\caption{Scenario \textbf{A5.2} --- ASL Hyperparameters}
	\begin{center}
		\begin{tabular}{|c|r|l|c|}
			\hline
			\textbf{Method} & \textbf{Symbol} & \textbf{Short Description} & \textbf{Value} \\ \hline
			all & TS & total time steps & \num{5e6} \\ \hline
			ASL & $\text{\textit{\textbf{w}}}_\pi$ & policy sym. loss weight & $0.2$  \\ 
			\hline
			ASL & $\boldsymbol{k}_\text{s}$ & max. distribution shift & 0.4 \\ \hline
			ASL & $\boldsymbol{k}_\text{d}$[reflec] & dead zone (sym. plane) & 0.2 \\ \hline
			ASL & $\boldsymbol{\hat{g}}(a)$ form & function form of $\boldsymbol{g}(a)$ & $y=mx+b$ \\ \hline
			MSL & $\text{\textit{\textbf{w}}}_\pi$ & policy sym. loss weight & 0.02  \\ 
			\hline
			PSL & $\text{\textit{\textbf{w}}}_\pi$ & policy sym. loss weight & 0.004  \\ \hline
		\end{tabular}
		\label{tab_asl:asl_hyperparameters_A5.2}
	\end{center}
\end{table}

\begin{table}[ht]
	\caption{Scenario \textbf{A6} --- ASL Hyperparameters}
	\begin{center}
		\begin{tabular}{|c|r|l|c|}
			\hline
			\textbf{Method} & \textbf{Symbol} & \textbf{Short Description} & \textbf{Value} \\ \hline
			all & TS & total time steps & \num{5e6} \\ \hline
			ASL & $\text{\textit{\textbf{w}}}_\pi$ & policy sym. loss weight & $0.05$  \\ 
			\hline
			ASL & $\boldsymbol{k}_\text{s}$ & max. distribution shift & 0.1 \\ \hline
			ASL & $\boldsymbol{k}_\text{d}$[reflec] & dead zone (sym. plane) & 0.2 \\ \hline
			ASL & $\boldsymbol{\hat{g}}(a)$ form & function form of $\boldsymbol{g}(a)$ & $y=mx+b$ \\ \hline
			MSL & $\text{\textit{\textbf{w}}}_\pi$ & policy sym. loss weight & 0.01  \\ 
			\hline
			PSL & $\text{\textit{\textbf{w}}}_\pi$ & policy sym. loss weight & 0.0005  \\ \hline
		\end{tabular}
		\label{tab_asl:asl_hyperparameters_A6}
	\end{center}
\end{table}

\clearpage % flushes all floats before proceeding

\section{Evaluation results}\label{app:C}
%\vspace{-10cm}

\begin{figure*}[!h]
\begin{adjustwidth}{-1.55cm}{-1.55cm}

	\newcommand\myscale[0]{0.848}
	\centering 
	\foreach \n in {1,...,20}{
	    \includegraphics[scale=\myscale]{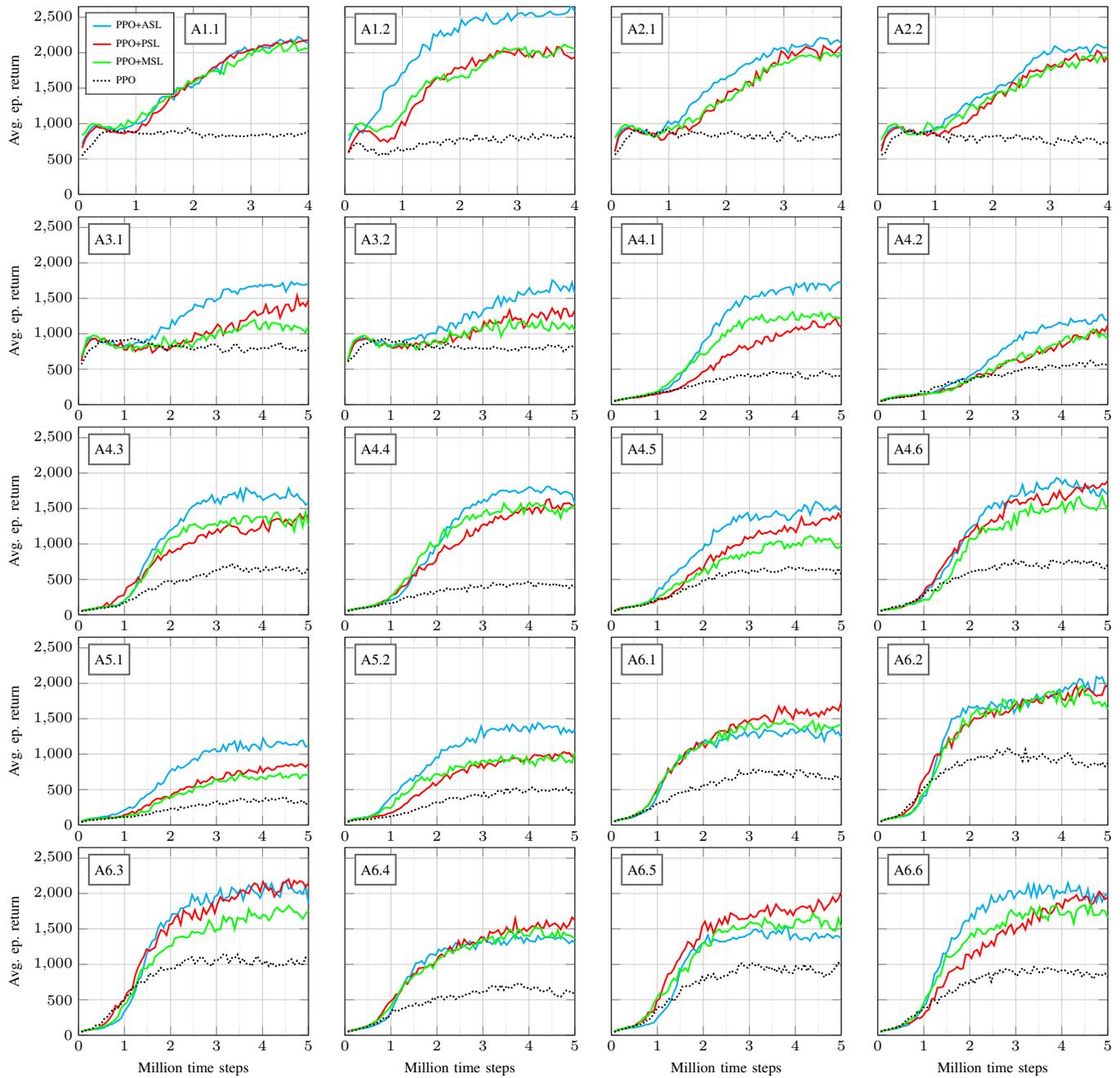}
	}
\end{adjustwidth}
\captionsetup{width=1.1\linewidth}
	\caption{Average learning curves for a batch of 12 instances per algorithm, per scenario. The deterministic neural network output of each policy was evaluated every 15 iterations, i.e., every 61440 time steps. The evaluated algorithms are: PPO without extensions (dotted black), PPO with Adaptive Symmetry Learning (solid blue), PPO with Proximal Symmetry Learning (solid red), PPO with generalized Mirror Symmetry Loss (solid green).}
	\label{fig:res_return}
\end{figure*}

\begin{figure*}[ht]
\begin{adjustwidth}{-1.55cm}{-1.55cm}

	\newcommand\myscale[0]{0.905}
	\centering 
	\foreach \n in {1,...,20}{
	    \includegraphics[scale=\myscale]{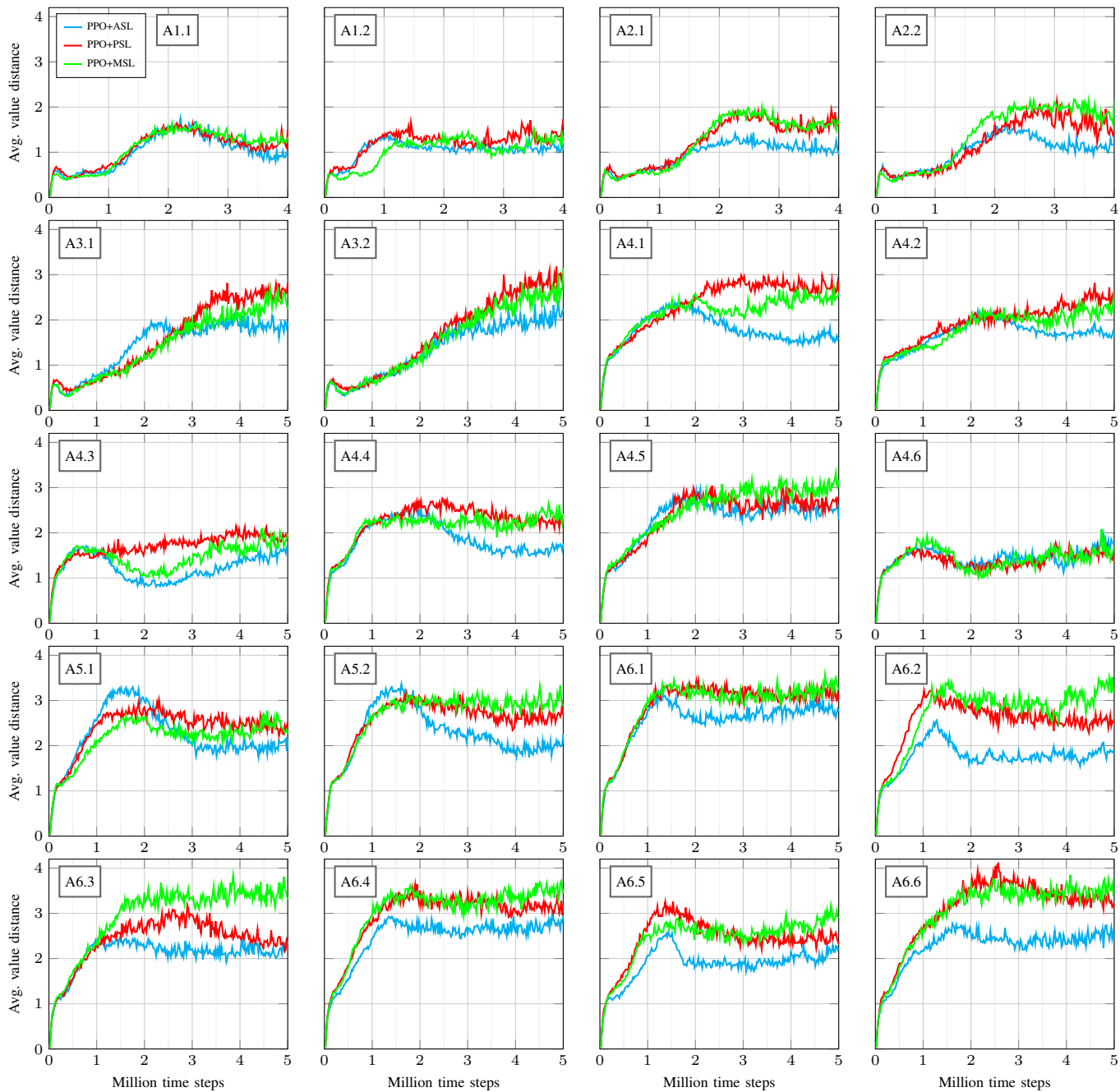}
	}
\end{adjustwidth}
\captionsetup{width=1.1\linewidth}
	\caption{Average absolute difference between the value estimates of the explored state and the corresponding symmetric state. Each data point is obtained every 5 iterations and is an average of all the explored states in the last iteration, averaged over 12 instances. The evaluated algorithms are: PPO with Adaptive Symmetry Learning (blue), PPO with Proximal Symmetry Learning (red), PPO with generalized Mirror Symmetry Loss (green).}
	
	\label{fig:res_valdiff}

\end{figure*}

\begin{figure*}[ht]
	\vspace{-2cm}
\begin{adjustwidth}{-1.55cm}{-1.55cm}
	\newcommand\myscale[0]{0.90}
	\centering 
	\foreach \n in {1,...,20}{
	    \includegraphics[scale=\myscale]{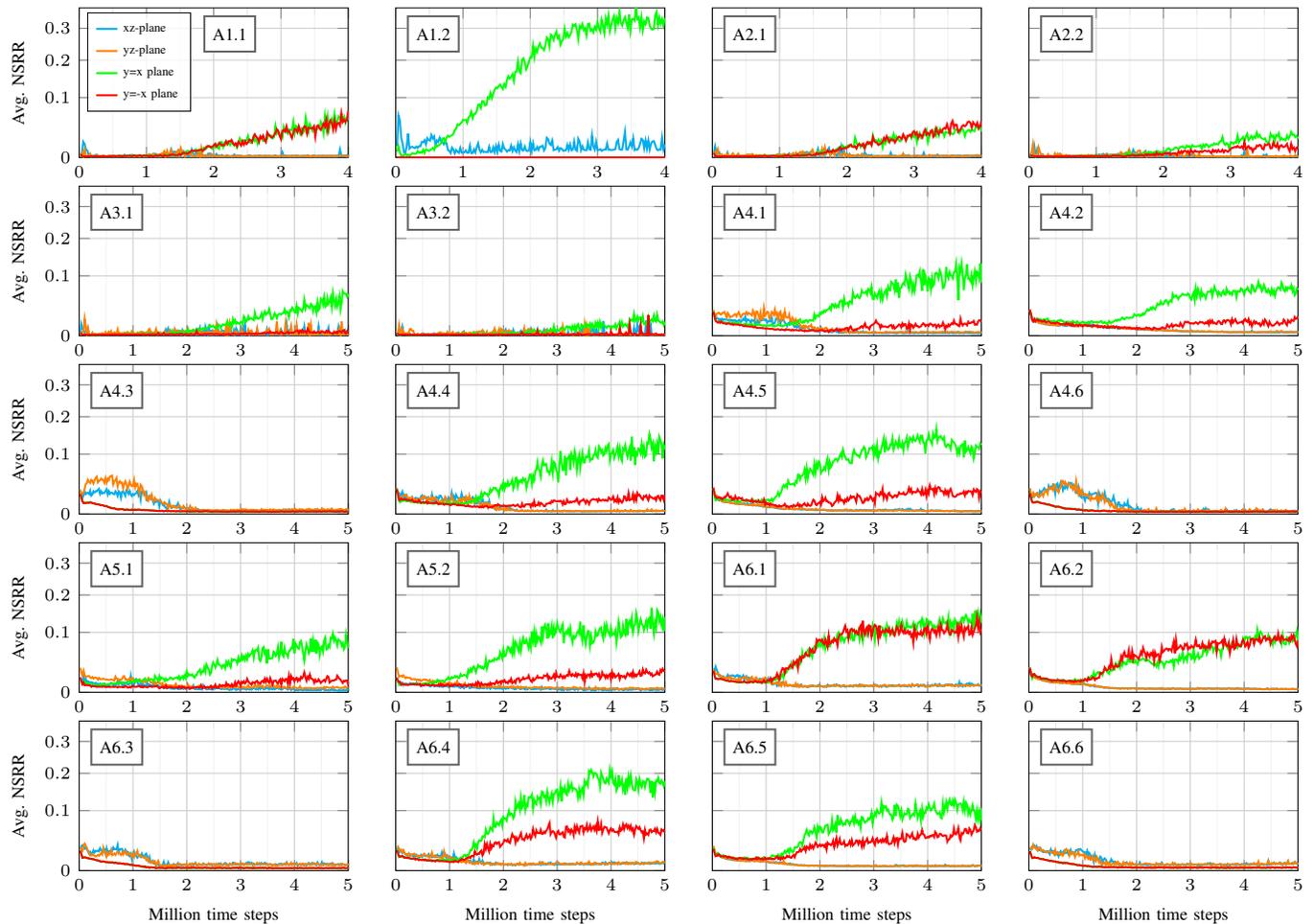}
	}
\end{adjustwidth}
\captionsetup{width=1.1\linewidth}
\vspace{-0.2cm}
	\caption{Average neutral state rejection ratio (NSRR), i.e. ratio of states that were rejected by the Adaptive Symmetry Learning algorithm (see Section~\ref{sec:neutral_exclusion}). Each data point is an average of 12 instances per symmetry plane, per scenario, obtained every 5 iterations. The evaluated symmetry planes are: xz-plane (blue), yz-plane (orange), y=x plane (green), y=-x plane (red).}
	\label{fig:res_nsrr}
\end{figure*}

\begin{figure*}[ht]
\begin{adjustwidth}{-1.55cm}{-1.55cm}

	\newcommand\myscale[0]{0.90}
	\centering 
	\foreach \n in {1,...,8}{
	    \includegraphics[scale=\myscale]{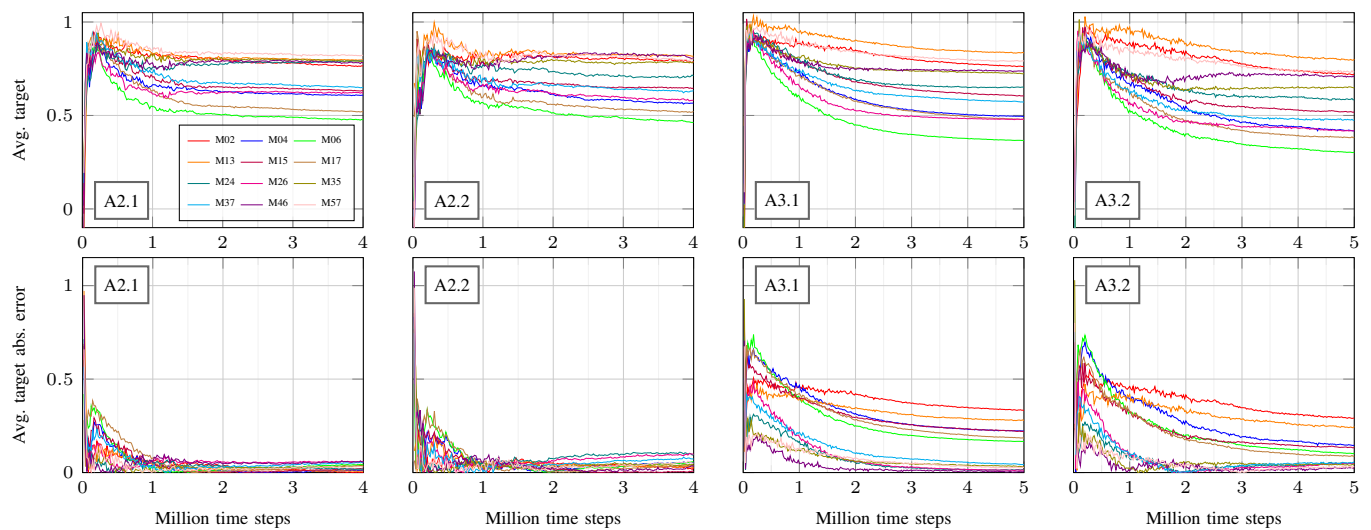}
	}
\end{adjustwidth}
\captionsetup{width=1.1\linewidth}
\vspace{-0.2cm}
	\caption{Average symmetry transformation (multiplier) target (above) and the respective absolute error (below) for the Adaptive Symmetry Learning algorithm. Each data point is an average of 12 instances per pair of symmetric action elements ($m_{0\rightarrow 2}$ to $m_{5\rightarrow 7}$), per scenario (\textbf{A2} and \textbf{A3}), obtained every 5 iterations.}
	\label{fig:res_target1}
\end{figure*}

\begin{figure*}[ht]
\begin{adjustwidth}{-1.55cm}{-1.55cm}

	\newcommand\myscale[0]{0.897}
	\centering 
	\foreach \n in {1,...,28}{
	    \includegraphics[scale=\myscale]{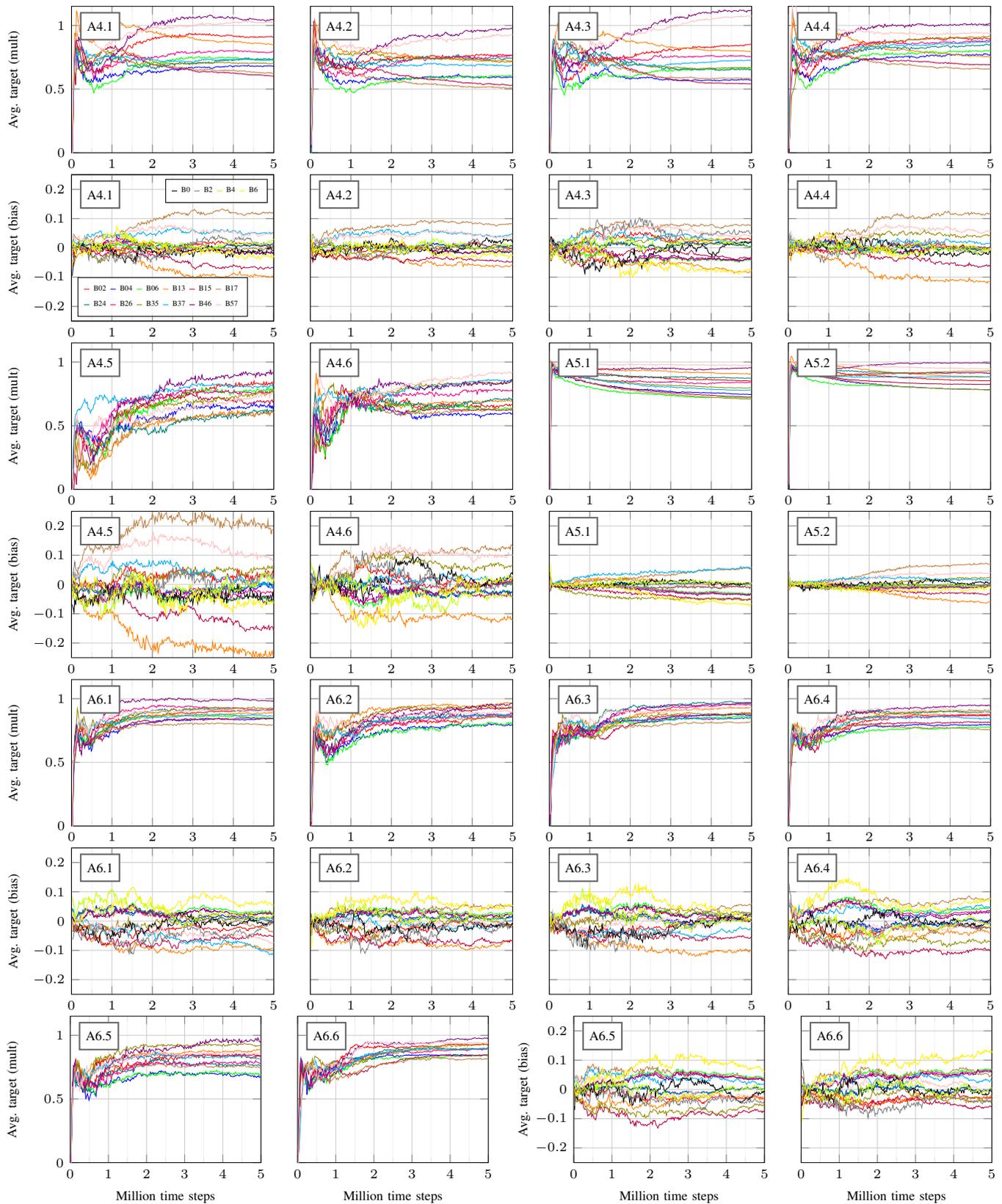}
	}
\end{adjustwidth}
	\captionsetup{width=1.1\linewidth}
	\caption{Average symmetry transformation (multiplier/bias) targets for the Adaptive Symmetry Learning algorithm. Each data point is an average of 12 instances per pair of symmetric action elements ($m_{0\rightarrow 2}$ to $m_{5\rightarrow 7}$ and $b_{0\rightarrow 2}$ to $b_{5\rightarrow 7}$) and reflexive relations/singles ($b_0$, $b_2$, $b_4$ and $b_6$), per scenario (\textbf{A4} to \textbf{A6}), obtained every 5 iterations.}
	
	\label{fig:res_target2}

\end{figure*}

\end{document}